\def\paperID{####} 
\def\confName{CVPR}
\def\confYear{2026}

\def\ONESPACE{ }
\def\OURMETHOD{ORCA}
\def\OUREXTENDED{Orchestrated Reasoning with Collaborative Agents for Document Visual Question Answering}

\def\paperTitle{\OURMETHOD: \OUREXTENDED}

\def\authorBlock{
    Aymen Lassoued$^{1,2}$, Mohamed Ali Souibgui$^{3}$, Yousri Kessentini$^{1}$ \\ 
    $^{1}$Digital Research Center of Sfax, SMARTS Laboratory, Sfax, Tunisia. \\
    $^{2}$École Polytechnique de Tunisie, University of Carthage \\
    $^{3}$Computer Vision Center, Universitat Autònoma de Barcelona\\

    {\tt\small \{aymen.lassoued@ept.ucar.tn, msouibgui@cvc.uab.cat, yousri.kessentini@crns.rnrt.tn\}}
}
\newif\ifreview
\reviewfalse  

\newif\ifarxiv
\arxivtrue 

\newif\ifcamera
\cameratrue 

\newif\ifrebuttal
\rebuttalfalse  

\newcommand{\arxiv}{%
  \reviewfalse
  \arxivtrue
  \camerafalse
  \rebuttalfalse
}

\arxiv
\pdfoutput=1
\documentclass[10pt,twocolumn,letterpaper]{article}
\ifreview \usepackage[review]{cvpr} \fi
\ifarxiv \usepackage[pagenumbers]{cvpr} \fi
\ifrebuttal \usepackage[rebuttal]{cvpr} \fi
\ifcamera \usepackage{cvpr} \fi

\usepackage{float}

\usepackage{algorithm}
\usepackage{algpseudocode}
\usepackage{graphicx}	
\usepackage{amsmath}	
\usepackage{amssymb}	
\usepackage{booktabs}
\usepackage{times}
\usepackage{microtype}
\usepackage{epsfig}
\usepackage{caption}
\usepackage{float}
\usepackage{placeins}
\usepackage{color, colortbl}
\usepackage{stfloats}
\usepackage{enumitem}
\usepackage{tabularx}
\usepackage{xstring}
\usepackage{multirow}
\usepackage{xspace}
\usepackage{url}
\usepackage{subcaption}
\usepackage{xcolor}
\usepackage{pifont}
\usepackage[hang,flushmargin]{footmisc}
\definecolor{myGreen}{RGB}{34, 139, 34}
\definecolor{myRed}{HTML}{FF6347}

\ifcamera \usepackage[accsupp]{axessibility} \fi

% ...

\ifarxiv  \fi

\newcommand{\R}[1]{{%
    \textbf{%
        \ifstrequal{#1}{1}{\textcolor{red}{R#1}}{%
        \ifstrequal{#1}{2}{\textcolor{blue}{R#1}}{%
        \ifstrequal{#1}{3}{\textcolor{magenta}{R#1}}{%
        \ifstrequal{#1}{4}{\textcolor{teal}{R#1}}{%
                           \textcolor{cyan}{R#1}%
        }}}}%
    }%
}}
\usepackage{tcolorbox} 
\definecolor{graypurple}{RGB}{80, 60, 100} 
\definecolor{lightgraypurple}{RGB}{210, 200, 240} 

\tcbset{
    colback=lightgraypurple!20, 
    colframe=graypurple, 
    coltitle=white, 
    fonttitle=\bfseries, 
    arc=5mm, 
    boxrule=1pt, 
    sharp corners,
    before skip=5pt, 
    after skip=5pt 
}

\usepackage{xr-hyper}

\makeatletter
\newcommand*{\addFileDependency}[1]{
  \typeout{(#1)}
  \@addtofilelist{#1}
  \IfFileExists{#1}{}{\typeout{No file #1.}}
}

\makeatother
\newcommand*{\myexternaldocument}[1]{
    \externaldocument{#1}
    \addFileDependency{#1.tex}
    \addFileDependency{#1.aux}
}

\definecolor{cvprblue}{rgb}{0.21,0.49,0.74}
\usepackage[pagebackref,breaklinks,colorlinks,allcolors=cvprblue]{hyperref}
\usepackage[capitalize]{cleveref}
\crefname{section}{Sec.}{Secs.}
\crefname{table}{Table}{Tables}
\crefname{figure}{Fig.}{Figs.}

\ifarxiv \crefname{appendix}{App.}{Apps.}
\else \crefname{appendix}{Suppl.}{Suppls.} \fi

\unless\ifarxiv \myexternaldocument{_supplementary} \fi

\begin{document}
\title{\paperTitle}
\author{\authorBlock}
\maketitle

\begin{abstract}
Document Visual Question Answering (DocVQA) remains challenging for existing Vision-Language Models (VLMs), especially under complex reasoning and multi-step workflows. Current approaches struggle to decompose intricate questions into manageable sub-tasks and often fail to leverage specialized processing paths for different document elements. We present \OURMETHOD: \OUREXTENDED, a novel multi-agent framework that addresses these limitations through strategic agent coordination and iterative refinement. \OURMETHOD \ONESPACE begins with a reasoning agent that decomposes queries into logical steps, followed by a routing mechanism that activates task-specific agents from a specialized agent dock.
Our framework leverages a set of specialized AI agents, each dedicated to a distinct modality, enabling fine-grained understanding and collaborative reasoning across diverse document components.
To ensure answer reliability, \OURMETHOD \ONESPACE employs a debate mechanism with stress-testing, and when necessary, a thesis-antithesis adjudication process. This is followed by a sanity checker to ensure format consistency. Extensive experiments on three benchmarks demonstrate that our approach achieves significant improvements over state-of-the-art methods, establishing a new paradigm for collaborative agent systems in vision-language reasoning.
\end{abstract}

\section{Introduction}
\label{sec:intro}

\begin{figure}[tp]
    \centering
    \includegraphics[width=0.9\linewidth, height=1.6\textheight, keepaspectratio]{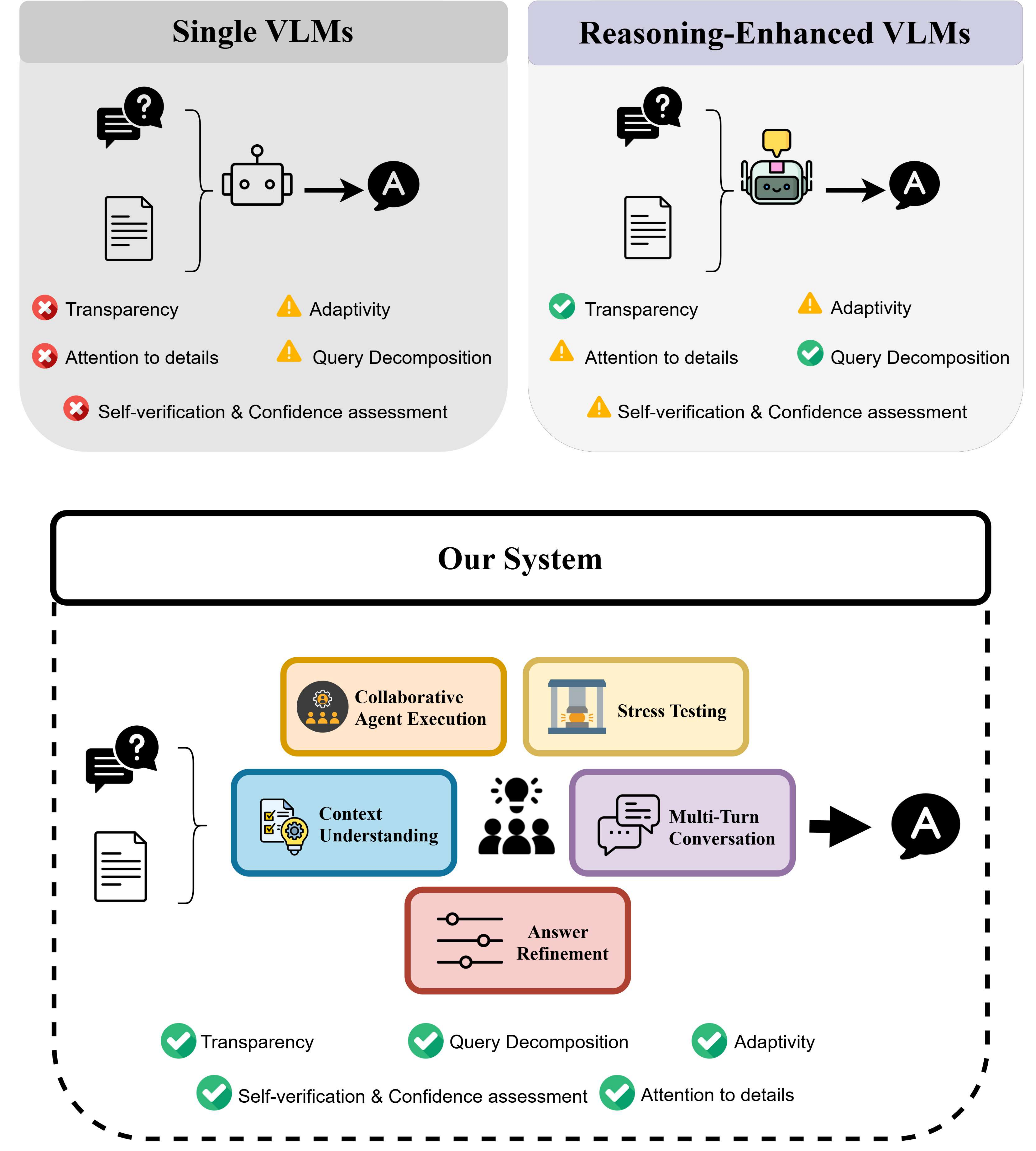}
    \caption{\textbf{Comparison of different approaches for DocVQA}. 
    Single-model VLMs and reasoning-enhanced VLMs lack critical capabilities such as adaptivity and self-verification. In contrast, \OURMETHOD{} introduces a feature-oriented, multi-agent design 
    achieving improved DocVQA performance as well as the missing capabilities
    in one unified framework.
    }
    \label{fig:intro_fig}
    \vspace{-2em}
\end{figure}

Answering questions based on single-page document images (DocVQA) \cite{mathew2021docvqa} requires more than simple information extraction. Questions often span multiple document modalities (e.g., text, tables, figures, handwritten content) and demand complex reasoning that current Vision Language Models (VLMs) struggle to perform reliably \cite{xu2025llava}. While VLMs have demonstrated impressive capabilities in document visual understanding \cite{bai2025qwen25vl, guo2025seed1, coreteam2025mimovl}, they frequently fall short when faced with multi-step reasoning, coordination across document elements, or specialized handling of diverse content types. 

Most existing DocVQA approaches typically rely on a single model to handle all aspects of document understanding, leading to suboptimal performance when questions involve heterogeneous information sources. For example, a question about data in a table with handwritten annotations requires expertise in both structured data extraction and OCR/HTR capabilities, skills that general-purpose models execute inconsistently \cite{lamm2024can}. Moreover, these models typically produce answers directly, without planning or exposing the reasoning steps behind their predictions.


Recently, some work has explored chain-of-thought (CoT) prompting and related techniques \cite{wei2022chain}, which encourage models to articulate intermediate reasoning steps before generating answers
\cite{zhang2024docassistant, zhang2024improvecot}.  These methods have improved both interpretability and accuracy on complex question answering tasks.
However, they still rely on a single model to handle all reasoning steps and document modalities, lacking mechanisms for content-aware specialization, self-verification, or adaptive agent selection based on document components. They do not employ specialized agents tailored to specific document elements (tables, charts, handwritten text), nor do they incorporate debate mechanisms to stress-test predictions or resolve conflicts between competing interpretations. Furthermore, without iterative refinement or cross-validation mechanisms, these models often produce answers without adequate confidence assessment \cite{banerjee2025llms}, limiting their reliability for complex DocVQA scenarios that require coordination across diverse document components.


To address these limitations, we introduce \textbf{\OURMETHOD}, a {multi-agent} framework that integrates explicit reasoning with collaborative execution for DocVQA and operates through five key stages. \textbf{(1) Context Understanding:} A thinking agent analyzes the document and question to generate a structured reasoning path and initial hypotheses. \textbf{(2) Collaborative Agent Execution:} Guided by the reasoning path, specialized agents tailored to document components such as tables, figures, forms, and handwritten text are dynamically activated to generate the answers. This introduces both reasoning capability and content-aware specialization. \textbf{(3) Debate Session} and \textbf{(4) Multi-turn Conversation:} To ensure reliability, \OURMETHOD{} incorporates a self-verification mechanism in which agents engage in iterative debate and reflection to stress-test and reconcile divergent responses. \textbf{(5) Answer Refinement:} A validation stage attends to fine-grained details and formatting consistency to refine the final output.

As illustrated in Figure~\ref{fig:intro_fig}, our design delivers key advantages over single-model and reasoning-enhanced DocVQA approaches, including transparency, query decomposition, adaptivity, self-verification, and fine-grained attention to detail.
Our contributions can be summarized as follows:
\begin{itemize}
    \item We propose a multi-agent framework that integrates explicit reasoning, specialized document understanding, and adversarial verification for robust single-page DocVQA.
    \item We achieved top performance on nearly all standard benchmarks compared to current state-of-the-art methods, demonstrating that our collaborative architecture with built-in debate and verification mechanisms produces more accurate and reliable answers for complex document question answering.
    \item We perform ablation studies to validate the contribution of each component, particularly the reasoning-guided agent selection and multi-turn conversation stages.
\end{itemize}
\section{Related Work}
\label{sec:related}

\noindent \textbf{Vision-Language Models for Document Understanding.} 
Document Visual Question Answering (DocVQA) has progressed from processing simple text-based documents to handling visually rich documents containing diverse content types~\cite{mathew2021docvqa, mathew2022infographicvqa, biten2019icdar, tito2023hierarchical}. Early approaches employed specialized multimodal transformers such as LayoutLM~\cite{xu2020layoutlm,huang2022layoutlmv3}, TILT~\cite{powalski2021going}, and Donut~\cite{kim2022donut}, which jointly model textual content, spatial layout, and visual features. Recent Vision-Language Models (VLMs), including BLIP-2~\cite{li2023blip2}, Qwen3-VL~\cite{qwen3vl2025}, InternVL~\cite{zhu2025internvl3,chen2023internvl}, and GLM-4.5V~\cite{zeng2024glm4v}, have demonstrated strong capabilities by integrating language understanding with visual perception~\cite{liu2023improved,liu2023visual,dai2023instructblip}. These models process document images directly, preserving layout and visual context. However, questions spanning multiple content modalities—such as extracting data from tables or interpreting figures alongside textual explanations—remain challenging, as they require coordination across different types of document elements~\cite{luo2024layoutllm,hu2024mplug,chen2024mllm}.

\noindent \textbf{Reasoning and Verification in Language Models.}
Explicit reasoning mechanisms have become increasingly important for complex question answering tasks. In~\cite{wei2022chain}, Chain-of-Thought (CoT) prompting introduced the concept of generating intermediate reasoning steps to improve complex question answering. Building on this,  Self-Consistency~\cite{wang2022self} demonstrated the benefits of sampling multiple reasoning paths. Extensions such as ReAct~\cite{yao2022react} and Reflexion~\cite{shinn2023reflexion} incorporate external tool use and iterative refinement. More recently, models with extended reasoning capabilities like DeepSeek-R1~\cite{deepseek2024r1} and OpenAI's o1~\cite{openai2024o1} have shown the value of longer thinking time for problem-solving. Parallel to these advances, verification mechanisms emerged to address the trustworthiness of model outputs, including those based on debate and argumentation~\cite{irving2018aisafety,du2023improving,chan2023chateval,liang2023encouraging}, which explore how multiple models can challenge and refine predictions through structured interaction. Together, these developments  inspire designing systems where reasoning transparency and cross-validation are essential.

\begin{figure*}[tp]
    \centering
    \includegraphics[width=0.95\textwidth, height=0.5\textheight, keepaspectratio]{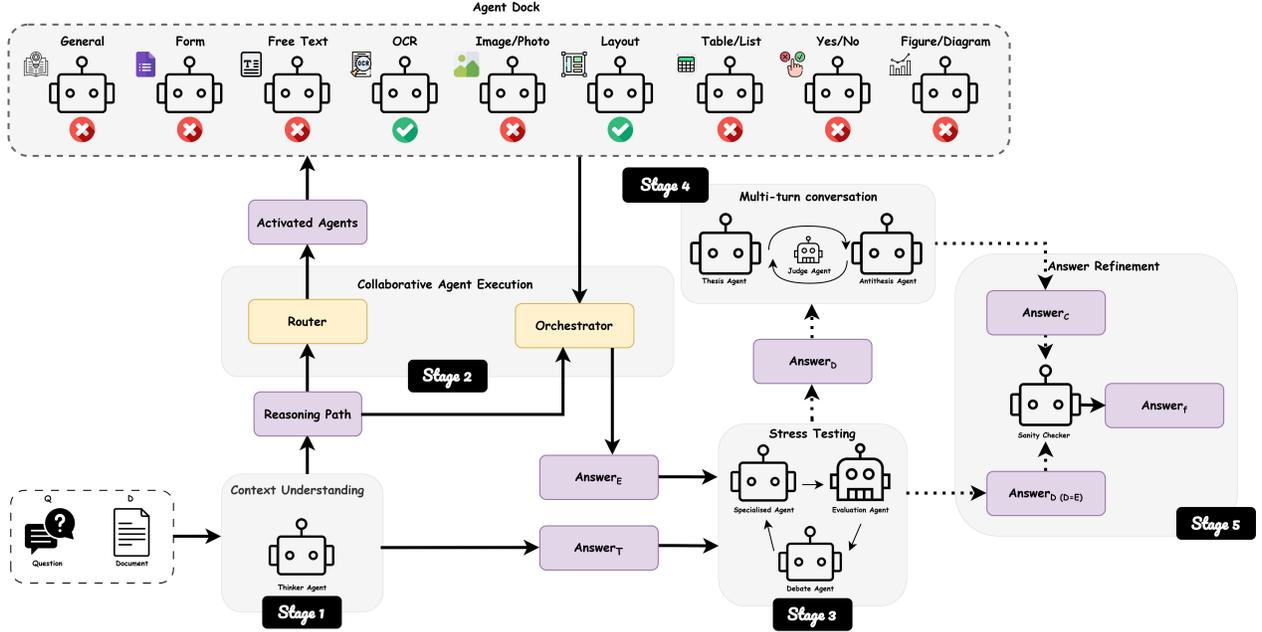}
    \caption{Overview of \textbf{\OURMETHOD}: A reasoning-guided multi-agent framework for Document Visual Question Answering operating through five stages: \textbf{(1) Context Understanding:} A thinker agent analyzes the question and document to generate both a reasoning path and initial answer ($a_T$). \textbf{(2) Collaborative Agent Execution:} A router selects relevant specialized agents from a dock of nine expert types (OCR, Layout, Table/List, Figure/Diagram, Form, Free Text, Image/Photo, Yes/No, and General), which an orchestrator sequences for optimal execution to produce an expert answer ($a_E$). \textbf{(3) Stress Testing:} When $a_E$ differs from $a_T$, a debate agent generates challenging questions to stress-test the specialized agent's confidence, with an evaluation agent assessing the responses to produce $a_D$. \textbf{(4) Multi-turn Conversation:} If stress testing indicates uncertainty, thesis and antithesis agents engage in structured three-turn debate under judge supervision to resolve conflicts and generate $a_C$. \textbf{(5) Answer Refinement:} A sanity checker performs final formatting corrections to ensure consistency with document conventions, producing the final answer ($a_F$).}
    \label{fig:main_fig}
    \vspace{-1.5em}
\end{figure*}

\noindent \textbf{Multi-Agent Frameworks.}
Multi-agent systems coordinate specialized components to tackle complex tasks~\cite{wu2023autogen,li2023camel,kim2024mdagents,hong2023metagpt}. Architectures such as Visual ChatGPT~\cite{wu2023visualchatgpt} and HuggingGPT~\cite{shen2023hugginggpt} use language models as controllers to orchestrate specialized vision and language modules. In document understanding contexts, different agents can focus on specific aspects such as table extraction, OCR, or layout analysis~\cite{su2020adapting,kannan2024smart,li2025metal}. The challenge lies in effectively routing questions to appropriate specialized models and coordinating their outputs when multiple modalities are involved.Our work explores how explicit reasoning can guide agent selection, how sequential orchestration can enable information flow between specialists, and how debate mechanisms can reconcile conflicting predictions when the thinker and expert agents reach divergent conclusions. Unlike prior orchestration frameworks such as Visual ChatGPT~\cite{wu2023visualchatgpt} and HuggingGPT~\cite{shen2023hugginggpt}, which rely on hand-crafted routing rules, ORCA introduces a VLM trained specifically for document-type routing via constrained generation with Turbo DFS decoding. In contrast to chain-of-thought and tool-use approaches, our reasoning path masking mechanism explicitly prevents confirmation bias in downstream agents. Finally, rather than applying verification universally, ORCA employs conditional activation, engaging debate in only 8.3\% of instances, concentrating computational overhead on cases where genuine uncertainty arises.

\section{Methodology}
\label{sec:method}

This section details our proposed framework for Document Visual Question Answering. Our approach employs a five-stage pipeline as illustrated in Figure \ref{fig:main_fig}.  At its core, our approach integrates explicit reasoning, specialized agent collaboration, and adversarial verification to handle the diverse and complex nature of document questions. We describe each stage in detail below.

\noindent \textbf{Problem Formulation.} Given a single-page document $\mathcal{D}$ and a natural language question $q$, our goal is to generate an accurate answer $a$ by reasoning over the document's multimodal content, which may include text, tables, figures, forms, and handwritten elements.

\subsection{Stage 1: Context Understanding}
\label{sec:context_understanding}

The first stage establishes the reasoning foundation for our framework. We employ a thinker agent $A_{\text{think}}$ based on GLM-4.5V-9B with thinking capabilities to analyze the question and document image jointly. The thinker agent generates two critical outputs: (1) a structured reasoning path $\mathcal{R}$ that decomposes the question into logical steps, and (2) an initial answer $a_T$ based on this reasoning process.

\begin{equation}
    (\mathcal{R}, a_T) = A_{\text{think}}(q, \mathcal{D})
\end{equation}

The reasoning path $\mathcal{R} = \{r_1, r_2, \ldots, r_n\}$ consists of $n$ intermediate reasoning steps that describe the cognitive process required to answer the question. For example, for a question "What is the total revenue in Q3?", the reasoning path might specify: $r_1$: "Locate the quarterly revenue table", $r_2$: "Find the Q3 column", $r_3$: "Extract the total revenue value". This explicit reasoning serves as a guide for subsequent agent selection and orchestration.

\subsection{Stage 2: Collaborative Agent Execution}
\label{sec:collaborative_execution}

The second stage dynamically selects and sequences specialized agents based on the reasoning path. This stage consists of three components: the agent dock, router, and orchestrator.

\noindent \textbf{Agent Dock.} We maintain  nine specialized agents, each designed to handle specific document content types:
\begin{itemize}
    \item $A_{\text{figure}}$: Handles diagrams and charts
    \item $A_{\text{yesno}}$: Processes yes/no questions
    \item $A_{\text{table}}$: Extracts information from tables and lists
    \item $A_{\text{layout}}$: Analyzes document layout structure
    \item $A_{\text{image}}$: Interprets photographs and images
    \item $A_{\text{ocr}}$: Recognizes handwritten and difficult text
    \item $A_{\text{text}}$: Processes free-form textual content
    \item $A_{\text{form}}$: Handles structured forms
    \item $A_{\text{other}}$: Addresses miscellaneous content types
\end{itemize}

All specialized agents are based on variants of Qwen3-VL-8B, fine-tuned for their respective tasks.

\noindent \textbf{Router.} The router agent $A_{\text{route}}$, 
plays a critical role in our framework by determining which specialized agents should process each document-question pair. We formulate this as a multi-label classification problem over nine agent types and employ several optimization strategies to ensure efficient and accurate routing.
Given the reasoning path $\mathcal{R}$, question $q$, and document $\mathcal{D}$, the router must predict a binary activation vector $\mathbf{v} \in \{0,1\}^9$ indicating which agents should be activated. 
Note that multiple agents can be simultaneously relevant for a single document-question pair.

We train the router on the Single-Page Document VQA dataset with ground-truth agent annotations. To improve model robustness and generalization, we incorporate data augmentation techniques. We employ Qwen2.5-VL-7B as the base architecture for $A_{\text{route}}$, unlike standard classification approaches that apply a sigmoid threshold to output logits, we treat routing as a constrained generation task. Thus, we use {Turbo DFS} (Depth-First Search with score-guided pruning) for decoding. Given the ranked candidate sequences from Turbo DFS, we apply a \textit{union strategy} to extract the final agent activation set. More details about training the router are available in Appendix \ref{supp:router}.


Our trained router can be used, therefore, to analyze the reasoning path $\mathcal{R}$, question $q$, and document $\mathcal{D}$ to determine which specialized agents are required. It outputs a binary activation vector $\mathbf{v} \in \{0,1\}^9$, where $v_i = 1$ indicates that agent $i$ should be activated.

\begin{equation}
    \mathbf{v} = A_{\text{route}}(q, \mathcal{D}, \mathcal{R})
\end{equation}

Thus, from the router we have $\mathcal{A}_{\text{active}} = \{A_i \mid v_i = 1\}$ denote the set of activated agents.

\noindent \textbf{Orchestrator.} The orchestrator determines the optimal execution order for activated agents. Given $n$ activated agents, it produces a sequence $\sigma = (\sigma_1, \sigma_2, \ldots, \sigma_n)$ where $\sigma_i \in \mathcal{A}_{\text{active}}$ represents the agent to execute at step $i$.

\begin{equation}
    \sigma = \text{Orchestrate}(\mathcal{A}_{\text{active}}, \mathcal{R}, q, \mathcal{D})
\end{equation}

Agents execute sequentially, with each agent receiving the output from its predecessor. For agent $\sigma_i$, the input consists of the question $q$, document $\mathcal{D}$, and the answer $a_{i-1}$ from the previous agent:

\begin{equation}
    a_i = \sigma_i(q, \mathcal{D}, a_{i-1})
\end{equation}

The final agent $\sigma_n$ additionally receives a masked version of the reasoning path $\mathcal{R}^*$, where occurrences of the answer are masked if they appear frequently in the reasoning steps. Specifically, if the answer $a_T$ appears more than the threshold $\tau$ we mask all occurrences.

\begin{equation}
    a_E = \sigma_n(q, \mathcal{D}, a_{n-1}, \mathcal{R}^*)
\end{equation}

So,  $a_E$ is the expert answer, representing the specialized agent's final answer. 

\subsection{Stage 3: Stress Testing Session}
\label{sec:debate}

In this stage, we compare the expert answer  $a_E$ with the thinker's answer $a_T$.  If $a_E = a_T$, we proceed directly to Stage 5 (sanity checking). If not, 
we initiate a stress testing session to assess the confidence of the expert system. The stress testing session consists of three agents (debate, specialized and evaluation) working together to stress-test the specialized agent's answer. This process is detailed in what follows.

\noindent \textbf{Debate Agent.} The goal of the debate agent $A_{\text{debate}}$ is to probe the reasoning behind the answer and identify potential weaknesses.
Hence, it generates challenging follow-up questions $q_{\text{debate}}$ based on the document $\mathcal{D}$, original question $q$, and expert answer $a_E$. 

\begin{equation}
    q_{\text{debate}} = A_{\text{debate}}(q, \mathcal{D}, a_E)
\end{equation}

\noindent \textbf{Specialized Agent.} The specialized agent $\sigma_n$ is the final agent from Stage 2 (with the final answer), in this stage, it receives the debate question and must provide: (1) a response $r_{\text{debate}}$ to the debate question, and (2) a potentially revised answer $a'_E$ to the original question.

\begin{equation}
    (r_{\text{debate}}, a'_E) = \sigma_n(q_{\text{debate}}, q, \mathcal{D}, a_E)
\end{equation}

\noindent \textbf{Evaluation Agent.} The evaluation agent $A_{\text{eval}}$ is an LLM-based model that assesses whether the specialized agent: (1) provided a coherent response, (2) stayed on-topic, and (3) maintained its original answer. The evaluation produces a binary decision $d \in \{\text{pass}, \text{fail}\}$.

\begin{equation}
    d = A_{\text{eval}}(q_{\text{debate}}, r_{\text{debate}}, a_E, a'_E)
\end{equation}

This stress testing repeats for two turns. If the specialized agent passes both turns (maintaining $a_E$ consistently with coherent responses), we set $a_D = a_E$ and proceed to Stage 5. If it fails in either turn, we proceed to Stage 4 (multi-turn conversation).

\subsection{Stage 4: Multi-turn Conversation}
\label{sec:communication}

This stage is executed if the stress testing session indicates uncertainty. At this point, we engage a multi-turn communication protocol engaging three distinct agents: thesis, antithesis, and judge.
Here,the thesis agent $A_{\text{thesis}}$ (same backbone as the specialized agents) advocates for answer $a_E$, while the antithesis agent $A_{\text{anti}}$ (InternVL3-8B-hf) generates an alternative answer $a_{\text{alt}}$ and argues against $a_E$.

\begin{equation}
    a_{\text{alt}} = A_{\text{anti}}(q, \mathcal{D}, a_E)
\end{equation}

If $A_{\text{anti}}$ cannot generate a distinct alternative (i.e., $a_{\text{alt}} = a_E$ or $a_{\text{alt}}$ contains $a_E$), we accept $a_E$ and proceed to Stage 5. Otherwise, we initiate a three-turn  debate with a defined protocol.

\noindent \textbf{Conversation Protocol.} Each turn $t$ consists of structured exchanges between the two agents. The antithesis agent presents its argument in a structured format containing three components:
\begin{itemize}
    \item \texttt{[REFERENCE]}: Evidence from the document supporting its position
    \item \texttt{[CRITICISM]}: Critique of the thesis agent's answer
    \item \texttt{[CONCLUSION]}: Its proposed answer and reasoning
\end{itemize}

\begin{equation}
    \text{arg}_{\text{anti}}^{(t)} = A_{\text{anti}}(q, \mathcal{D}, a_E, \text{summary}^{(t-1)})
\end{equation}

\noindent The thesis agent receives only the \texttt{[REFERENCE]} and \texttt{[CRITICISM]} components and responds by defending its position and addressing the criticism:

\begin{equation}
    \text{arg}_{\text{thesis}}^{(t)} = A_{\text{thesis}}(q, \mathcal{D}, a_E, \text{arg}_{\text{anti}}^{(t)}[\text{REF, CRIT}], \text{summary}^{(t-1)})
\end{equation}

During this debate, the  judge agent $A_{\text{judge}}$ (LLM-based) performs three functions: (1) Evaluates the \texttt{[CONCLUSION]} sections to determine if either agent has been convinced to change its position, (2) Generates a summary of the discussion for the next turn and (3) If no convincement occurs after three turns, analyzes the full debate transcript linguistically to determine which agent demonstrated greater confidence. Thus, after each turn $t$:
\begin{equation}
    (\text{convinced}, \text{summary}^{(t)}) = A_{\text{judge}}(\text{arg}_{\text{thesis}}^{(t)}, \text{arg}_{\text{anti}}^{(t)})
\end{equation}

The final answer $a_C$ is determined when one agent is convinced or after three turns, based on the judge's confidence assessment. 

\subsection{Stage 5: Answer Refinement}
\label{sec:refinement}

The final stage ensures formatting consistency between the predicted answer and the source document. A sanity checker agent $A_{\text{sanity}}$ receives the question $q$, document $\mathcal{D}$, and the answer from the previous stage (either $a_E$, $a_D$, or $a_C$). It performs two specific operations: Correction of missing spaces that appear in the document but not in the answer and Adjustment of punctuation to match the document's formatting conventions
\begin{equation}
    a_F = A_{\text{sanity}}(q, \mathcal{D}, a_{\text{prev}})
\end{equation}

Hence, $a_F$ is the final output answer. This stage ensures that the answer maintains fidelity to the document's original formatting, which is particularly important for DocVQA evaluation metrics.

More detailed implementation details can be found in Appendix \ref{supp:algos}

\vspace{-0.5em}
\section{Experiments}
\label{sec:experiments}

We evaluate our framework on three document understanding benchmarks to answer the following questions: (1) Does our reasoning-guided multi-agent approach improve DocVQA accuracy compared to existing VLMs? and (2) What is the contribution of each stage in our pipeline?

\subsection{Experiment Setup}

\noindent\textbf{Implementation Details.} Our framework consists of multiple specialized components: a thinker agent, nine specialized agents in the agent dock, a router agent, debate agents, thesis and antithesis agents, an evaluation agent (LLM-based), a judge agent (LLM-based), and a sanity checker. For \OURMETHOD (Qwen3VL-8B-Instruct), the thinker agent uses GLM-4.5V-9B~\cite{zeng2024glm4v} with thinking capabilities, while all other agents (specialized agents, debate agents, thesis agent, and sanity checker) are based on Qwen3VL-8B-Instruct~\cite{qwen2025qwen3-vl-8b}, and the antithesis agent is based on InternVL3-8B-hf~\cite{zhu2025internvl3, InternVL3-8B-hf}. The evaluation and judge agents use Qwen3-1.7B~\cite{qwen2025qwen3-1.7b} for robust assessment. The multi-agent debate mechanism enables collaborative reasoning, while the thinking agent provides enhanced chain-of-thought reasoning. All experiments are conducted on 4 NVIDIA L4 GPUs, each with 24GB VRAM (96GB total VRAM) and 175GB RAM. Details of agent configurations and hyperparameters are provided in the supplementary materials.
We evaluate our approach on three challenging document understanding benchmarks
(Single-Page DocVQA~\cite{mathew2021docvqa}, InfographicsVQA~\cite{mathew2022infographicvqa} and  OCRBench-v2 (en)~\cite{liu2024ocrbench}), 
Following standard evaluation protocols~\cite{mathew2021docvqa,mathew2022infographicvqa}, we report {ANLS} scores for {Single-Page DocVQA} and {InfographicsVQA}. 
For {OCRBench-v2}, we employ its official multi-dimensional evaluation suite with six task-specific metrics. The final score represents the average across all dimensions. More details about implementation details are available in Appendix~\ref{appendix:impl_details}.


\subsection{Main Results}
The overall performance of our method compared to other approaches is available in
Table~\ref{tab:main-results} and Table~\ref{tab:ocrbench_v2_split}, where we test the performance for DocVQA and OCRBench, respectively.
\begin{table}[t]
\centering
\caption{Performance comparison across DocVQA benchmarks. Models are categorized by architectural paradigm.}
\label{tab:main-results}
\resizebox{\columnwidth}{!}{
\begin{tabular}{l|c|cc|c}
\toprule
\textbf{Model} & \textbf{Open-source} & \textbf{DocVQA} & \textbf{InfoVQA} & \textbf{Avg.} \\
\midrule
\multicolumn{5}{l}{\textit{Document Understanding Models}} \\
LayoutLMv2 LARGE~\cite{xu2021layoutlmv2} & \checkmark & 86.7 & 28.3 & 57.5 \\
Text-Monkey~\cite{li2024monkey} & \checkmark & 66.7 & 28.6 & 47.7 \\
DocOwl-1.5~\cite{hu2024mplug} & \checkmark & 82.2 & 50.7 & 66.5 \\
\midrule
\multicolumn{5}{l}{\textit{General-Purpose VLMs}} \\
Qwen2-VL~\cite{bai2025qwen25vl} & \checkmark & 96.7 & 84.7 & 90.7 \\
InternVL2-Pro~\cite{chen2024internvl} & \checkmark & 95.1 & 83.3 & 89.2 \\
Molmo-72B~\cite{deitke2024molmo} & \checkmark & 93.5 & 81.9 & 87.7 \\
DeepSeek-VL2~\cite{wu2024deepseekvl2mixtureofexpertsvisionlanguagemodels} & \checkmark & 93.3 & 78.1 & 85.7 \\
LLaVA-One-Vision-8B~\cite{li2024llava} & \checkmark & 94.8 & 78.4 & 86.6 \\
Gemini Pro 1.5~\cite{team2024gemini} & \texttimes & 86.5 & 72.7 & 79.6 \\
Claude-3.7 Sonnet~\cite{claude2024anthropic} & \texttimes & 94.1 & 65.5 & 79.8 \\
Qwen-VL-Max (single)~\cite{wang2024qwen2} & \texttimes & 93.1 & 73.4 & 83.3 \\
GPT-4 Turbo + Textract~\cite{achiam2023gpt} & \texttimes & 87.4 & 71.9 & 79.7 \\
GPT-4o~\cite{achiam2023gpt} & \texttimes & 93.0 & 82.1 & 87.6 \\
\midrule
\multicolumn{5}{l}{\textit{Reasoning-Enhanced OS Models}} \\
MiMo-VL-7B-RL~\cite{coreteam2025mimovl} & \checkmark & 95.0 & 88.1 & 91.6 \\
VideoLLaMA3-7B~\cite{zhang2025videollama} & \checkmark & 95.0 & 78.9 & 87.0 \\
Gemma-3-27B-IT~\cite{team2025gemma} & \checkmark & 86.6 & 70.6 & 78.6 \\
\midrule
\multicolumn{5}{l}{\textit{Baseline VLMs (Single-Model)}} \\
Qwen2.5-VL-7B-Instruct~\cite{qwen2025qwen2.5-vl-7b} & \checkmark & 95.7 & 77.7 & 86.7 \\
Qwen3VL-4B-Instruct~\cite{qwen2025qwen3-vl-4b} & \checkmark & 95.3 & 80.3 & 87.8 \\
Qwen3VL-8B-Instruct~\cite{qwen2025qwen3-vl-8b} & \checkmark & 96.1 & 83.1 & 89.6 \\
\midrule
\multicolumn{5}{l}{\textit{\OURMETHOD{} (Multi-Agent Framework)}} \\
\OURMETHOD{} (Qwen2.5-VL-7B) & \checkmark & 96.4 {\scriptsize \textcolor{darkgray}{(+0.7)}} & 86.9 {\scriptsize \textcolor{darkgray}{(+9.2)}} & 91.7 {\scriptsize \textcolor{darkgray}{(+5.0)}} \\
\OURMETHOD{} (Qwen3VL-4B) & \checkmark & 96.0 {\scriptsize \textcolor{darkgray}{(+0.7)}} & 85.4 {\scriptsize \textcolor{darkgray}{(+5.1)}} & 90.7 {\scriptsize \textcolor{darkgray}{(+2.9)}} \\
\OURMETHOD{} (Qwen3VL-8B) & \checkmark & \textbf{97.2} {\scriptsize \textcolor{darkgray}{(+1.1)}} & \textbf{88.0} {\scriptsize \textcolor{darkgray}{(+4.9)}} & \textbf{92.6} {\scriptsize \textcolor{darkgray}{(+3.0)}} \\
\midrule
\multicolumn{5}{l}{\textit{Relative Improvements}} \\
Average Gain & -- & +0.8\% & +6.4\% & +3.6\% \\
\bottomrule
\end{tabular}
}
\end{table}

\noindent\textbf{Performance on DocVQA Tasks.} 
As shown in Table~\ref{tab:main-results}, we evaluate across two document visual question answering benchmarks: DocVQA~\cite{mathew2021docvqa} and InfographicVQA~\cite{mathew2022infographicvqa}. We categorize the models into five groups: (1) Document Understanding Models, designed for document-centric tasks; (2) General-Purpose VLMs, vision-language models for broad multimodal tasks including both open-source and closed-source variants; (3) Reasoning-Enhanced Open-Source Models, approaches that incorporate explicit reasoning capabilities; (4) Baseline VLMs (Single-Model), the base models upon which our framework builds; and (5) \OURMETHOD{} (Multi-Agent Framework), our proposed models. 

As can be seen, the DocVQA benchmark has become highly competitive, with several models achieving strong results. We begin by document understanding models. These models perform well on single-page DocVQA tasks, where answers are simply extractive. However, they struggle on the InfographicVQA benchmark, which requires higher-level reasoning, cross-modal grounding, and complex layout understanding. Next, for General-purpose VLMs, they  demonstrate much stronger generalization capabilities. Models such as Qwen2-VL~\cite{bai2025qwen25vl} and InternVL2-Pro~\cite{chen2024internvl} exhibit remarkable performance gains across both benchmarks. These models can be further enhanced when equipped with explicit reasoning mechanisms, as shown by recent reasoning-oriented open-source systems such as MiMo-VL-7B-RL~\cite{coreteam2025mimovl}, which achieve notable improvements, particularly on InfographicVQA. Finally, it is clear that our proposed \OURMETHOD{} model achieves substantial improvements on both benchmarks. On DocVQA, our framework attains the best results among all compared approaches, delivering consistent yet modest average gains of +0.8\%. This corresponds to a 28.2\% relative error reduction (3.9\%$\rightarrow$2.8\%), reflecting substantial progress in a low-error regime. On InfographicVQA, we observe a significant improvement of +6.4\% on average. Notably, InfographicVQA demands sophisticated integration of textual and visual information across intricate infographic layouts. This validates our hypothesis that orchestrated multi-agent collaboration with specialized reasoning capabilities excels in complex scenarios where single models often struggle.



\begin{table}[t]
\centering
\caption{OCRBench-v2 performance comparison between baseline VLMs and ORCA across model scales. The results marked with * are not reproduced, they are given by the papers' authors.}
\label{tab:ocrbench_v2_split}
\resizebox{\columnwidth}{!}{
\setlength{\tabcolsep}{3pt}
\begin{tabular}{l|cccccccc|c}
\hline
Model & Rec & Ref & Spo & Ext & Par & Cal & Und & Rea & Avg \\
\hline
\multicolumn{10}{l}{\textit{Open-source LMMs}} \\
LLaVA-Next-8B* \cite{liu2024llava-next} & 41.3 & 18.8 & 0 & 49.5 & 21.2 & 17.3 & 55.2 & 48.9 & 31.5 \\
LLaVA-OV-7B* \cite{li2024llava} & 46.0 & 20.8 & 0.1 & 58.3 & 25.3 & 23.3 & 64.4 & 53.0 & 36.4 \\
TextMonkey* \cite{li2024monkey} & 39.1 & 0.7 & 0 & 19.0 & 12.2 & 19.0 & 61.1 & 40.2 & 23.9 \\
Molmo-7B* \cite{deitke2024molmo} & 52.4 & 21.3 & 0.1 & 45.5 & 7.6 & 28.5 & 65.3 & 55.0 & 34.5 \\
Cambrian-34B* \cite{tong2024cambrian} & 45.3 & 21.5 & 0 & 53.6 & 19.2 & 19.5 & 63.5 & 55.5 & 34.7 \\
Pixtral-12B* \cite{agrawal2024pixtral} & 48.9 & 21.6 & 0 & 66.3 & 35.5 & 29.8 & 66.9 & 53.7 & 40.3 \\
Nemotron Nano V2 VL* \cite{nvidia2024nemotron-nano} & 67.6 & 54.5 & 36.2 & \textbf{92.0} & 26.6 & \textbf{80.4} & 75.5 & 57.0 & 61.2 \\
Llama Nemotron VL 8B* \cite{nvidia2024llama-nemotron} & 70.2 & \textbf{69.1} & \textbf{61.8} & 81.4 & 39.2 & 31.9 & 73.1 & 54.7 & 60.2 \\
InternVL-3-14B* \cite{internvl3-2024} & 67.3 & 36.9 & 11.2 & 89.0 & 38.4 & 38.4 & 79.2 & 60.5 & 52.6 \\
Qwen2.5-VL-7B* \cite{qwen2025qwen2.5-vl-7b} & 68.8 & 25.7 & 1.2 & 80.2 & 30.4 & 38.2 & 73.2 & 56.2 & 46.7 \\
Qwen3-Omni-A3B* \cite{qwen2024qwen3-omni} & 72.3 & 62.0 & 45.6 & 93.5 & 20.8 & 67.0 & 74.1 & 55.3 & 61.3 \\
Qwen3-VL-4B-Instruct* \cite{qwen2025qwen3-vl-4b} & - & - & - & - & - & - & - & - & 63.7 \\
Qwen3-VL-8B-Instruct* \cite{qwen2025qwen3-vl-8b} & - & - & - & - & - & - & - & - & 65.4 \\
\hline
\multicolumn{10}{l}{\textit{\OURMETHOD{} (Multi-Agent Framework)}} \\
\OURMETHOD{} (Qwen2.5-VL-7B) & 72.3 & 28.6 & 3.4 & 84.7 & 34.2 & 41.9 & 75.8 & 61.5 & 50.3 {\scriptsize \textcolor{darkgray}{(+3.6)}} \\
\OURMETHOD{} (Qwen3VL-4B) & 81.1 & 48.2 & 24.3 & 89.3 & 60.1 & 57.2 & 86.6 & 71.6 & 64.8 {\scriptsize \textcolor{darkgray}{(+1.1)}} \\
\OURMETHOD{} (Qwen3VL-8B) & \textbf{83.2} & 51.3 & 28.4 & 89.8 & \textbf{62.3} & 59.4 & \textbf{88.7} & \textbf{73.7} & \textbf{67.1} {\scriptsize \textcolor{darkgray}{(+1.7)}} \\
\hline
\end{tabular}
}
\end{table}

\noindent\textbf{Performance on OCRBench-v2 Multi-Task.}
We further evaluate the models on OCRBench-v2 across eight critical OCR subtasks: Recognition (Rec), Referring (Ref), Spotting (Spo), Extraction (Ext), Parsing (Par), Calculation (Cal), Understanding (Und), and Reasoning (Rea). The results are presented in  Table~\ref{tab:ocrbench_v2_split}.
As can be seen, \OURMETHOD{} achieves consistent improvements across all model scales, with gains inversely correlated to model capacity: +3.6 points for Qwen2.5-VL-7B versus +1.7 for Qwen3VL-8B. This suggests that our approach of multi-agent collaboration provides greater benefits for smaller models by compensating for individual capacity limitations through specialized agent expertise. Notably, our framework demonstrates especially strong improvements in challenging tasks such as Understanding, Reasoning, and Spotting for \OURMETHOD{} over its single-model baseline, directly reflecting \OURMETHOD{}'s design objectives of structured reasoning and modality-specific specialization. These gains are concentrated precisely where specialized agent collaboration is most effective. \OURMETHOD{} with Qwen3VL-8B configuration reaches 67.1\% average performance, demonstrating that our framework

addresses the inherent complexity of document understanding tasks that challenge monolithic vision-language models.

\noindent\textbf{Generalization to Chart-Centric Reasoning.}
To evaluate generalization beyond document VQA, we additionally assess \OURMETHOD{} on ChartQA, a benchmark requiring visual and numerical reasoning over charts and figures. \OURMETHOD{} (Qwen3VL-8B) achieves 90.1\%, improving over the single-model baseline of 85.7\% (+4.4\%). This improvement is consistent with \OURMETHOD{}'s design, as chart understanding benefits directly from the dedicated figure/diagram agent and the multi-turn debate mechanism for resolving numerical ambiguities. Beyond document-centric tasks, \OURMETHOD{} also improves performance on VQAv2 by +4.7\% over the single-model baseline, suggesting that the orchestrated reasoning pipeline extends to broader vision-language question answering scenarios.

\subsection{Inference Latency and Cost Analysis}

Table~\ref{tab:latency} reports inference latency measured on 4$\times$ NVIDIA L4 GPUs using vLLM acceleration. \OURMETHOD{} introduces moderate overhead relative to single-model baselines, mitigated by three key optimizations: (1) vLLM inference acceleration providing approximately 5$\times$ speedup over standard sequential execution; (2) conditional execution that bypasses the debate stages when the thinker and expert agents agree, which occurs in 77\% of cases; and (3) backbone reuse across stages, avoiding redundant model loading. In the early-termination regime (Stages 1--2 only), \OURMETHOD{} incurs approximately 4--6$\times$ overhead while delivering +2--3\% improvement on complex tasks, making it suitable for latency-sensitive deployments. Full pipeline execution is recommended for accuracy-critical applications such as processing higly sensitive documents, where the accuracy-latency trade-off is favorable. Notably, scaling monolithic models beyond 100B parameters achieves comparable accuracy at substantially higher memory and deployment cost, making \OURMETHOD{} a compute-efficient alternative that improves reasoning quality through structured orchestration rather than brute-force parameter scaling. Furthermore, intermediate reasoning traces and routing decisions generated during inference can be logged and reused as supervision signals for training on downstream tasks, further amortizing the compute cost over time.

\begin{table}[t]
\centering
\caption{Inference latency comparison on 4$\times$ NVIDIA L4 GPUs with vLLM optimization. Early-termination applies when thinker and expert agents agree, bypassing Stages 3--5 (77\% of cases).}
\label{tab:latency}
\resizebox{\columnwidth}{!}{%
\begin{tabular}{l|c|cc}
\toprule
\textbf{Configuration} & \textbf{Latency (s)} & \textbf{DocVQA} & \textbf{InfoVQA} \\
\midrule
Baseline (Qwen3VL-8B)           & 0.3--0.8  & 96.1 & 83.1 \\
\OURMETHOD{} Early-Termination  & 2.9--4.5  & 96.7 & 86.9 \\
\OURMETHOD{} Full Pipeline      & 9.6--13.1 & 97.2 & 88.0 \\
\bottomrule
\end{tabular}
}
\end{table}

\subsection{Ablation Studies}

We systematically evaluate each component's contribution using the three benchmarks. We start by ablating individual stages in Table~\ref{tab:ablation_stages} and report the following findings.

\begin{table}[t]
\centering
\caption{Stage-wise ablation study on Qwen3VL-8B. Each row isolates the impact of removing a specific stage from the complete \OURMETHOD{} framework.}
\label{tab:ablation_stages}
\resizebox{\columnwidth}{!}{
\begin{tabular}{l|ccc}
\toprule
\textbf{Configuration} & \textbf{DocVQA} & \textbf{InfoVQA} & \textbf{OCRBench-v2} \\
\midrule
\OURMETHOD{} (Full) & \textbf{97.2} & \textbf{88.0} & \textbf{67.1} \\
\midrule
\multicolumn{4}{l}{\textit{Individual Component Ablations}} \\
w/o Stage 1 (Reasoning) & 96.5 {\scriptsize (-0.7)} & 84.9 {\scriptsize (-3.1)} & 66.1 {\scriptsize (-1.0)} \\
w/o Stage 2 (Collaborative Agents) & 96.3 {\scriptsize (-0.9)} & 84.1 {\scriptsize (-3.9)} & 66.0 {\scriptsize (-1.1)} \\
w/o Stage 3 (Stress Testing) & 97.0 {\scriptsize (-0.2)} & 87.5 {\scriptsize (-0.5)} & 66.9 {\scriptsize (-0.2)} \\
w/o Stage 4 (Multi-turn Debate) & 96.9 {\scriptsize (-0.3)} & 87.2 {\scriptsize (-0.8)} & 66.7 {\scriptsize (-0.4)} \\
w/o Stage 5 (Answer Refinement) & 97.1 {\scriptsize (-0.1)} & 87.9 {\scriptsize (-0.1)} & 67.0 {\scriptsize (-0.1)} \\
\midrule
\multicolumn{4}{l}{\textit{Cumulative Ablations}} \\
w/o Stages 4+5 & 96.8 {\scriptsize (-0.4)} & 87.6 {\scriptsize (-0.4)} & 66.9 {\scriptsize (-0.2)} \\
w/o Stages 3+4+5 (All Verification) & 96.7 {\scriptsize (-0.5)} & 86.9 {\scriptsize (-1.1)} & 66.3 {\scriptsize (-0.8)} \\
w/o Stages 2--5 & 96.2 {\scriptsize (-1.0)} & 84.1 {\scriptsize (-3.9)} & 65.6 {\scriptsize (-1.5)} \\
Baseline (Single Model) & 96.1 & 83.1 & 65.4 \\
\bottomrule
\end{tabular}
}
\end{table}

\noindent\textbf{Core Components Drive Performance.} 
Stages 1 (Reasoning) and 2 (Collaborative Agents) constitute the framework's core components. Removing Stage 1 causes substantial drops across all benchmarks, demonstrating that strategic reasoning paths provide essential task decomposition. Stage 2 ablation shows even larger impact, with performance approaching baseline levels, confirming that multi-agent specialization represents our core architectural innovation.

\noindent\textbf{Verification Stages Provide Targeted Refinement.} 
Stages 3--5 contribute incrementally with smaller magnitude improvements. Stage 3 (Stress Testing) yields 0.2--0.5 point gains, Stage 4 (Multi-turn Debate) adds 0.3--0.8 points, while Stage 5's minimal impact primarily addresses formatting inconsistencies rather than semantic accuracy. Cumulatively removing all verification stages causes slight degradation, indicating complementary rather than foundational roles.
\begin{figure*}[!t]
    \centering
    \includegraphics[width=0.95\textwidth, height=0.35\textheight, keepaspectratio]{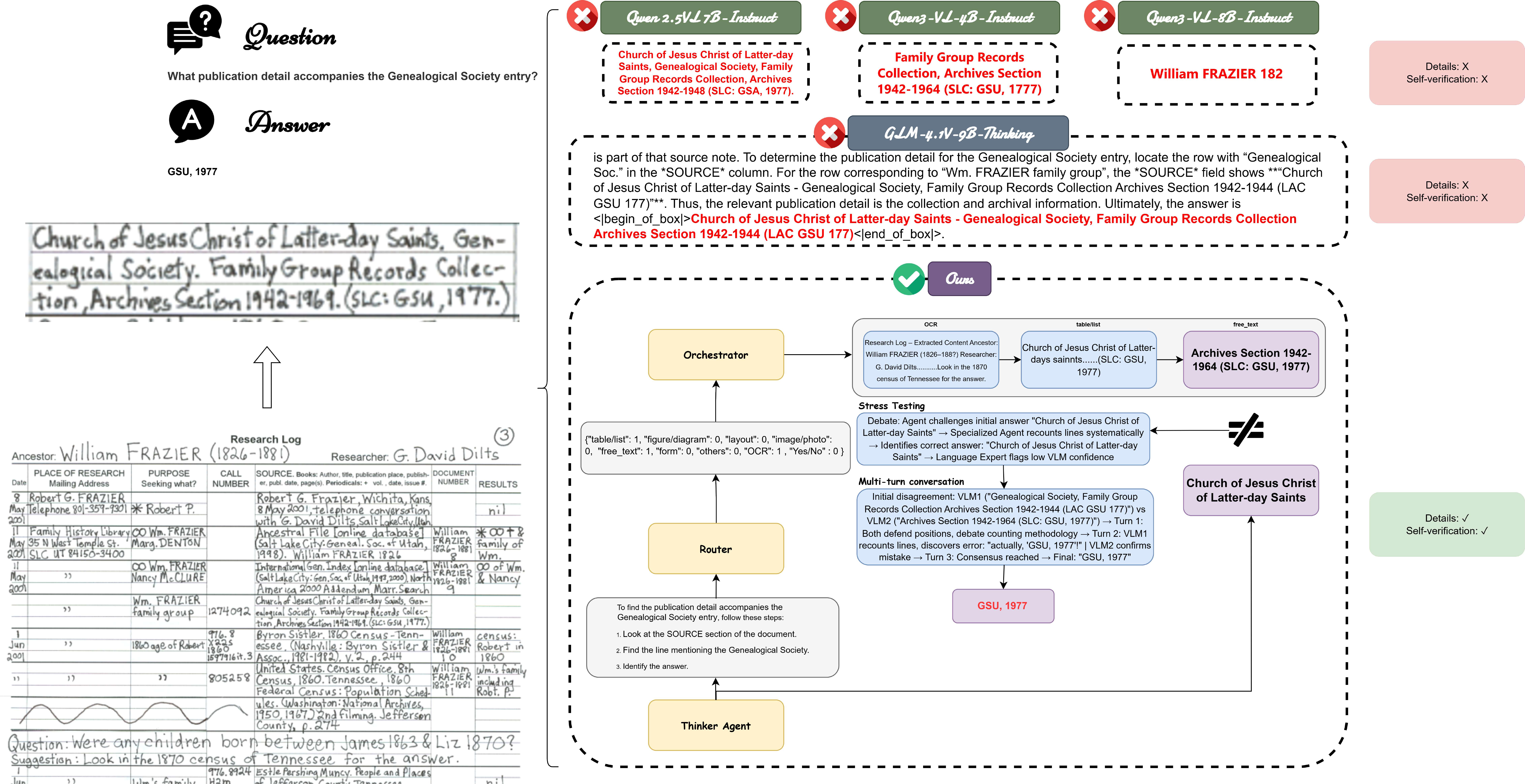}
    \caption{A case study demonstrating \textbf{\OURMETHOD}'s multi-agent reasoning pipeline on a complex visual document question. Better viewed with zoom. 
    \textbf{Question:} What publication detail accompanies the Genealogical Society entry?. \textbf{GT answer:} ``GSU, 1977''.}
    \label{fig:case_study}
    \vspace{-0.5em}
\end{figure*}

\noindent\textbf{Selective Activation Explains Modest Verification Impact.} 
Following the previous findings, we investigate the cause of the complementary roles of stages 3 and 4. An analysis of 500 randomly sampled test instances from the validation sets of the single-page DocVQA and InfographicsVQA datasets reveals that Stage 3 activates only when expert and thinker agents disagree ($a_E \neq a_T$), occurring in 23.4\% of cases. Among these, 35.7\% fail stress testing and proceed to Stage 4 debate. Consequently, multi-turn debate is engaged in merely 8.3\% of instances. This low activation frequency accounts for modest aggregate contributions, though these stages potentially enhance reliability in ambiguous edge cases where disagreement signals genuine uncertainty.

\begin{table}[t]
\centering
\caption{Reasoning path masking ablation on Qwen3VL-8B. Masking prevents confirmation bias while preserving strategic guidance.}
\label{tab:ablation_masking}
\resizebox{\columnwidth}{!}{
\begin{tabular}{l|ccc}
\toprule
\textbf{Component Variation} & \textbf{DocVQA} & \textbf{InfoVQA} & \textbf{OCRBench-v2} \\
\midrule
No reasoning masking & 96.5 & 87.6 & 66.4 \\
Reasoning masking (Full) & \textbf{97.2} & \textbf{88.0} & \textbf{67.1} \\
\midrule
Performance gain & +0.7 & +0.4 & +0.7 \\
\bottomrule
\end{tabular}
}
\end{table}

\noindent\textbf{Answer Masking Mitigates Confirmation Bias.} 
Finally, we investigate the role of the thinker's answer masking. Table~\ref{tab:ablation_masking} demonstrates that exposing the thinker's answer to specialized agents causes consistent degradation across benchmarks. Without masking, later agents exhibit pronounced anchoring bias toward the thinker's preliminary conclusion rather than conducting independent analysis. By masking the answer while preserving the reasoning path, we ensure agents concentrate on designated subtasks without premature influence.
\subsection{Case Study}
Figure~\ref{fig:case_study} presents an example to demonstrate the workflow of \OURMETHOD. The visualization illustrates  transparent reasoning, self-verification, confidence assessment, and attention to fine-grained details. The multi-agent pipeline systematically decomposes the problem, with each specialist contributing domain-specific expertise while maintaining interpretable intermediate outputs. Thus, providing several advantages over single-VLM approaches in terms of reasoning, robustness, and explainability. More details about executing \OURMETHOD{} on similar cases are available in Appendix~\ref{app:sec:results}.

\subsection{Error Analysis}

We analyzed 100 incorrect predictions from ORCA on the Single-Page DocVQA and InfographicsVQA validation sets. Each error was traced to the first pipeline stage responsible for the failure. The main error sources are: (1) \textbf{Reasoning errors} (43\%), where the thinker agent produces an incorrect reasoning path that misguides downstream agents; (2) \textbf{Router errors} (27\%), caused by incorrect agent selection that omits relevant document elements or activates unsuitable specialists; (3) \textbf{Agent coordination failures} (18\%), where early-stage errors propagate through sequential execution; and (4) \textbf{Over-refinement} (12\%), in which verification stages over-analyze or override initially correct answers.

A key advantage of ORCA’s modular architecture is that individual components can be upgraded independently as stronger models become available, enabling continuous system improvement without architectural changes.

\section{Conclusion}
We present ORCA, a multi-agent framework for Document Visual Question Answering that integrates explicit reasoning decomposition with specialized agent collaboration to address the limitations of single-model VLMs. Our modular architecture enables component upgrades as foundation models evolve. In future, to further improve \OURMETHOD{},we plain to optimize  the router via reinforcement learning with task-specific rewards to learn agent selection beyond the current supervised labels; orchestration ordering can be learned through policy gradients with intermediate answer quality as the state representation; and the debate mechanism can be refined via multi-agent PPO with adversarial rewards. We also plan to extend \OURMETHOD{} to multi-page document undersranding, introducing long-context routing and inter-page agentic reasoning.


\section*{Acknowledgements}
The contribution of Mohamed Ali Souibgui to this work has been supported by Consolidated Research Group 2021 SGR 01559, by project PID2023-146426NB-100 funded by MCIU/AEI/10.13039/501100011033 and FSE+, and by the European Union’s Horizon Europe programme under grant agreement No 101070617 (Project ELSA)and No 101214398 (Project ELLIOT).
{\small
\bibliographystyle{ieeenat_fullname}
\bibliography{main}

\begin{thebibliography}{74}
\providecommand{\natexlab}[1]{#1}
\providecommand{\url}[1]{\texttt{#1}}
\expandafter\ifx\csname urlstyle\endcsname\relax
  \providecommand{\doi}[1]{doi: #1}\else
  \providecommand{\doi}{doi: \begingroup \urlstyle{rm}\Url}\fi

\bibitem[Achiam et~al.(2023)Achiam, Adler, Agarwal, Ahmad, Akkaya, Aleman, Almeida, Altenschmidt, Altman, Anadkat, et~al.]{achiam2023gpt}
Josh Achiam, Steven Adler, Sandhini Agarwal, Lama Ahmad, Ilge Akkaya, Florencia~Leoni Aleman, Diogo Almeida, Janko Altenschmidt, Sam Altman, Shyamal Anadkat, et~al.
\newblock Gpt-4 technical report.
\newblock \emph{arXiv preprint arXiv:2303.08774}, 2023.

\bibitem[Agrawal et~al.(2024)Agrawal, Antoniak, Hanna, Bout, Chaplot, Chudnovsky, Costa, De~Monicault, Garg, Gervet, et~al.]{agrawal2024pixtral}
Pravesh Agrawal, Szymon Antoniak, Emma~Bou Hanna, Baudouin Bout, Devendra Chaplot, Jessica Chudnovsky, Diogo Costa, Baudouin De~Monicault, Saurabh Garg, Theophile Gervet, et~al.
\newblock Pixtral 12b.
\newblock \emph{arXiv preprint arXiv:2410.07073}, 2024.

\bibitem[Anthropic(2025)]{claude2024anthropic}
Anthropic.
\newblock Claude.
\newblock \url{https://www.anthropic.com/claude}, 2025.
\newblock Accessed: 11 November 2025.

\bibitem[Bai et~al.(2023)Bai, Bai, Yang, Wang, Tan, Wang, Lin, Zhou, and Zhou]{wang2024qwen2}
Jinze Bai, Shuai Bai, Shusheng Yang, Shijie Wang, Sinan Tan, Peng Wang, Junyang Lin, Chang Zhou, and Jingren Zhou.
\newblock Qwen-vl: A versatile vision-language model for understanding, localization, text reading, and beyond.
\newblock \emph{arXiv preprint arXiv:2308.12966}, 2023.

\bibitem[Bai et~al.(2025)Bai, Chen, Liu, Wang, Ge, Song, Dang, Wang, Wang, Tang, et~al.]{bai2025qwen25vl}
Shuai Bai, Keqin Chen, Xuejing Liu, Jialin Wang, Wenbin Ge, Sibo Song, Kai Dang, Peng Wang, Shijie Wang, Jun Tang, et~al.
\newblock Qwen2.5-vl technical report.
\newblock \emph{arXiv preprint arXiv:2502.13923}, 2025.

\bibitem[Banerjee and Lavie(2005)]{banerjee2005meteor}
Satanjeev Banerjee and Alon Lavie.
\newblock Meteor: An automatic metric for mt evaluation with improved correlation with human judgments.
\newblock In \emph{Proceedings of the ACL Workshop on Intrinsic and Extrinsic Evaluation Measures for Machine Translation and/or Summarization}, pages 65--72, 2005.

\bibitem[Banerjee et~al.(2025)Banerjee, Agarwal, and Singla]{banerjee2025llms}
Sourav Banerjee, Ayushi Agarwal, and Saloni Singla.
\newblock Llms will always hallucinate, and we need to live with this.
\newblock In \emph{Intelligent Systems Conference}, pages 624--648. Springer, 2025.

\bibitem[Biten et~al.(2019)Biten, Tito, Mafla, Gomez, Rusi{\~n}ol, Valveny, Jawahar, and Karatzas]{biten2019icdar}
Ali~Furkan Biten, Rub{\`e}n Tito, Andr{\'e}s Mafla, Lluis Gomez, Mar{\c{c}}al Rusi{\~n}ol, Ernest Valveny, CV Jawahar, and Dimosthenis Karatzas.
\newblock Scene text visual question answering.
\newblock In \emph{2019 International Conference on Document Analysis and Recognition (ICDAR)}, pages 1291--1296. IEEE, 2019.

\bibitem[Chan et~al.(2023)Chan, Chen, Su, Yu, Xue, Zhang, Fu, and Liu]{chan2023chateval}
Chi-Min Chan, Weize Chen, Yusheng Su, Jianxuan Yu, Wei Xue, Shanghang Zhang, Jie Fu, and Zhiyuan Liu.
\newblock Chateval: Towards better llm-based evaluators through multi-agent debate.
\newblock \emph{arXiv preprint arXiv:2308.07201}, 2023.

\bibitem[Chen et~al.(2024{\natexlab{a}})Chen, Zhang, Xu, and Zhao]{chen2024mllm}
Jiaqi Chen, Zeyu Zhang, Chengcheng Xu, and Zhou Zhao.
\newblock Multimodal large language models: A survey.
\newblock \emph{arXiv preprint arXiv:2405.07538}, 2024{\natexlab{a}}.

\bibitem[Chen et~al.(2023)Chen, Wu, Wang, Su, Chen, Xing, Zhong, Zhang, Zhu, Lu, et~al.]{chen2023internvl}
Zhe Chen, Jiannan Wu, Wenhai Wang, Weijie Su, Guo Chen, Sen Xing, Muyan Zhong, Qinglong Zhang, Xizhou Zhu, Lewei Lu, et~al.
\newblock Internvl: Scaling up vision foundation models and aligning for generic visual-linguistic tasks.
\newblock \emph{arXiv preprint arXiv:2312.14238}, 2023.

\bibitem[Chen et~al.(2024{\natexlab{b}})Chen, Wu, Wang, Su, Chen, Xing, Zhong, Zhang, Zhu, Lu, et~al.]{chen2024internvl}
Zhe Chen, Jiannan Wu, Wenhai Wang, Weijie Su, Guo Chen, Sen Xing, Muyan Zhong, Qinglong Zhang, Xizhou Zhu, Lewei Lu, et~al.
\newblock Internvl: Scaling up vision foundation models and aligning for generic visual-linguistic tasks.
\newblock In \emph{Proceedings of the IEEE/CVF conference on computer vision and pattern recognition}, pages 24185--24198, 2024{\natexlab{b}}.

\bibitem[Dai et~al.(2023)Dai, Li, Li, Tiong, Zhao, Wang, Li, Fung, and Hoi]{dai2023instructblip}
Wenliang Dai, Junnan Li, Dongxu Li, Anthony Meng~Huat Tiong, Junqi Zhao, Weisheng Wang, Boyang Li, Pascale Fung, and Steven Hoi.
\newblock Instructblip: Towards general-purpose vision-language models with instruction tuning.
\newblock In \emph{Advances in Neural Information Processing Systems}, pages 49250--49267, 2023.

\bibitem[{DeepSeek-AI}(2025)]{deepseek2024r1}
{DeepSeek-AI}.
\newblock Deepseek-r1: Incentivizing reasoning capability in llms via reinforcement learning.
\newblock \emph{arXiv preprint arXiv:2501.12948}, 2025.

\bibitem[Deitke et~al.(2024)Deitke, Clark, Lee, Tripathi, Yang, Park, Salehi, Muennighoff, Lo, Soldaini, et~al.]{deitke2024molmo}
Matt Deitke, Christopher Clark, Sangho Lee, Rohun Tripathi, Yue Yang, Jae~Sung Park, Mohammadreza Salehi, Niklas Muennighoff, Kyle Lo, Luca Soldaini, et~al.
\newblock Molmo and pixmo: Open weights and open data for state-of-the-art multimodal models.
\newblock \emph{arXiv e-prints}, pages arXiv--2409, 2024.

\bibitem[Du et~al.(2023)Du, Li, Torralba, Tenenbaum, and Mordatch]{du2023improving}
Yilun Du, Shuang Li, Antonio Torralba, Joshua~B Tenenbaum, and Igor Mordatch.
\newblock Improving factuality and reasoning in language models through multiagent debate.
\newblock In \emph{International Conference on Machine Learning}, pages 8633--8656. PMLR, 2023.

\bibitem[Fu et~al.(2025)Fu, Kuang, Song, Huang, Yang, Li, Zhu, Luo, Wang, Lu, Li, Tang, Shan, Lin, Liu, Wu, Feng, Liu, Huang, Tang, Chen, Jin, Liu, and Bai]{liu2024ocrbench}
Ling Fu, Zhebin Kuang, Jiajun Song, Mingxin Huang, Biao Yang, Yuzhe Li, Linghao Zhu, Qidi Luo, Xinyu Wang, Hao Lu, Zhang Li, Guozhi Tang, Bin Shan, Chunhui Lin, Qi Liu, Binghong Wu, Hao Feng, Hao Liu, Can Huang, Jingqun Tang, Wei Chen, Lianwen Jin, Yuliang Liu, and Xiang Bai.
\newblock Ocrbench v2: An improved benchmark for evaluating large multimodal models on visual text localization and reasoning.
\newblock \emph{arXiv preprint arXiv:2501.00321}, 2025.
\newblock Version 2, revised 5 Jun 2025.

\bibitem[Guo et~al.(2025)Guo, Wu, Zhu, Leng, Shi, Chen, Fan, Wang, Jiang, Wang, et~al.]{guo2025seed1}
Dong Guo, Faming Wu, Feida Zhu, Fuxing Leng, Guang Shi, Haobin Chen, Haoqi Fan, Jian Wang, Jianyu Jiang, Jiawei Wang, et~al.
\newblock Seed1.5-vl technical report.
\newblock \emph{arXiv preprint arXiv:2505.07062}, 2025.

\bibitem[Hong et~al.(2024)Hong, Zheng, Chen, Cheng, Zhang, Wang, Yau, Lin, Zhou, Ran, et~al.]{hong2023metagpt}
Sirui Hong, Xiawu Zheng, Jonathan Chen, Yuheng Cheng, Ceyao Zhang, Zili Wang, Steven Ka~Shing Yau, Zijuan Lin, Liyang Zhou, Chenyu Ran, et~al.
\newblock Metagpt: Meta programming for a multi-agent collaborative framework.
\newblock In \emph{International Conference on Learning Representations}, 2024.

\bibitem[Hu et~al.(2024)Hu, Xu, Ye, Yan, Zhang, Zhang, Li, Zhang, Jin, Huang, et~al.]{hu2024mplug}
Anwen Hu, Haiyang Xu, Jiabo Ye, Ming Yan, Liang Zhang, Bo Zhang, Chen Li, Ji Zhang, Qin Jin, Fei Huang, et~al.
\newblock mplug-docowl 1.5: Unified structure learning for ocr-free document understanding.
\newblock \emph{arXiv preprint arXiv:2403.12895}, 2024.

\bibitem[Huang et~al.(2022)Huang, Lv, Cui, Lu, and Wei]{huang2022layoutlmv3}
Yupan Huang, Tengchao Lv, Lei Cui, Yutong Lu, and Furu Wei.
\newblock Layoutlmv3: Pre-training for document ai with unified text and image masking.
\newblock In \emph{Proceedings of the 30th ACM International Conference on Multimedia}, pages 4083--4091, 2022.

\bibitem[{InternVL Team}(2024)]{internvl3-2024}
{InternVL Team}.
\newblock Internvl-3: Scaling up vision-language models.
\newblock \url{https://internvl.github.io/}, 2024.
\newblock Accessed: 2024-12-29.

\bibitem[{InternVL Team}(2025)]{InternVL3-8B-hf}
{InternVL Team}.
\newblock Internvl3-8b-hf.
\newblock \url{https://huggingface.co/OpenGVLab/InternVL3-8B-hf}, 2025.
\newblock Accessed: 2025-01-01.

\bibitem[Irving et~al.(2018)Irving, Christiano, and Amodei]{irving2018aisafety}
Geoffrey Irving, Paul Christiano, and Dario Amodei.
\newblock Ai safety via debate.
\newblock \emph{arXiv preprint arXiv:1805.00899}, 2018.

\bibitem[Kannan et~al.(2024)Kannan, Bharadhwaj, Jain, and Jayaraman]{kannan2024smart}
Sai Kannan, Homanga Bharadhwaj, Aishwarya Jain, and Dinesh Jayaraman.
\newblock Smart: Scalable multi-agent real-time simulation via next-token prediction.
\newblock \emph{arXiv preprint arXiv:2405.15677}, 2024.

\bibitem[Kim et~al.(2022)Kim, Hong, Yim, Nam, Park, Yim, Hwang, Yun, Han, and Park]{kim2022donut}
Geewook Kim, Teakgyu Hong, Moonbin Yim, JeongYeon Nam, Jinyoung Park, Jinyeong Yim, Wonseok Hwang, Sangdoo Yun, Dongyoon Han, and Seunghyun Park.
\newblock Ocr-free document understanding transformer.
\newblock In \emph{European Conference on Computer Vision}, pages 498--517. Springer, 2022.

\bibitem[Kim et~al.(2024)Kim, Cho, Kim, Song, and Choi]{kim2024mdagents}
Yubin Kim, Chanwoo Cho, Hyewon Kim, Sik Song, and Edward Choi.
\newblock Mdagents: An adaptive collaboration of llms for medical decision-making.
\newblock \emph{arXiv preprint arXiv:2404.15155}, 2024.

\bibitem[Lamm and Keuper(2024)]{lamm2024can}
Bianca Lamm and Janis Keuper.
\newblock Can visual language models replace ocr-based visual question answering pipelines in production? a case study in retail.
\newblock \emph{arXiv preprint arXiv:2408.15626}, 2024.

\bibitem[Li et~al.(2024{\natexlab{a}})Li, Zhang, Guo, Zhang, Li, Zhang, Zhang, Zhang, Li, Liu, et~al.]{li2024llava}
Bo Li, Yuanhan Zhang, Dong Guo, Renrui Zhang, Feng Li, Hao Zhang, Kaichen Zhang, Peiyuan Zhang, Yanwei Li, Ziwei Liu, et~al.
\newblock Llava-onevision: Easy visual task transfer.
\newblock \emph{arXiv preprint arXiv:2408.03326}, 2024{\natexlab{a}}.

\bibitem[Li et~al.(2023{\natexlab{a}})Li, Hammoud, Itani, Khizbullin, and Ghanem]{li2023camel}
Guohao Li, Hasan Abed Al~Kader Hammoud, Hani Itani, Dmitrii Khizbullin, and Bernard Ghanem.
\newblock Camel: Communicative agents for ``mind'' exploration of large language model society.
\newblock In \emph{Advances in Neural Information Processing Systems}, pages 51991--52008, 2023{\natexlab{a}}.

\bibitem[Li et~al.(2023{\natexlab{b}})Li, Li, Savarese, and Hoi]{li2023blip2}
Junnan Li, Dongxu Li, Silvio Savarese, and Steven Hoi.
\newblock Blip-2: Bootstrapping language-image pre-training with frozen image encoders and large language models.
\newblock In \emph{International Conference on Machine Learning}, pages 19730--19742. PMLR, 2023{\natexlab{b}}.

\bibitem[Li et~al.(2024{\natexlab{b}})Li, Yang, Liu, Ma, Zhang, Yang, Sun, Liu, and Bai]{li2024monkey}
Zhang Li, Biao Yang, Qiang Liu, Zhiyin Ma, Shuo Zhang, Jingxu Yang, Yabo Sun, Yuliang Liu, and Xiang Bai.
\newblock Monkey: Image resolution and text label are important things for large multi-modal models.
\newblock In \emph{Proceedings of the IEEE/CVF Conference on Computer Vision and Pattern Recognition}, pages 26763--26773, 2024{\natexlab{b}}.

\bibitem[Li et~al.(2025)Li, Wang, Zhang, and Chen]{li2025metal}
Zhiwei Li, Xiaoqiang Wang, Shuo Zhang, and Yong Chen.
\newblock Metal: Towards multi-agent large language models for legal case retrieval.
\newblock \emph{arXiv preprint arXiv:2501.09234}, 2025.

\bibitem[Liang et~al.(2023)Liang, He, Jiao, Wang, Wang, Wang, Yang, Tu, and Shi]{liang2023encouraging}
Tian Liang, Zhiwei He, Wenxiang Jiao, Xing Wang, Yan Wang, Rui Wang, Yujiu Yang, Zhaopeng Tu, and Shuming Shi.
\newblock Encouraging divergent thinking in large language models through multi-agent debate.
\newblock In \emph{Proceedings of the 2023 Conference on Empirical Methods in Natural Language Processing}, pages 9006--9021, 2023.

\bibitem[Liu et~al.(2023{\natexlab{a}})Liu, Li, Li, and Lee]{liu2023improved}
Haotian Liu, Chunyuan Li, Yuheng Li, and Yong~Jae Lee.
\newblock Improved baselines with visual instruction tuning.
\newblock In \emph{Proceedings of the IEEE/CVF Conference on Computer Vision and Pattern Recognition}, pages 26296--26306, 2023{\natexlab{a}}.

\bibitem[Liu et~al.(2023{\natexlab{b}})Liu, Li, Wu, and Lee]{liu2023visual}
Haotian Liu, Chunyuan Li, Qingyang Wu, and Yong~Jae Lee.
\newblock Visual instruction tuning.
\newblock In \emph{Advances in Neural Information Processing Systems}, pages 34892--34916, 2023{\natexlab{b}}.

\bibitem[Liu et~al.(2024)Liu, Li, Li, Li, Zhang, Shen, and Lee]{liu2024llava-next}
Haotian Liu, Chunyuan Li, Yuheng Li, Bo Li, Yuanhan Zhang, Sheng Shen, and Yong~Jae Lee.
\newblock Llava-next: Improved reasoning, ocr, and world knowledge, 2024.

\bibitem[Luo et~al.(2024)Luo, Sibue, Chen, Tang, Huang, and Feng]{luo2024layoutllm}
Masato Luo, Brian Sibue, Xin Chen, Huaxiu Tang, Simeng Huang, and Yiheng Feng.
\newblock Layoutllm: Large language model instruction tuning for visually-rich document understanding.
\newblock \emph{arXiv preprint arXiv:2403.05252}, 2024.

\bibitem[Mathew et~al.(2021)Mathew, Karatzas, and Jawahar]{mathew2021docvqa}
Minesh Mathew, Dimosthenis Karatzas, and C.~V. Jawahar.
\newblock Docvqa: A dataset for vqa on document images.
\newblock In \emph{Proceedings of the IEEE/CVF Winter Conference on Applications of Computer Vision (WACV)}, pages 2200--2209, 2021.

\bibitem[Mathew et~al.(2022)Mathew, Karatzas, and Jawahar]{mathew2022infographicvqa}
Minesh Mathew, Dimosthenis Karatzas, and CV Jawahar.
\newblock Infographicvqa.
\newblock In \emph{Proceedings of the IEEE/CVF Winter Conference on Applications of Computer Vision}, pages 1697--1706, 2022.

\bibitem[{NVIDIA}(2024{\natexlab{a}})]{nvidia2024llama-nemotron}
{NVIDIA}.
\newblock Llama nemotron vl 8b.
\newblock \url{https://developer.nvidia.com/}, 2024{\natexlab{a}}.
\newblock Accessed: 2024-12-29.

\bibitem[{NVIDIA}(2024{\natexlab{b}})]{nvidia2024nemotron-nano}
{NVIDIA}.
\newblock Nemotron nano v2 vl.
\newblock \url{https://developer.nvidia.com/}, 2024{\natexlab{b}}.
\newblock Accessed: 2024-12-29.

\bibitem[{OpenAI}(2024)]{openai2024o1}
{OpenAI}.
\newblock Learning to reason with llms.
\newblock Technical report, OpenAI, 2024.
\newblock Available at: \url{https://openai.com/index/learning-to-reason-with-llms/}.

\bibitem[Papineni et~al.(2002)Papineni, Roukos, Ward, and Zhu]{papineni2002bleu}
Kishore Papineni, Salim Roukos, Todd Ward, and Wei-Jing Zhu.
\newblock Bleu: a method for automatic evaluation of machine translation.
\newblock In \emph{Proceedings of Annual Meeting of the Association for Computational Linguistics}, pages 311--318, 2002.

\bibitem[Powalski et~al.(2021)Powalski, Borchmann, Jurkiewicz, Dwojak, Pietruszka, and Pa{\l}ka]{powalski2021going}
Rafa{\l} Powalski, {\L}ukasz Borchmann, Dawid Jurkiewicz, Tomasz Dwojak, Micha{\l} Pietruszka, and Gabriela Pa{\l}ka.
\newblock Going full-tilt boogie on document understanding with text-image-layout transformer.
\newblock In \emph{International Conference on Document Analysis and Recognition}, pages 732--747. Springer, 2021.

\bibitem[{Qwen Team}(2024)]{qwen2024qwen3-omni}
{Qwen Team}.
\newblock Qwen3-omni: Multimodal large language model.
\newblock \url{https://qwenlm.github.io/}, 2024.
\newblock Accessed: 2024-12-29.

\bibitem[{Qwen Team}(2025{\natexlab{a}})]{qwen2025qwen2.5-vl-7b}
{Qwen Team}.
\newblock Qwen2.5-vl-7b-instruct.
\newblock \url{https://huggingface.co/Qwen/Qwen2.5-VL-7B-Instruct}, 2025{\natexlab{a}}.
\newblock Accessed: 2025-01-01.

\bibitem[{Qwen Team}(2025{\natexlab{b}})]{qwen2025qwen3-1.7b}
{Qwen Team}.
\newblock Qwen3-1.7b.
\newblock \url{https://huggingface.co/Qwen/Qwen3-1.7B}, 2025{\natexlab{b}}.
\newblock Accessed: 2025-01-01.

\bibitem[{Qwen Team}(2025{\natexlab{c}})]{qwen2025qwen3-vl-4b}
{Qwen Team}.
\newblock Qwen3-vl-4b-instruct.
\newblock \url{https://huggingface.co/Qwen/Qwen3-VL-4B-Instruct}, 2025{\natexlab{c}}.
\newblock Accessed: 2025-01-01.

\bibitem[{Qwen Team}(2025{\natexlab{d}})]{qwen2025qwen3-vl-8b}
{Qwen Team}.
\newblock Qwen3-vl-8b-instruct.
\newblock \url{https://huggingface.co/Qwen/Qwen3-VL-8B-Instruct}, 2025{\natexlab{d}}.
\newblock Accessed: 2025-01-01.

\bibitem[{Qwen Team}(2025{\natexlab{e}})]{qwen3vl2025}
{Qwen Team}.
\newblock Qwen3-vl: The most powerful vision-language model in the qwen series.
\newblock \url{https://github.com/QwenLM/Qwen3-VL}, 2025{\natexlab{e}}.
\newblock Accessed: 2025-11-08.

\bibitem[Shen et~al.(2023)Shen, Song, Tan, Li, Lu, and Zhuang]{shen2023hugginggpt}
Yongliang Shen, Kaitao Song, Xu Tan, Dongsheng Li, Weiming Lu, and Yueting Zhuang.
\newblock Hugginggpt: Solving ai tasks with chatgpt and its friends in hugging face.
\newblock \emph{Advances in Neural Information Processing Systems}, 36:\penalty0 40075--40093, 2023.

\bibitem[Shinn et~al.(2023)Shinn, Cassano, Gopinath, Narasimhan, and Yao]{shinn2023reflexion}
Noah Shinn, Federico Cassano, Ashwin Gopinath, Karthik Narasimhan, and Shunyu Yao.
\newblock Reflexion: Language agents with verbal reinforcement learning.
\newblock \emph{Advances in Neural Information Processing Systems}, 36:\penalty0 8634--8652, 2023.

\bibitem[Su et~al.(2020)Su, Zhong, Zhang, Chen, Li, and Lin]{su2020adapting}
Tong Su, Yifei Zhong, Kai Zhang, Qiyuan Chen, Xiang Li, and Stephen Lin.
\newblock Adapting document images to display constraints.
\newblock In \emph{Proceedings of the 28th ACM International Conference on Multimedia}, pages 1617--1625, 2020.

\bibitem[Team et~al.(2024)Team, Georgiev, Lei, Burnell, Bai, Gulati, Tanzer, Vincent, Pan, Wang, et~al.]{team2024gemini}
Gemini Team, Petko Georgiev, Ving~Ian Lei, Ryan Burnell, Libin Bai, Anmol Gulati, Garrett Tanzer, Damien Vincent, Zhufeng Pan, Shibo Wang, et~al.
\newblock Gemini 1.5: Unlocking multimodal understanding across millions of tokens of context.
\newblock \emph{arXiv preprint arXiv:2403.05530}, 2024.

\bibitem[Team et~al.(2025{\natexlab{a}})Team, Kamath, Ferret, Pathak, Vieillard, Merhej, Perrin, Matejovicova, Ram{\'e}, Rivi{\`e}re, et~al.]{team2025gemma}
Gemma Team, Aishwarya Kamath, Johan Ferret, Shreya Pathak, Nino Vieillard, Ramona Merhej, Sarah Perrin, Tatiana Matejovicova, Alexandre Ram{\'e}, Morgane Rivi{\`e}re, et~al.
\newblock Gemma 3 technical report.
\newblock \emph{arXiv preprint arXiv:2503.19786}, 2025{\natexlab{a}}.

\bibitem[Team et~al.(2025{\natexlab{b}})Team, Hong, Yu, Gu, Wang, Gan, Tang, Cheng, Qi, Ji, Pan, Duan, Wang, Wang, Cheng, He, Su, Yang, Pan, Zeng, Wang, Chen, Shi, Pang, Zhang, Yin, Yang, Chen, Xu, Zhu, Chen, Chen, Chen, Lin, Wang, Chen, Lei, Gong, Pan, Liu, Xu, Zhang, Zheng, Yang, Zhong, Huang, Zhao, Xue, Tu, Meng, Zhang, Luo, Hao, Tong, Li, Jia, Liu, Zhang, Lyu, Fan, Huang, Wang, Xue, Wang, Wang, An, Du, Shi, Huang, Niu, Wang, Yue, Li, Zhang, Wang, Wang, Zhang, Xue, Hou, Du, Wang, Zhang, Liu, Xu, Li, Huang, Dong, and Tang]{zeng2024glm4v}
V Team, Wenyi Hong, Wenmeng Yu, Xiaotao Gu, Guo Wang, Guobing Gan, Haomiao Tang, Jiale Cheng, Ji Qi, Junhui Ji, Lihang Pan, Shuaiqi Duan, Weihan Wang, Yan Wang, Yean Cheng, Zehai He, Zhe Su, Zhen Yang, Ziyang Pan, Aohan Zeng, Baoxu Wang, Bin Chen, Boyan Shi, Changyu Pang, Chenhui Zhang, Da Yin, Fan Yang, Guoqing Chen, Jiazheng Xu, Jiale Zhu, Jiali Chen, Jing Chen, Jinhao Chen, Jinghao Lin, Jinjiang Wang, Junjie Chen, Leqi Lei, Letian Gong, Leyi Pan, Mingdao Liu, Mingde Xu, Mingzhi Zhang, Qinkai Zheng, Sheng Yang, Shi Zhong, Shiyu Huang, Shuyuan Zhao, Siyan Xue, Shangqin Tu, Shengbiao Meng, Tianshu Zhang, Tianwei Luo, Tianxiang Hao, Tianyu Tong, Wenkai Li, Wei Jia, Xiao Liu, Xiaohan Zhang, Xin Lyu, Xinyue Fan, Xuancheng Huang, Yanling Wang, Yadong Xue, Yanfeng Wang, Yanzi Wang, Yifan An, Yifan Du, Yiming Shi, Yiheng Huang, Yilin Niu, Yuan Wang, Yuanchang Yue, Yuchen Li, Yutao Zhang, Yuting Wang, Yu Wang, Yuxuan Zhang, Zhao Xue, Zhenyu Hou, Zhengxiao Du, Zihan Wang, Peng Zhang, Debing Liu, Bin Xu, Juanzi Li,
  Minlie Huang, Yuxiao Dong, and Jie Tang.
\newblock Glm-4.5v and glm-4.1v-thinking: Towards versatile multimodal reasoning with scalable reinforcement learning, 2025{\natexlab{b}}.

\bibitem[Tito et~al.(2023)Tito, Karatzas, and Valveny]{tito2023hierarchical}
Rub{\`e}n Tito, Dimosthenis Karatzas, and Ernest Valveny.
\newblock Hierarchical multimodal transformers for visual question answering.
\newblock In \emph{Proceedings of the IEEE/CVF Conference on Computer Vision and Pattern Recognition}, pages 10495--10504, 2023.

\bibitem[Tong et~al.(2024)Tong, Brown~II, Wu, Woo, IYER, Akula, Yang, Yang, Middepogu, Wang, et~al.]{tong2024cambrian}
Shengbang Tong, Ellis~L Brown~II, Penghao Wu, Sanghyun Woo, Arjun~Jyoti IYER, Sai~Charith Akula, Shusheng Yang, Jihan Yang, Megha Middepogu, Zichen Wang, et~al.
\newblock Cambrian-1: A fully open, vision-centric exploration of multimodal llms.
\newblock In \emph{Advances in Neural Information Processing Systems}, 2024.

\bibitem[Wang et~al.(2023)Wang, Wei, Schuurmans, Le, Chi, Narang, Chowdhery, and Zhou]{wang2022self}
Xuezhi Wang, Jason Wei, Dale Schuurmans, Quoc Le, Ed Chi, Sharan Narang, Aakanksha Chowdhery, and Denny Zhou.
\newblock Self-consistency improves chain of thought reasoning in language models.
\newblock In \emph{International Conference on Learning Representations}, 2023.

\bibitem[Wei et~al.(2022)Wei, Wang, Schuurmans, Bosma, Ichter, Xia, Chi, Le, and Zhou]{wei2022chain}
Jason Wei, Xuezhi Wang, Dale Schuurmans, Maarten Bosma, Brian Ichter, Fei Xia, Ed Chi, Quoc Le, and Denny Zhou.
\newblock Chain-of-thought prompting elicits reasoning in large language models.
\newblock \emph{arXiv preprint arXiv:2201.11903}, 2022.
\newblock Version 6, revised 10 Jan 2023.

\bibitem[Wu et~al.(2023{\natexlab{a}})Wu, Yin, Qi, Wang, Tang, and Duan]{wu2023visualchatgpt}
Chenfei Wu, Shengming Yin, Weizhen Qi, Xiaodong Wang, Zecheng Tang, and Nan Duan.
\newblock Visual chatgpt: Talking, drawing and editing with visual foundation models.
\newblock \emph{arXiv preprint arXiv:2303.04671}, 2023{\natexlab{a}}.

\bibitem[Wu et~al.(2023{\natexlab{b}})Wu, Bansal, Zhang, Wu, Li, Zhu, Jiang, Zhang, Zhang, Liu, et~al.]{wu2023autogen}
Qingyun Wu, Gagan Bansal, Jieyu Zhang, Yiran Wu, Beibin Li, Erkang Zhu, Li Jiang, Xiaoyun Zhang, Shaokun Zhang, Jiale Liu, et~al.
\newblock Autogen: Enabling next-gen llm applications via multi-agent conversation.
\newblock \emph{arXiv preprint arXiv:2308.08155}, 2023{\natexlab{b}}.

\bibitem[Wu et~al.(2024)Wu, Chen, Pan, Liu, Liu, Dai, Gao, Ma, Wu, Wang, Xie, Wu, Hu, Wang, Sun, Li, Piao, Guan, Liu, Xie, You, Dong, Yu, Zhang, Zhao, Wang, and Ruan]{wu2024deepseekvl2mixtureofexpertsvisionlanguagemodels}
Zhiyu Wu, Xiaokang Chen, Zizheng Pan, Xingchao Liu, Wen Liu, Damai Dai, Huazuo Gao, Yiyang Ma, Chengyue Wu, Bingxuan Wang, Zhenda Xie, Yu Wu, Kai Hu, Jiawei Wang, Yaofeng Sun, Yukun Li, Yishi Piao, Kang Guan, Aixin Liu, Xin Xie, Yuxiang You, Kai Dong, Xingkai Yu, Haowei Zhang, Liang Zhao, Yisong Wang, and Chong Ruan.
\newblock Deepseek-vl2: Mixture-of-experts vision-language models for advanced multimodal understanding, 2024.

\bibitem[{Xiaomi LLM-Core Team}(2025)]{coreteam2025mimovl}
{Xiaomi LLM-Core Team}.
\newblock Mimo-vl technical report.
\newblock \url{https://github.com/XiaomiMiMo/MiMo-VL}, 2025.

\bibitem[Xu et~al.(2025)Xu, Jin, Wu, Li, Song, Sun, and Yuan]{xu2025llava}
Guowei Xu, Peng Jin, Ziang Wu, Hao Li, Yibing Song, Lichao Sun, and Li Yuan.
\newblock Llava-cot: Let vision language models reason step-by-step.
\newblock In \emph{Proceedings of the IEEE/CVF International Conference on Computer Vision}, pages 2087--2098, 2025.

\bibitem[Xu et~al.(2020)Xu, Li, Cui, Huang, Wei, and Zhou]{xu2020layoutlm}
Yiheng Xu, Minghao Li, Lei Cui, Shaohan Huang, Furu Wei, and Ming Zhou.
\newblock Layoutlm: Pre-training of text and layout for document image understanding.
\newblock In \emph{Proceedings of the 26th ACM SIGKDD International Conference on Knowledge Discovery \& Data Mining}, pages 1192--1200, 2020.

\bibitem[Xu et~al.(2021)Xu, Xu, Lv, Cui, Wei, Wang, Lu, Florencio, Zhang, Che, et~al.]{xu2021layoutlmv2}
Yang Xu, Yiheng Xu, Tengchao Lv, Lei Cui, Furu Wei, Guoxin Wang, Yijuan Lu, Dinei Florencio, Cha Zhang, Wanxiang Che, et~al.
\newblock Layoutlmv2: Multi-modal pre-training for visually-rich document understanding.
\newblock In \emph{Proceedings of the 59th Annual Meeting of the Association for Computational Linguistics and the 11th International Joint Conference on Natural Language Processing (Volume 1: Long Papers)}, pages 2579--2591, 2021.

\bibitem[Yao et~al.(2023)Yao, Zhao, Yu, Du, Shafran, Narasimhan, and Cao]{yao2022react}
Shunyu Yao, Jeffrey Zhao, Dian Yu, Nan Du, Izhak Shafran, Karthik Narasimhan, and Yuan Cao.
\newblock React: Synergizing reasoning and acting in language models.
\newblock In \emph{International Conference on Learning Representations}, 2023.

\bibitem[Zhang et~al.(2025)Zhang, Li, Cheng, Hu, Yuan, Chen, Leng, Jiang, Zhang, Li, et~al.]{zhang2025videollama}
Boqiang Zhang, Kehan Li, Zesen Cheng, Zhiqiang Hu, Yuqian Yuan, Guanzheng Chen, Sicong Leng, Yuming Jiang, Hang Zhang, Xin Li, et~al.
\newblock Videollama 3: Frontier multimodal foundation models for image and video understanding.
\newblock \emph{arXiv preprint arXiv:2501.13106}, 2025.

\bibitem[Zhang et~al.(2024{\natexlab{a}})Zhang, Fan, and Zhang]{zhang2024docassistant}
Jinxu Zhang, Qiyuan Fan, and Yu Zhang.
\newblock Read and think: An efficient step-wise multimodal language model for document understanding and reasoning.
\newblock \emph{arXiv preprint arXiv:2403.00816}, 2024{\natexlab{a}}.

\bibitem[Zhang et~al.(2024{\natexlab{b}})Zhang, Zhang, Li, Zhang, Sun, Gan, Yang, Pang, and Yang]{zhang2024improvecot}
Ruohong Zhang, Bowen Zhang, Yanghao Li, Haotian Zhang, Zhiqing Sun, Zhe Gan, Yinfei Yang, Ruoming Pang, and Yiming Yang.
\newblock Improve vision language model chain-of-thought reasoning.
\newblock \emph{arXiv preprint arXiv:2410.16198}, 2024{\natexlab{b}}.

\bibitem[Zhong et~al.(2020)Zhong, ShafieiBavani, and Jimeno~Yepes]{teds}
Xu Zhong, Elaheh ShafieiBavani, and Antonio Jimeno~Yepes.
\newblock Image-based table recognition: data, model, and evaluation.
\newblock In \emph{Proceedings of European Conference on Computer Vision}, pages 564--580. Springer, 2020.

\bibitem[Zhu et~al.(2025)Zhu, Wang, Chen, Liu, Ye, Gu, Tian, Duan, Su, Shao, Gao, Cui, Wang, Cao, Liu, Wei, Zhang, Wang, Xu, Li, Wang, Deng, Li, He, Jiang, Luo, Wang, He, Shi, Zhang, Shao, He, Xiong, Qu, Sun, Jiao, Lv, Wu, Zhang, Deng, Ge, Chen, Wang, Dou, Lu, Zhu, Lu, Lin, Qiao, Dai, and Wang]{zhu2025internvl3}
Jinguo Zhu, Weiyun Wang, Zhe Chen, Zhaoyang Liu, Shenglong Ye, Lixin Gu, Hao Tian, Yuchen Duan, Weijie Su, Jie Shao, Zhangwei Gao, Erfei Cui, Xuehui Wang, Yue Cao, Yangzhou Liu, Xingguang Wei, Hongjie Zhang, Haomin Wang, Weiye Xu, Hao Li, Jiahao Wang, Nianchen Deng, Songze Li, Yinan He, Tan Jiang, Jiapeng Luo, Yi Wang, Conghui He, Botian Shi, Xingcheng Zhang, Wenqi Shao, Junjun He, Yingtong Xiong, Wenwen Qu, Peng Sun, Penglong Jiao, Han Lv, Lijun Wu, Kaipeng Zhang, Huipeng Deng, Jiaye Ge, Kai Chen, Limin Wang, Min Dou, Lewei Lu, Xizhou Zhu, Tong Lu, Dahua Lin, Yu Qiao, Jifeng Dai, and Wenhai Wang.
\newblock Internvl3: Exploring advanced training and test-time recipes for open-source multimodal models.
\newblock \emph{arXiv preprint arXiv:2504.10479}, 2025.

\end{thebibliography}
}

\ifarxiv \clearpage \appendix  
\clearpage
\setcounter{page}{1}
\maketitlesupplementary

\section{Implementation and Training Details}\label{appendix:impl_details}

\subsection{Datasets.} We evaluate our approach on three challenging document understanding benchmarks: 
(1) \textbf{Single-Page DocVQA}~\cite{mathew2021docvqa}: a standard benchmark for single-page document question answering covering diverse document types; 
(2) \textbf{InfographicsVQA}~\cite{mathew2022infographicvqa}: a dataset requiring integration of textual and visual cues to answer questions about infographics; and 
(3) \textbf{OCRBench-v2 (en)}~\cite{liu2024ocrbench}: a comprehensive benchmark for OCR and document understanding. 
These datasets span a variety of document structures, including forms, tables, charts, and mixed content, ensuring a comprehensive evaluation.

\subsection{Baselines.} We compare our method against state-of-the-art vision-language models (VLMs), including \textbf{Qwen2.5-VL-7B-Instruct}~\cite{bai2025qwen25vl}, \textbf{Qwen3VL-4B-Instruct}, and \textbf{Qwen3VL-8B-Instruct}~\cite{qwen3vl2025}. 
These baselines represent single-model systems without multi-agent collaboration or explicit reasoning decomposition.

\subsection{Evaluation Metrics.} 
Following standard evaluation protocols~\cite{mathew2021docvqa,mathew2022infographicvqa}, we report \textbf{ANLS} scores for \textbf{Single-Page DocVQA} and \textbf{InfographicsVQA}. 
For \textbf{OCRBench-v2}, we employ its official multi-dimensional evaluation suite with six task-specific metrics. The final score represents the average across all dimensions:

\begin{itemize}[leftmargin=10pt]
    \item \textbf{Parsing:} TEDS~\cite{teds} for structural similarity in format conversion.
    \item \textbf{Localization:} IoU for spatial overlap of text regions.
    \item \textbf{Extraction:} F1 score for relation and key information extraction.
    \item \textbf{Long Reading:} BLEU~\cite{papineni2002bleu}, METEOR~\cite{banerjee2005meteor}, F1, and edit distance for long-form comprehension.
    \item \textbf{Counting:} Normalized L1 distance for text instance enumeration.
    \item \textbf{Basic VQA:} Exact match for multiple-choice; substring matching ($\leq 5$ words) or ANLS for open-ended questions.
\end{itemize}

\section{Router}
In this Section, we provide further details on the router agent used in \OURMETHOD{}.
\subsection{Training Data and Augmentation}\label{supp:router}
We train the router on the Single-Page Document VQA dataset with ground-truth agent annotations. To enhance model robustness and generalization, we apply some data augmentation techniques:

\begin{itemize}
    \item \textbf{Back-translation}: Questions are translated through intermediate languages (French and Chinese) and then back to English, generating paraphrased variants while preserving semantic meaning

    \item \textbf{Document perturbations}: Minor transformations to document images (rotation, contrast adjustment) simulate real-world scanning variations

\end{itemize}

To ensure robust evaluation and prevent data leakage in the multi-label setting, we employ Multilabel Stratified K-Fold cross-validation with $n_{\text{splits}} = 8$. This stratification strategy preserves the distribution of label combinations across folds, which is critical given that some agent combinations are significantly rarer than others. We train the router on seven folds and validate on the remaining fold.

\subsection{Model Architecture and Optimization}
We employ Qwen2.5-VL-7B as the base architecture for $A_{\text{route}}$, fine-tuned on our augmented dataset. To optimize training efficiency for our English-focused benchmark evaluation, we apply several key techniques:
\begin{itemize}
    
\item \textbf{Vocabulary Shrinking.} We reduce the tokenizer vocabulary by identifying and removing tokens unused in our training corpus. This process:
\begin{itemize}
    \item Analyzes the actual token distribution in our DocVQA datasets
    \item Removes unused tokens while preserving special tokens and model configuration tokens
    \item Shrinks the embedding layers accordingly
\end{itemize}

This vocabulary reduction yields substantial benefits: reduced memory footprint (enabling larger batch sizes), faster training convergence, and decreased inference latency—critical for real-time routing decisions. For English-centric benchmarks, this approach typically reduces vocabulary size with no loss in task performance.

\item \textbf{Efficient Training Infrastructure.} We leverage Unsloth's optimized training framework combined with Flash Attention 2 for memory-efficient attention computation. Flash Attention 2 reduces memory complexity from $O(N^2)$ to $O(N)$ for sequence length $N$, enabling us to process high-resolution document images with longer context windows during training.

\end{itemize}
\subsection{Turbo DFS Decoding for Multi-Label Prediction}
\label{sec:turbo_dfs}

Unlike standard classification approaches that apply a sigmoid threshold to output logits, we treat routing as a constrained generation task and employ \textbf{Turbo DFS} (Depth-First Search with score-guided pruning) for decoding. This choice addresses fundamental limitations of traditional multi-label decoding:

\begin{itemize}
    \item \textit{Sampling-based methods} introduce non-determinism and may miss valid label combinations across runs
    \item \textit{Greedy decoding} returns a single sequence, potentially missing alternative valid agent combinations
    \item \textit{Beam search} explores only a fixed number of sequences without explicit probability thresholds
\end{itemize}

Turbo DFS offers several advantages for our multi-label routing task:

\noindent \textbf{Algorithm Overview.} Turbo DFS performs score-guided enumeration over token continuations, pruning branches whose cumulative negative log-likelihood exceeds a configurable threshold. Starting from the model's output logits:
\begin{enumerate}
    \item Compute token-level negative log-likelihoods (NLL) after temperature scaling: $\text{NLL}(t) = -\log P(t \mid \text{context})$
    \item For each candidate token $t$, calculate cumulative score: $s_{\text{new}} = s_{\text{prev}} + \text{NLL}(t)$
    \item Prune branches where $s_{\text{new}} > -\log(\text{min\_prob})$, with special handling for the greedy token (most probable continuation)
    \item Recursively explore unpruned branches up to \texttt{max\_new\_tokens} depth
    \item Return all valid sequences as ranked candidates with their cumulative scores
\end{enumerate}

\noindent \textbf{Deterministic Multi-Label Extraction.} Given the ranked candidate sequences from Turbo DFS, we employ a \textit{union strategy} to extract the final agent activation set:
\begin{equation}
    \mathcal{A}_{\text{active}} = \bigcup_{\substack{(s, \tau) \in \text{candidates} \\ e^{-s} \geq \text{min\_prob}}} \text{DecodeAgents}(\tau)
\end{equation}

where $s$ is the cumulative score, $\tau$ is the token sequence, and $\text{DecodeAgents}(\tau)$ maps token sequences to agent identifiers by decoding tokens and parsing agent labels from the resulting text. This union approach ensures high recall: any agent appearing in a high-probability candidate sequence is included in the final routing decision.
\noindent \textbf{Hyperparameters.} We configure Turbo DFS with:
\begin{itemize}
    \item $\text{min\_prob} = 0.02$ (accept sequences with probability $\geq 2\%$)
    \item $\text{max\_new\_tokens} = 3$ (agent labels are short)
    \item $\text{temperature} = 0.9$ (slight smoothing of probability distribution)
\end{itemize}

This decoding strategy provides deterministic, ranked agent selections with explicit confidence scores, enabling principled multi-label thresholding and supporting downstream reranking if needed.

\section{Algorithms}
\label{supp:algos}

\begin{algorithm}
\caption{Collaborative Agent Execution}
\label{alg:collaborative_execution}
\small
\begin{algorithmic}[1]
\Require Question $q$, Document $\mathcal{D}$, Reasoning path $\mathcal{R}$, Initial answer $a_T$, Agent dock $\{A_1, \ldots, A_9\}$
\Ensure Expert answer $a_E$
\State $\mathbf{v} \gets A_{\text{route}}(q, \mathcal{D}, \mathcal{R})$ \Comment{\textcolor{blue}{Activate agents}}
\State $\mathcal{A}_{\text{active}} \gets \{A_i \mid v_i = 1\}$
\State $\sigma \gets \text{Orchestrate}(\mathcal{A}_{\text{active}}, \mathcal{R}, q, \mathcal{D})$ \Comment{\textcolor{blue}{Determine order}}
\State $a_0 \gets \emptyset$ \Comment{\textcolor{blue}{Initialize}}
\For{$i = 1$ to $|\sigma| - 1$}
    \State $a_i \gets \sigma_i(q, \mathcal{D}, a_{i-1})$ \Comment{\textcolor{blue}{Sequential execution}}
\EndFor
\State $\mathcal{R}^* \gets \text{MaskAnswer}(\mathcal{R}, a_T, \tau)$ \Comment{\textcolor{blue}{Mask reasoning}}
\State $a_E \gets \sigma_n(q, \mathcal{D}, a_{n-1}, \mathcal{R}^*)$ \Comment{\textcolor{blue}{Final agent}}
\State \Return $a_E$
\end{algorithmic}
\end{algorithm}

\begin{algorithm}
\caption{Stress Testing Session}
\label{alg:debate}
\small
\begin{algorithmic}[1]
\Require Question $q$, Document $\mathcal{D}$, Expert answer $a_E$, Specialized agent $\sigma_n$
\Ensure Debate answer $a_D$, Proceed to Stage 4: $\text{flag}_{\text{comm}}$
\State $\text{flag}_{\text{comm}} \gets \text{False}$
\For{$t = 1$ to $2$} \Comment{\textcolor{blue}{Two debate turns}}
    \State $q_{\text{debate}} \gets A_{\text{debate}}(q, \mathcal{D}, a_E)$
    \State $(r_{\text{debate}}, a'_E) \gets \sigma_n(q_{\text{debate}}, q, \mathcal{D}, a_E)$
    \State $d \gets A_{\text{eval}}(q_{\text{debate}}, r_{\text{debate}}, a_E, a'_E)$
    \If{$d = \text{fail}$}
        \State $\text{flag}_{\text{comm}} \gets \text{True}$
        \State \textbf{break}
    \EndIf
\EndFor
\If{$\text{flag}_{\text{comm}} = \text{False}$}
    \State $a_D \gets a_E$ \Comment{\textcolor{blue}{Agent passed stress test}}
\Else
    \State $a_D \gets \text{None}$ \Comment{\textcolor{blue}{Proceed to Stage 4}}
\EndIf
\State \Return $a_D$, $\text{flag}_{\text{comm}}$
\end{algorithmic}
\end{algorithm}

\begin{algorithm}
\caption{Multi-turn Communication}
\label{alg:communication}
\small
\begin{algorithmic}[1]
\Require Question $q$, Document $\mathcal{D}$, Expert answer $a_E$
\Ensure Communication answer $a_C$
\State $a_{\text{alt}} \gets A_{\text{anti}}(q, \mathcal{D}, a_E)$
\If{$a_{\text{alt}} = a_E$ or $a_E \subset a_{\text{alt}}$}
    \State \Return $a_E$ \Comment{\textcolor{blue}{No alternative found}}
\EndIf
\State $\text{summary}^{(0)} \gets \emptyset$
\State $\text{transcript} \gets [\,]$
\For{$t = 1$ to $3$} \Comment{\textcolor{blue}{Three-turn debate}}
    \State $\text{arg}_{\text{anti}}^{(t)} \gets A_{\text{anti}}(q, \mathcal{D}, a_E, \text{summary}^{(t-1)})$
    \State $\text{arg}_{\text{thesis}}^{(t)} \gets A_{\text{thesis}}(q, \mathcal{D}, a_E,$
    \Statex \hspace*{\algorithmicindent}\hspace*{\algorithmicindent}$\text{arg}_{\text{anti}}^{(t)}[\text{REF, CRIT}], \text{summary}^{(t-1)})$
    \State $(\text{convinced}, \text{summary}^{(t)}) \gets A_{\text{judge}}(\text{arg}_{\text{thesis}}^{(t)}, \text{arg}_{\text{anti}}^{(t)})$
    \State $\text{transcript}.\text{append}(\text{arg}_{\text{anti}}^{(t)}, \text{arg}_{\text{thesis}}^{(t)})$
    \If{$\text{convinced} \neq \text{None}$}
        \State $a_C \gets \text{convinced.answer}$
        \State \Return $a_C$ \Comment{\textcolor{blue}{Early termination}}
    \EndIf
\EndFor
\State $a_C \gets A_{\text{judge}}.\text{FinalDecision}(\text{transcript})$ \Comment{\textcolor{blue}{Linguistic analysis}}
\State \Return $a_C$
\end{algorithmic}
\end{algorithm}

\section{Inference Latency and Cost Analysis}
\label{appendix:latency}

\subsection{Optimization Details}
Three optimizations reduce \OURMETHOD{}'s effective latency: (1) \textbf{vLLM acceleration} provides approximately 5$\times$ throughput improvement over naive Hugging Face inference via continuous batching and PagedAttention. (2) \textbf{Conditional execution} bypasses Stages 3 and 4 when the thinker and expert agents produce identical answers, occurring in 77\% of test instances. (3) \textbf{Backbone reuse} shares model weights across agents of the same architecture, reducing GPU memory overhead and eliminating redundant model initialization.

\subsection{ORCA-Lite Configuration}
For latency-sensitive scenarios, \OURMETHOD{}-Lite restricts the pipeline to Stages 1, 2 and 5 only, incurring approximately 4--7$\times$ the latency of a single-model baseline while delivering +2--3\% improvement on complex reasoning tasks.

\begin{table}[h]
\centering
\caption{\OURMETHOD{}-Lite vs.\ full pipeline accuracy--latency trade-off.}
\label{tab:orca_lite}
\resizebox{\columnwidth}{!}{
\begin{tabular}{l|c|ccc}
\toprule
\textbf{Configuration} & \textbf{Latency} & \textbf{DocVQA} & \textbf{InfoVQA} & \textbf{OCRBench-v2} \\
\midrule
Single Model                       & 0.3--0.8s           & 96.1 & 83.1 & 65.4 \\
\OURMETHOD{}-Lite    & 3.2--5.3s     & 96.8 & 87.0 & 66.4 \\
\OURMETHOD{} Full                  & 9.6--13.1s & 97.2 & 88.0 & 67.1 \\
\bottomrule
\end{tabular}
}
\end{table}

\section{Additional Results}\label{app:sec:results}

\subsection{Detailed DocVQA Performance Breakdown}

Table~\ref{tab:docvqa-detailed} presents a comprehensive performance breakdown on the DocVQA benchmark, evaluating all open-source models across different document types and question categories. The analysis covers nine distinct categories: \textbf{Figures/Diagrams}, \textbf{Forms}, \textbf{Tables/Lists}, \textbf{Layout}, \textbf{Free Text}, \textbf{Images/Photos}, \textbf{Handwriting}, \textbf{Yes/No} questions, and \textbf{Other} question types. This granular evaluation provides insights into model capabilities across diverse document understanding tasks.

Our multi-agent framework, \OURMETHOD{}, demonstrates consistent improvements across all categories when compared to baseline open-source models. Notably, \OURMETHOD{} with Qwen3VL-8B achieves the highest overall score of 97.2\%, with particularly strong performance on Yes/No questions (100\%) and Tables/Lists (97.8\%). The framework shows robust performance across challenging categories such as handwriting recognition (96.7\%) and form understanding (98.2\%), indicating its effectiveness in handling complex document layouts and varying text modalities.

\subsection{Detailed Infographics VQA Performance Breakdown}
Table~\ref{tab:infographics-detailed} provides an in-depth analysis of model performance on the Infographics VQA benchmark, which presents unique challenges due to the complex visual and textual information typical of infographics. The evaluation is structured across three dimensions: \textbf{Answer Type} (including image span, question span, multiple spans, and non-span answers), \textbf{Evidence Source} (table/list, textual, visual object, figure, and map), and \textbf{Required Operations} (comparison, arithmetic, and counting tasks). This multi-faceted categorization enables a comprehensive understanding of how models handle different aspects of infographic comprehension.

The results reveal that \OURMETHOD{} maintains strong performance across all three evaluation dimensions. With Qwen3VL-8B, our framework achieves an overall score of 88.0\%, demonstrating particularly notable capabilities in visual object recognition (94.1\%) and counting operations (91.4\%). Our approach shows consistent improvements across answer types and evidence sources. The framework excels at multi-span answers (83.1\%), a particularly challenging task requiring integration of information from multiple locations, and demonstrates solid performance on arithmetic operations (82.8\%), indicating robust reasoning capabilities. These results underscore the effectiveness of our multi-agent approach in handling the diverse and complex reasoning requirements inherent in infographic understanding.
\begin{table*}[t]
\centering
\caption{Detailed performance breakdown on DocVQA benchmark for open-source models. Results are shown across different document types and question categories.}
\label{tab:docvqa-detailed}
\scriptsize
\setlength{\tabcolsep}{3pt}
\begin{tabular}{l|cccccccccc}
\toprule
\textbf{Model} & \textbf{Fig/Diag} & \textbf{Form} & \textbf{Table/List} & \textbf{Layout} & \textbf{Free Text} & \textbf{Img/Photo} & \textbf{Handwr.} & \textbf{Yes/No} & \textbf{Others} & \textbf{Score} \\
\midrule

\multicolumn{11}{l}{\textit{Open-source models results}} \\
LayoutLMv2 LARGE & 65.7 & 89.5 & 87.7 & 87.9 & 87.1 & 72.9 & 67.3 & 55.2 & 81.0 & 86.7 \\
Qwen2-VL & 92.1 & \textbf{98.2} & 97.0 & 96.8 & \textbf{96.2} & \textbf{91.4} & 94.4 & 96.6 & 95.4 & 96.7 \\
InternVL2-Pro & 88.9 & 97.1 & 94.9 & 95.8 & 94.5 & 89.1 & 92.8 & 96.6 & 94.1 & 95.1 \\
Molmo-72B & 88.2 & 95.5 & 93.9 & 94.1 & 91.0 & 86.9 & 92.0 & 92.0 & 92.3 & 93.5 \\
DeepSeek-VL2 & 88.5 & 95.8 & 93.6 & 93.1 & 92.1 & 86.9 & 89.9 & 89.7 & 90.1 & 93.3 \\
LLaVA-One-Vision-8B & 90.0 & 96.7 & 95.3 & 95.3 & 92.7 & 85.1 & 92.1 & 93.1 & 94.4 & 94.8 \\
MiMo-VL-7B-RL & 91.6 & 97.1 & 96.6 & 93.9 & 93.4 & 86.0 & 94.6 & 95.4 & 92.9 & 95.0 \\
VideoLLaMA3-7B & 88.4 & 96.9 & 95.0 & 95.3 & 94.3 & 88.4 & 92.9 & 93.1 & 93.1 & 95.0 \\

\midrule
\multicolumn{11}{l}{\textit{\OURMETHOD{} (Multi-Agent Framework)}} \\
\OURMETHOD{} (Qwen2.5-VL-7B) & 91.8 & 97.8 & 97.2 & 96.9 & 95.2 & 91.0 & 95.8 & 96.6 & 95.4 & 96.4 \\
\OURMETHOD{} (Qwen3VL-4B) & 91.2 & 97.4 & 96.8 & 96.4 & 94.6 & 90.2 & 95.2 & 96.6 & 94.7 & 96.0 \\
\OURMETHOD{} (Qwen3VL-8B) & \textbf{93.2} & \textbf{98.2} & \textbf{97.8} & \textbf{97.6} & 95.6 & \textbf{91.4} & \textbf{96.7} & \textbf{100.0} & \textbf{96.6} & \textbf{97.2} \\
\bottomrule
\end{tabular}
\end{table*}
\begin{table*}[t]
\centering
\caption{Detailed performance breakdown on Infographics VQA benchmark for open-source models. Results are categorized by answer type, evidence source, and required operations.}
\label{tab:infographics-detailed}
\resizebox{\textwidth}{!}{
\begin{tabular}{l c cccc ccccc ccc}
\toprule
& & \multicolumn{4}{c}{\textbf{Answer type}} & \multicolumn{5}{c}{\textbf{Evidence}} & \multicolumn{3}{c}{\textbf{Operation}} \\
\cmidrule(lr){3-6} \cmidrule(lr){7-11} \cmidrule(lr){12-14}
\textbf{Method} & Score & Image span & Question span & Multiple spans & Non span & Table/List & Textual & Visual object & Figure & Map & Comparison & Arithmetic & Counting \\
\midrule
\multicolumn{14}{l}{\textit{Open-source models results}} \\
LayoutLMv2 LARGE & 28.3 & 34.3 & 27.6 & 6.4 & 11.1 & 24.5 & 38.6 & 14.4 & 26.0 & 31.1 & 19.0 & 11.3 & 11.6 \\
Qwen2-VL & 84.7 & 87.4 & 87.1 & 77.8 & 74.2 & 86.0 & 94.3 & 78.3 & 81.7 & 75.9 & 73.0 & 89.8 & 57.9 \\
InternVL2-Pro & 83.3 & 86.8 & 89.3 & 73.5 & 69.7 & 83.4 & 92.6 & 77.6 & 80.9 & 71.9 & 73.0 & 85.8 & 53.7 \\
Molmo-72B & 81.9 & 85.1 & 88.3 & 68.2 & 70.4 & 81.8 & 91.4 & 80.6 & 79.5 & 69.6 & 70.5 & 81.9 & 59.3 \\
DeepSeek-VL2 & 78.1 & 81.9 & 80.1 & 69.9 & 63.6 & 79.4 & 90.4 & 73.7 & 74.3 & 63.3 & 62.1 & 72.8 & 53.3 \\
LLaVA-One-Vision-8B & 78.4 & 82.2 & 83.1 & 64.6 & 65.0 & 79.6 & 89.9 & 70.6 & 74.7 & 62.6 & 62.7 & 75.8 & 54.5 \\
MiMo-VL-7B-RL & 88.1 & 90.1 & 89.5 & 84.5 & 80.9 & 90.1 & 93.3 & 83.7 & 85.8 & 73.0 & 82.6 & 89.1 & 74.4 \\
VideoLLaMA3-7B & 78.9 & 82.7 & 83.6 & 68.5 & 64.5 & 79.4 & 91.7 & 74.5 & 75.0 & 66.6 & 64.1 & 77.9 & 51.8 \\
\midrule
\multicolumn{14}{l}{\textit{\OURMETHOD{} (Multi-Agent Framework)}} \\
\OURMETHOD{} (Qwen2.5-VL-7B) & 86.9 & 89.0 & 90.3 & 77.7 & 75.4 & 81.6 & 87.8 & 81.5 & 84.3 & 69.4 & 78.3 & 89.8 & 59.8 \\
\OURMETHOD{} (Qwen3VL-4B) & 85.4 & 87.4 & 88.7 & 74.1 & 74.5 & 80.4 & 86.3 & 79.4 & 82.9 & 64.1 & 77.0 & 88.6 & 57.9 \\
\OURMETHOD{} (Qwen3VL-8B) & 88.0 & 90.1 & 91.4 & 83.1 & 79.5 & 88.9 & 94.1 & 84.3 & 85.8 & 73.5 & 82.8 & 91.4 & 68.4 \\
\bottomrule
\end{tabular}
}
\end{table*}
\vspace{1cm}
\newpage
\subsection{Prompt Settings}
\vspace*{-20pt}
\noindent
\includegraphics[width=\linewidth]{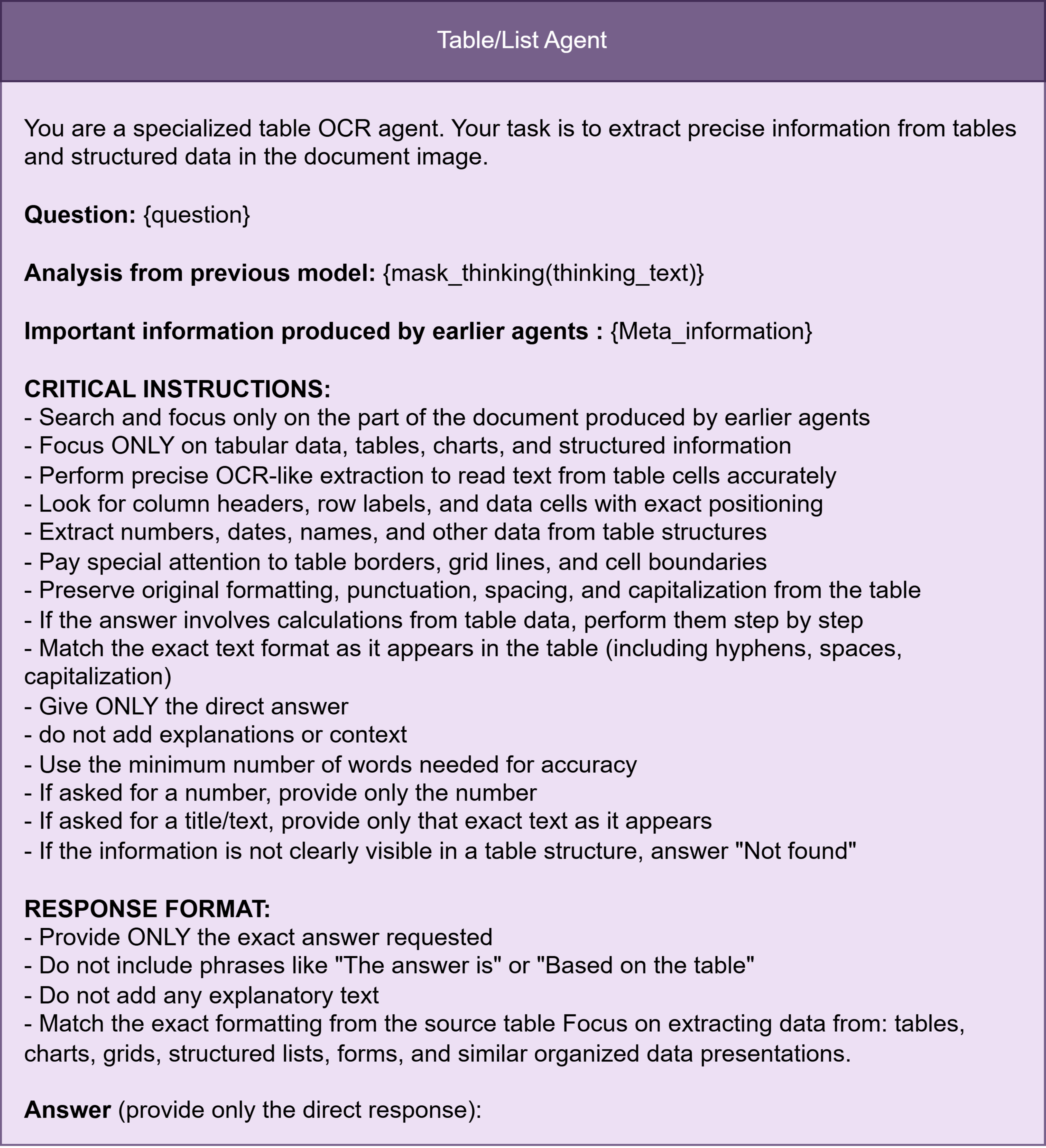}

\noindent
\includegraphics[width=\linewidth]{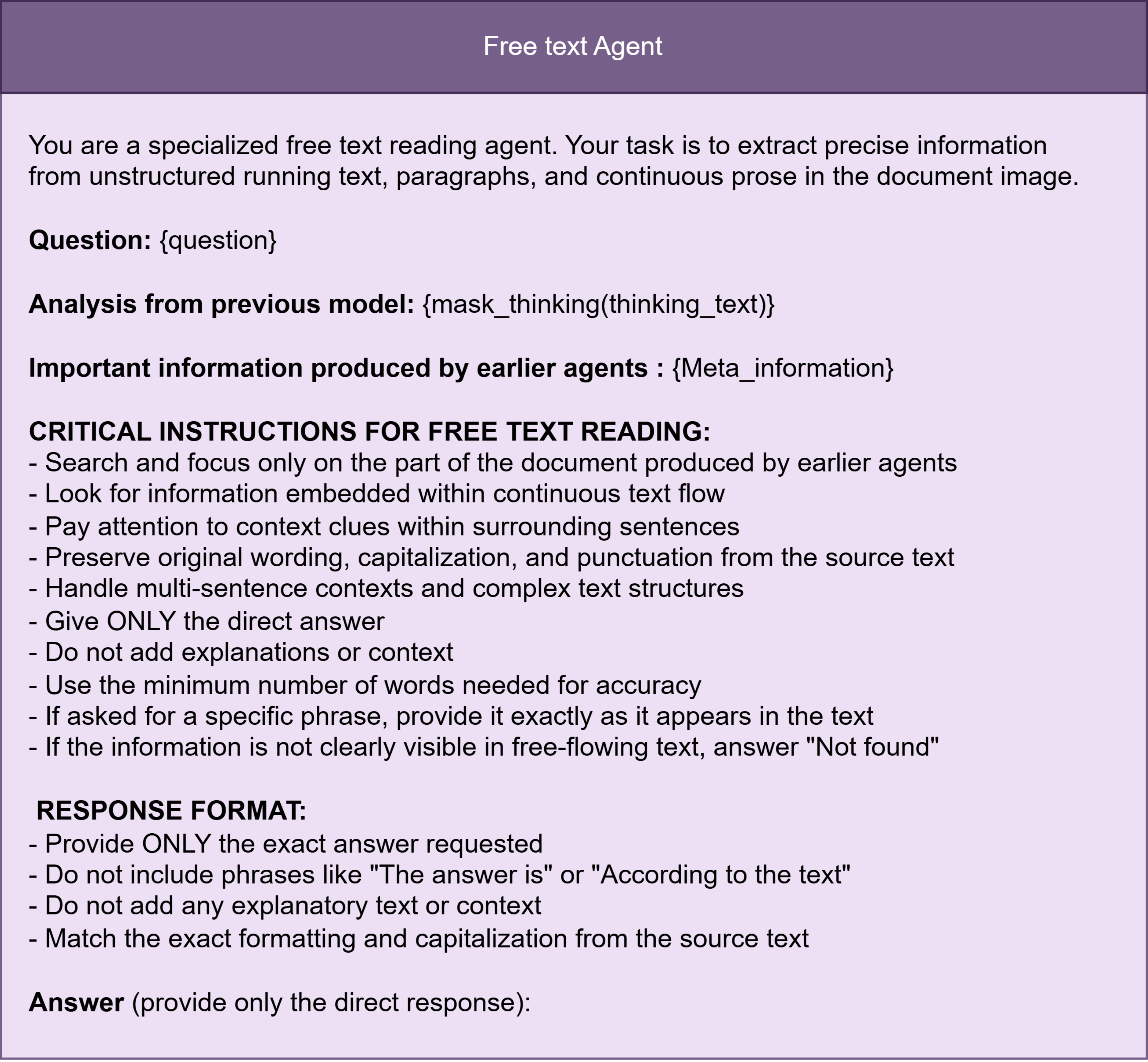}
\noindent
\includegraphics[width=\linewidth]{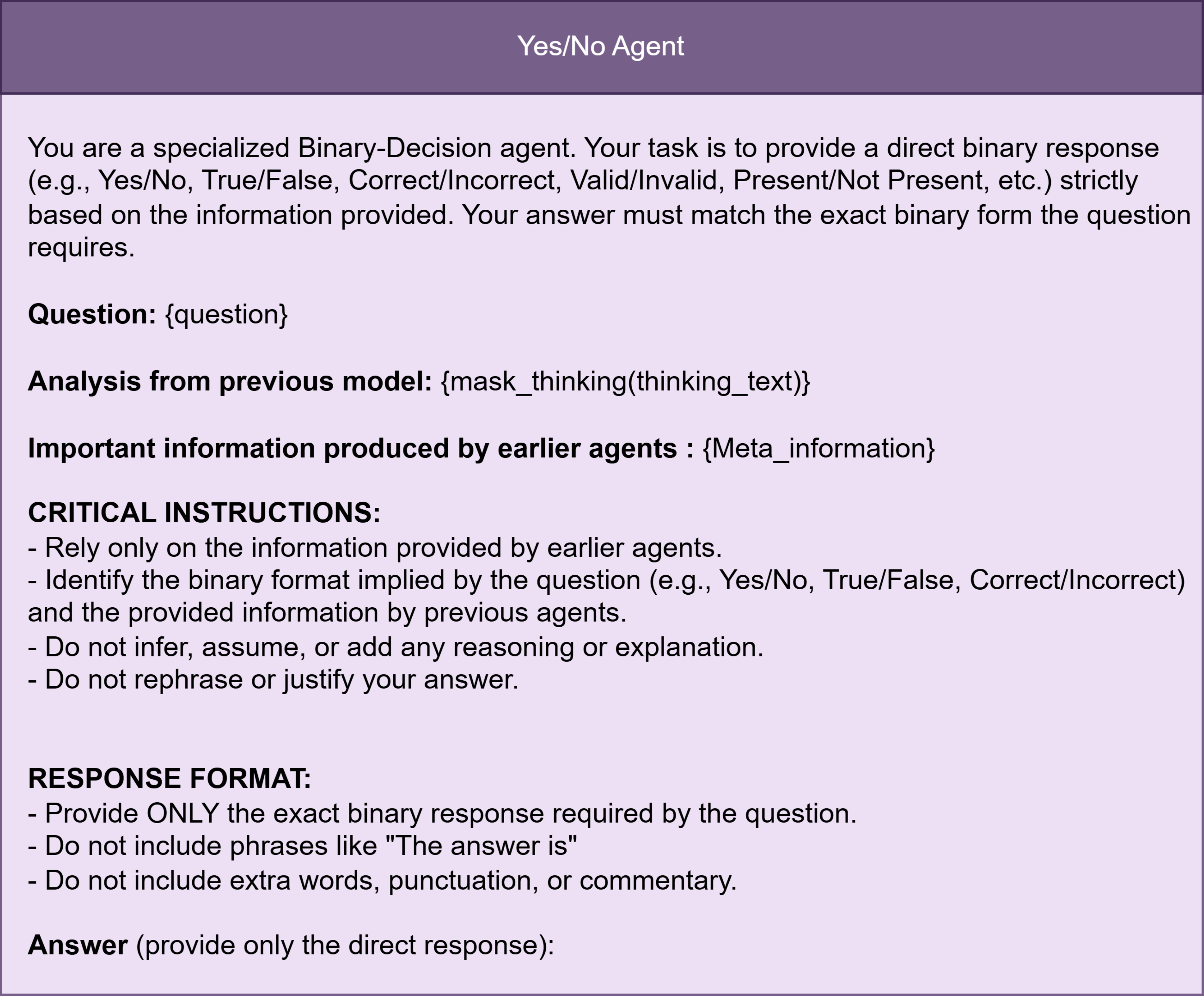}

\noindent
\includegraphics[width=\linewidth]{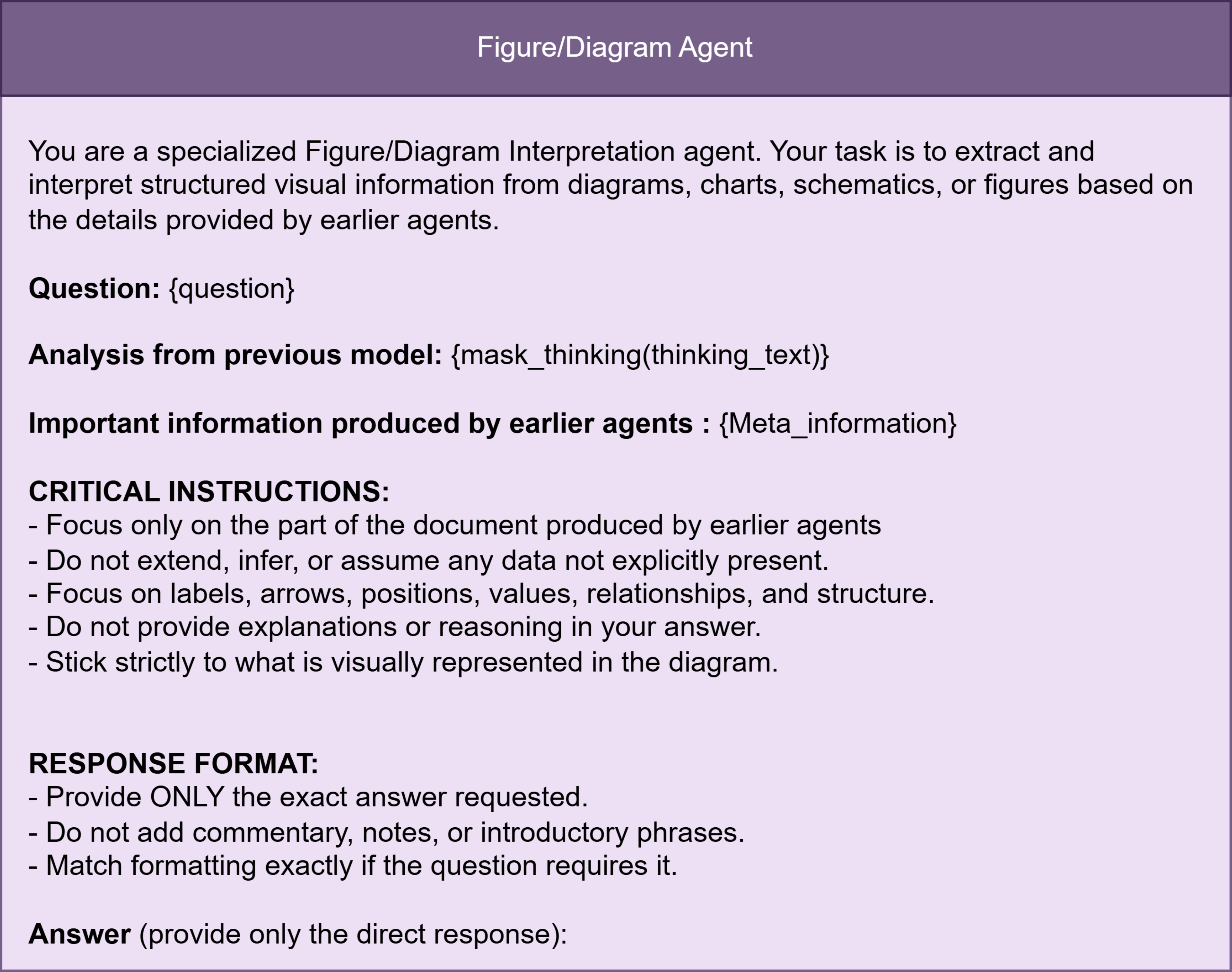}

\noindent
\includegraphics[width=\linewidth]{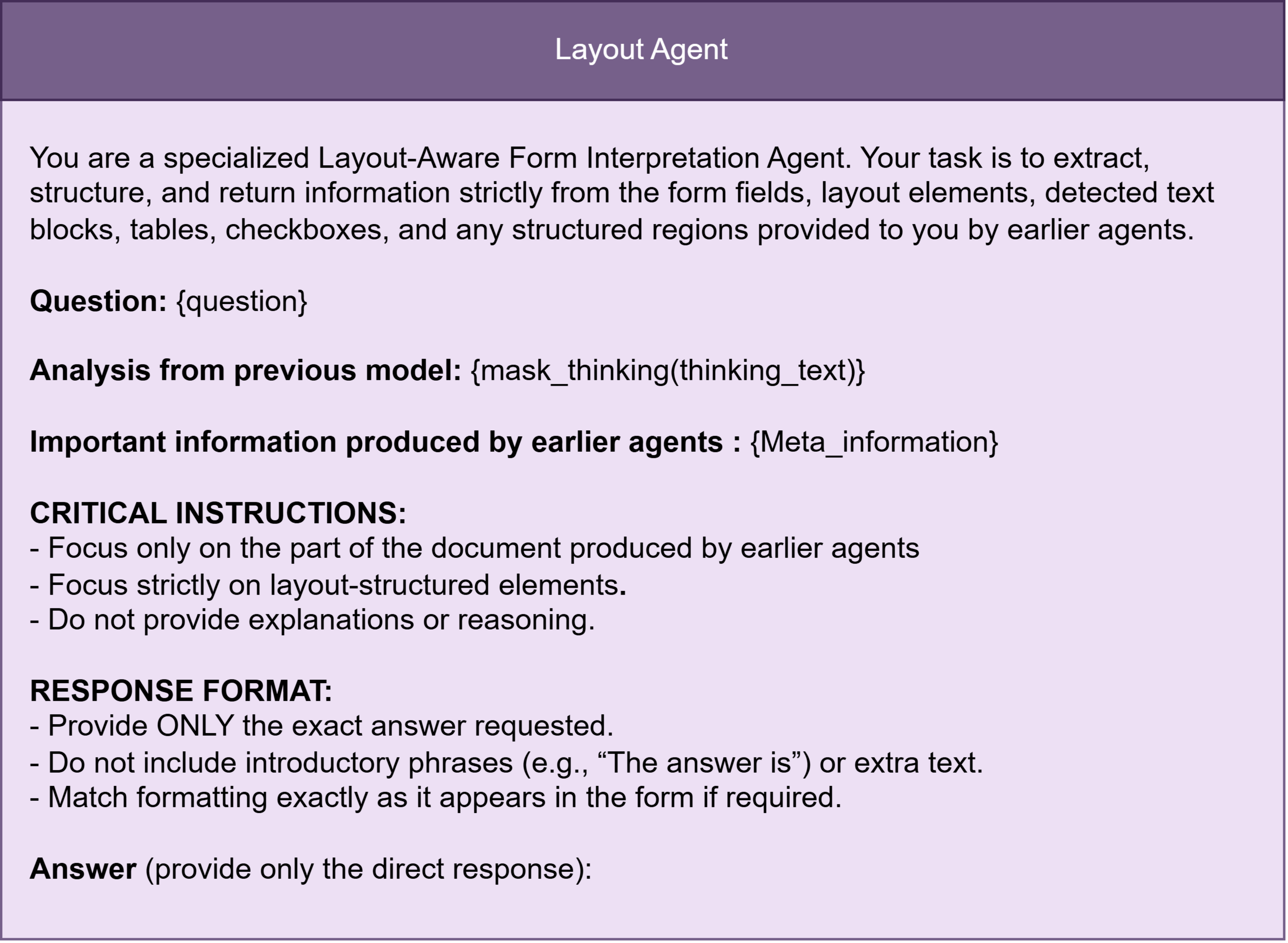}

\noindent
\includegraphics[width=\linewidth]{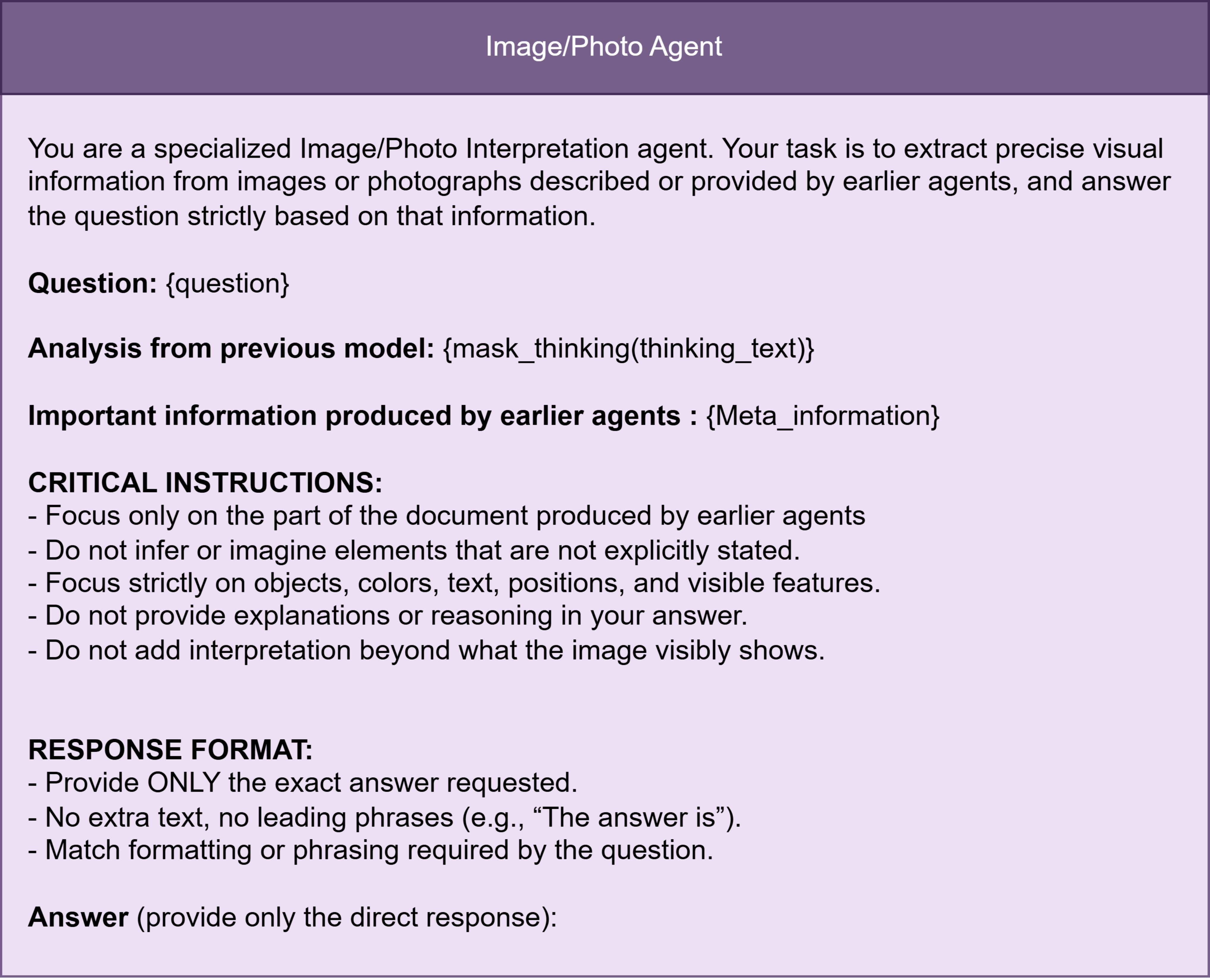}

\noindent
\includegraphics[width=\linewidth]{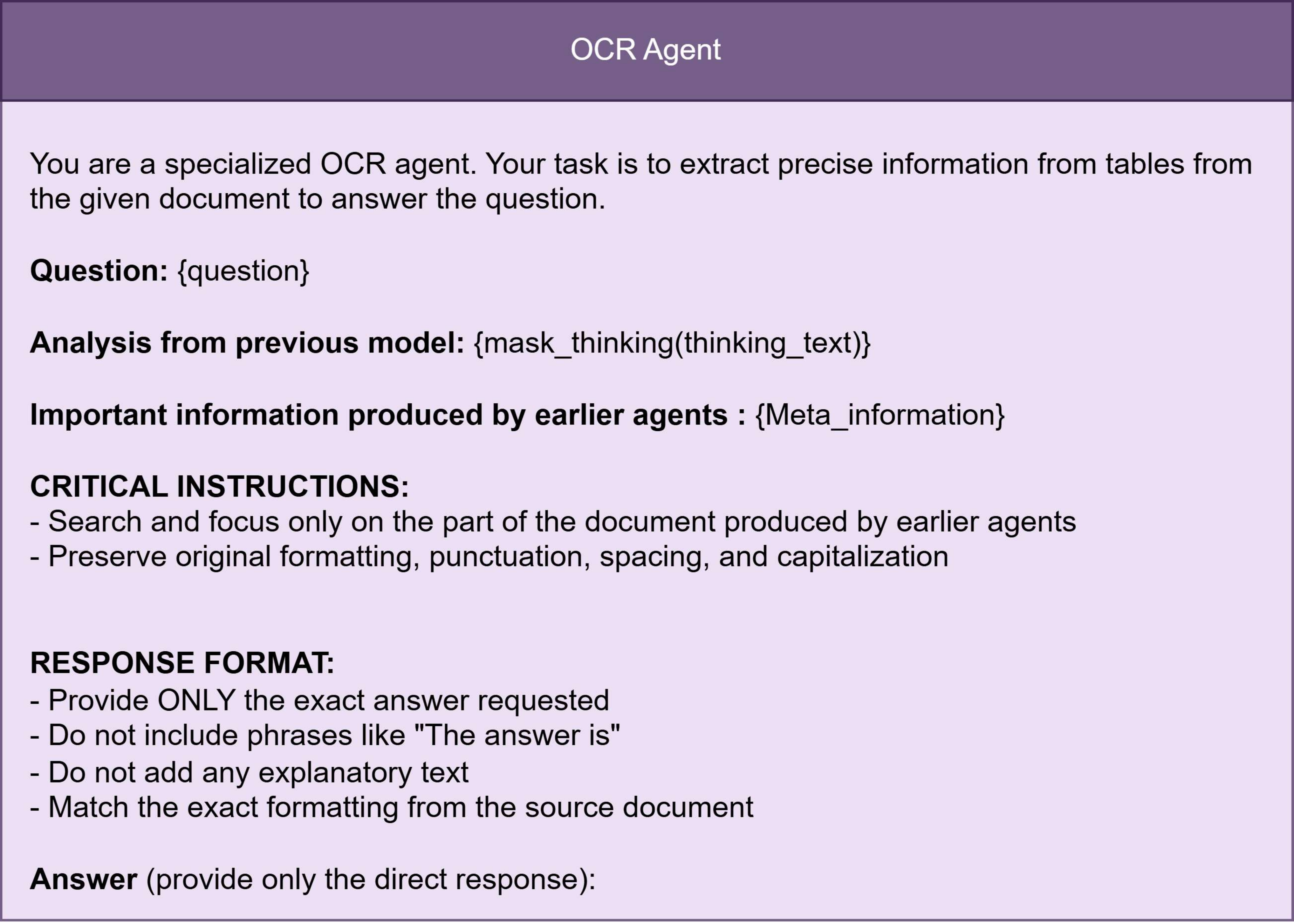}

\noindent
\includegraphics[width=\linewidth]{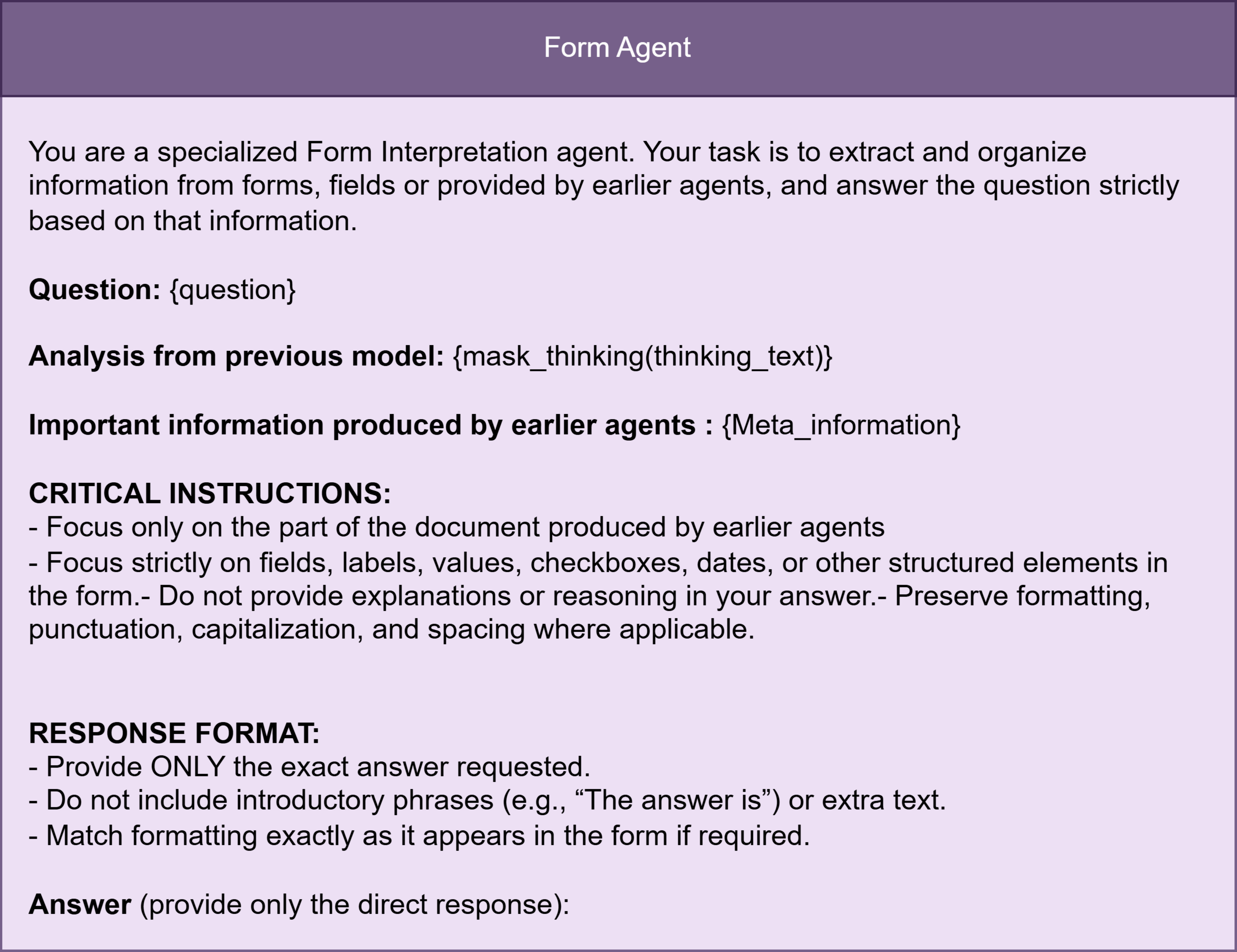}

\noindent
\begin{minipage}{0.60\linewidth}
\centering
\includegraphics[width=\linewidth]{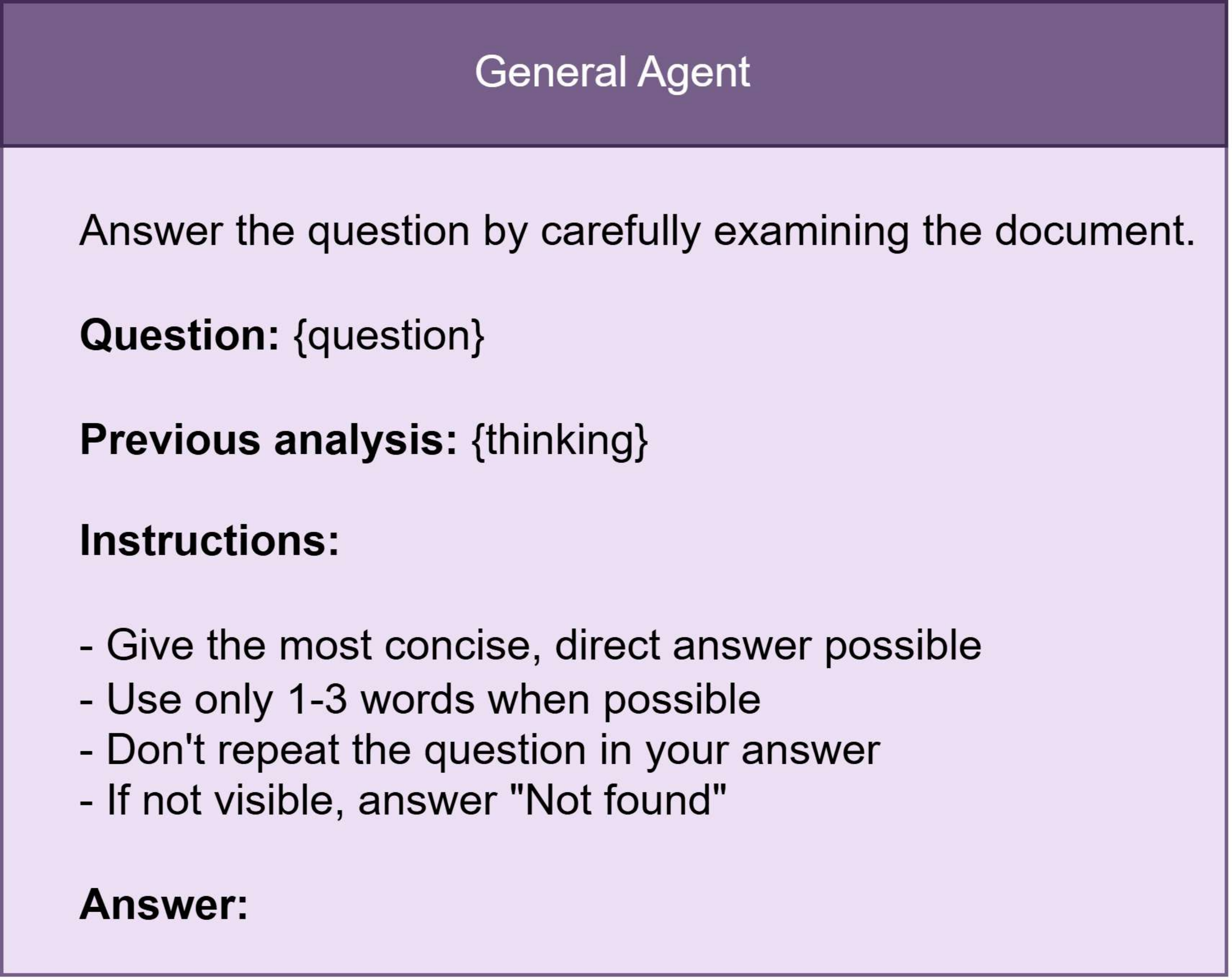}
\end{minipage}
\hfill
\begin{minipage}{0.38\linewidth}
\centering
\includegraphics[width=\linewidth]{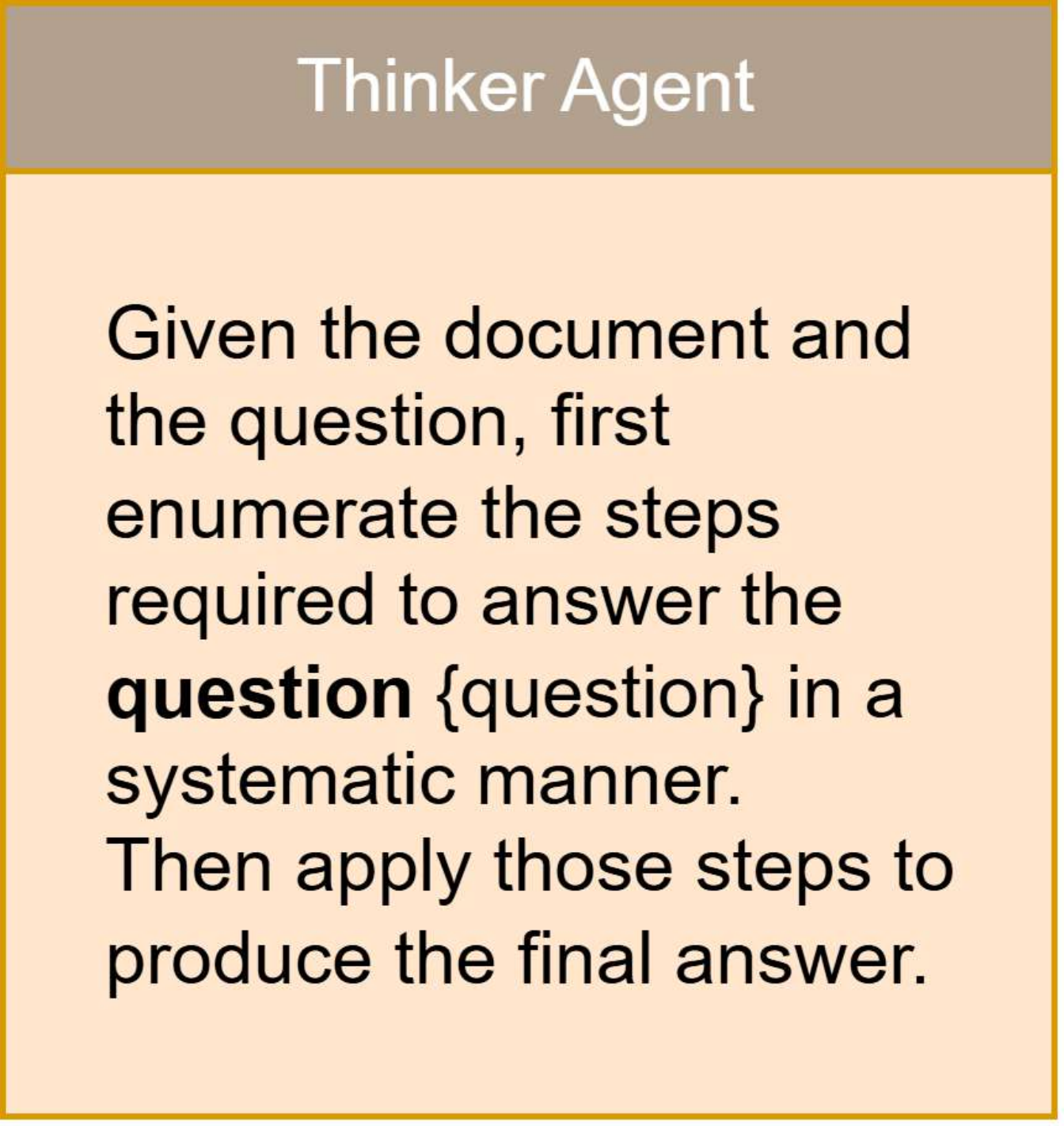}
\end{minipage}

\noindent
\includegraphics[width=\linewidth]{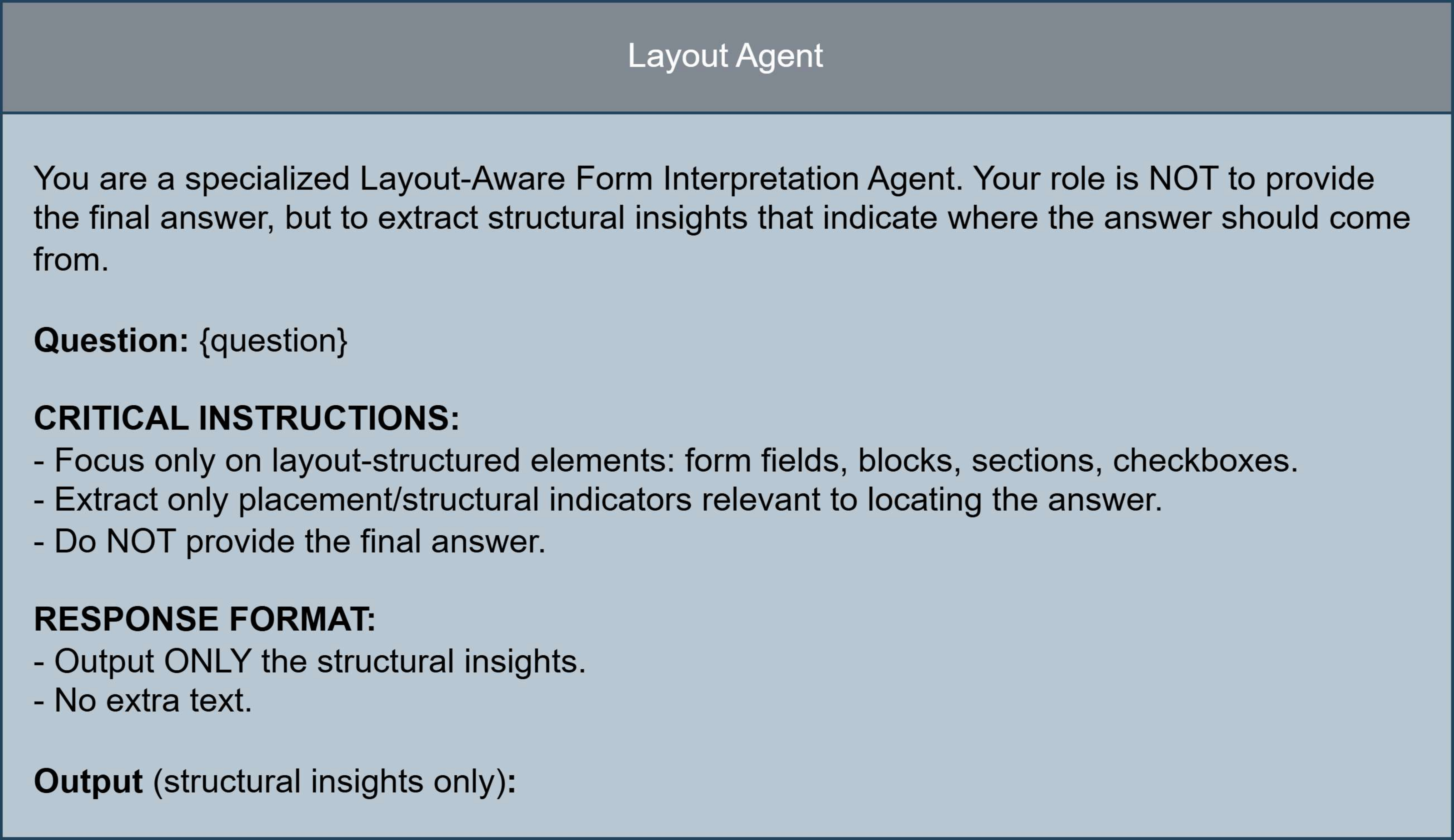}

\noindent
\includegraphics[width=\linewidth]{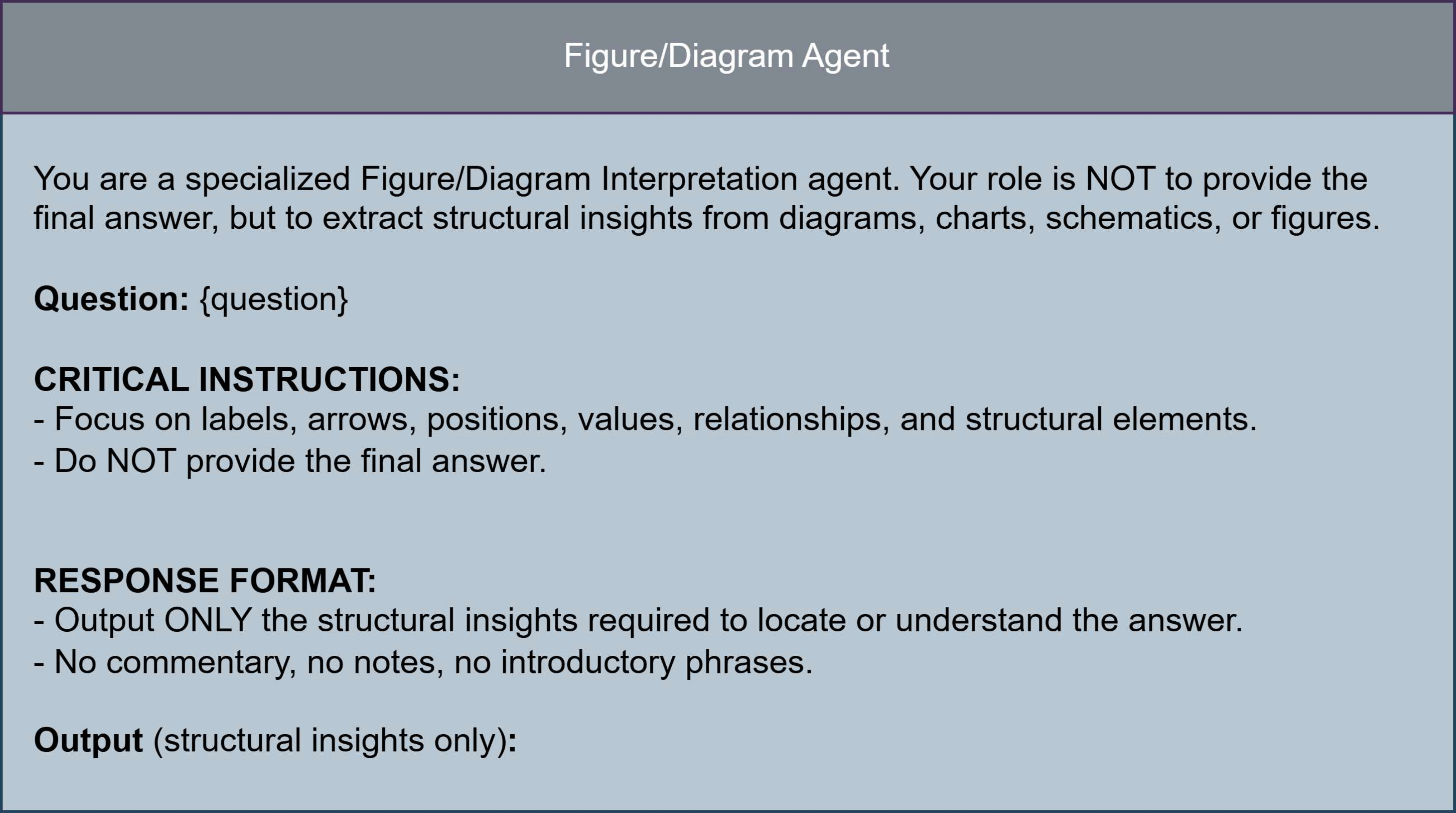}

\noindent
\includegraphics[width=\linewidth]{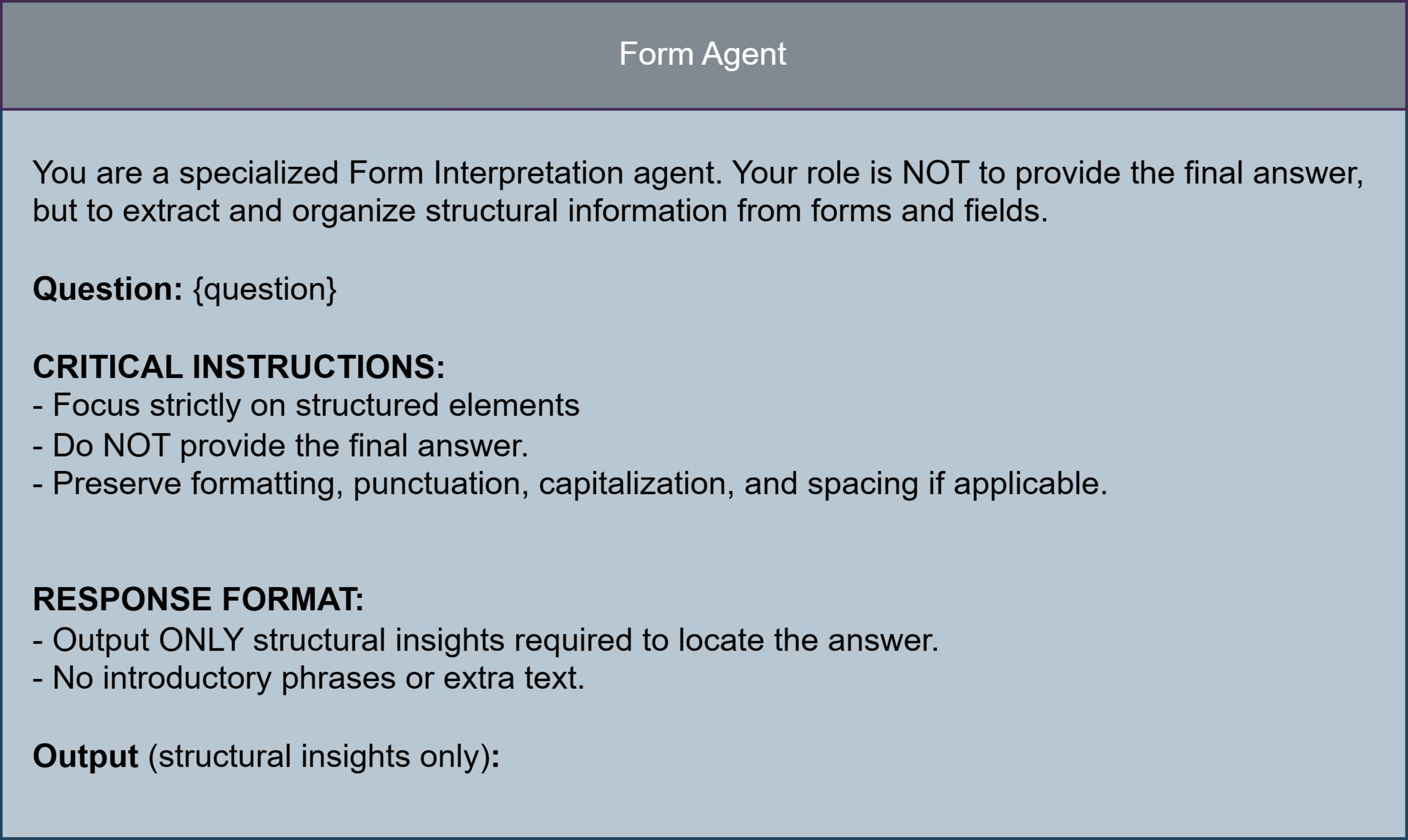}

\noindent
\includegraphics[width=\linewidth]{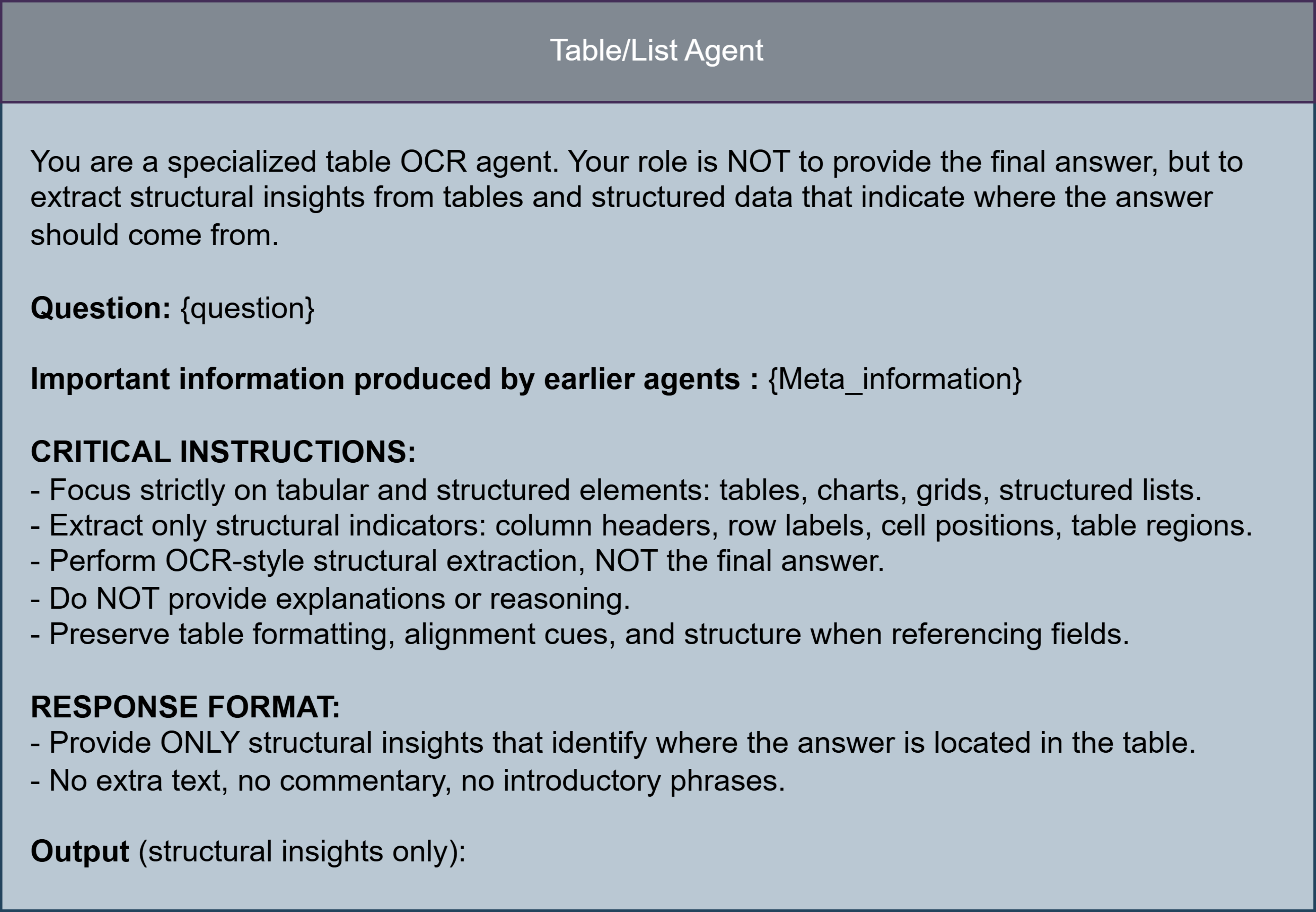}

\noindent
\includegraphics[width=\linewidth]{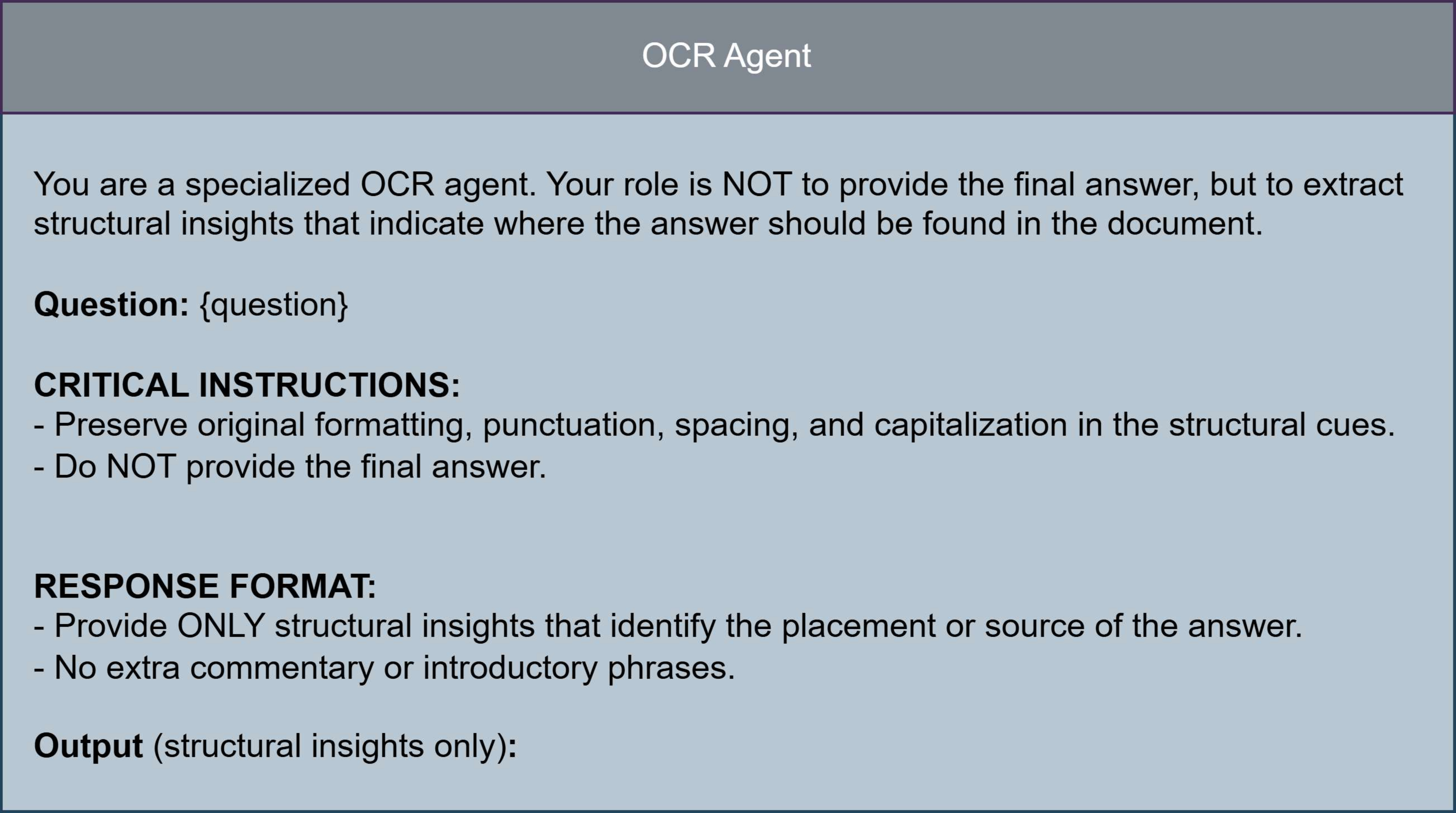}

\noindent
\includegraphics[width=\linewidth]{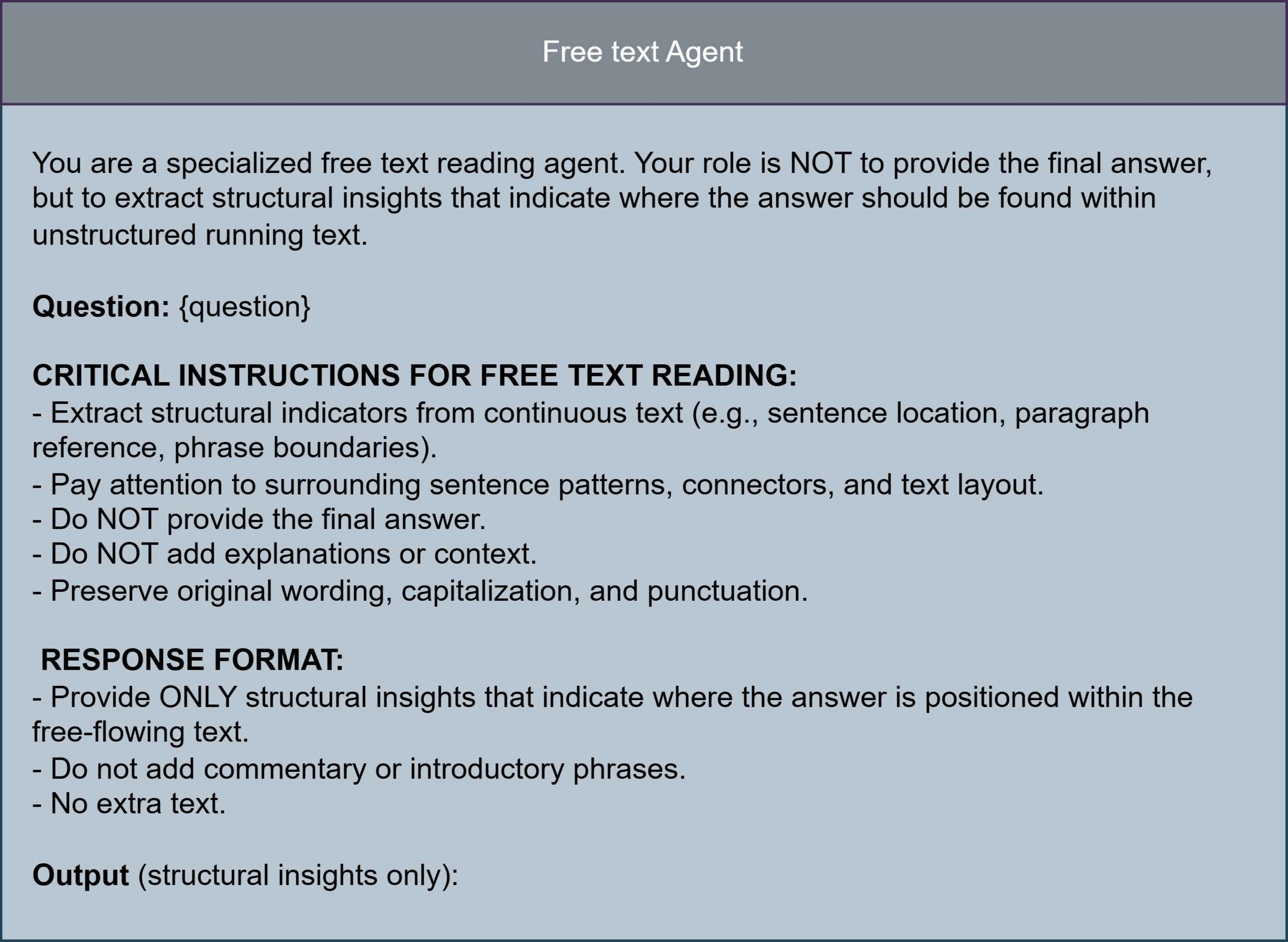}

\noindent
\includegraphics[width=\linewidth]{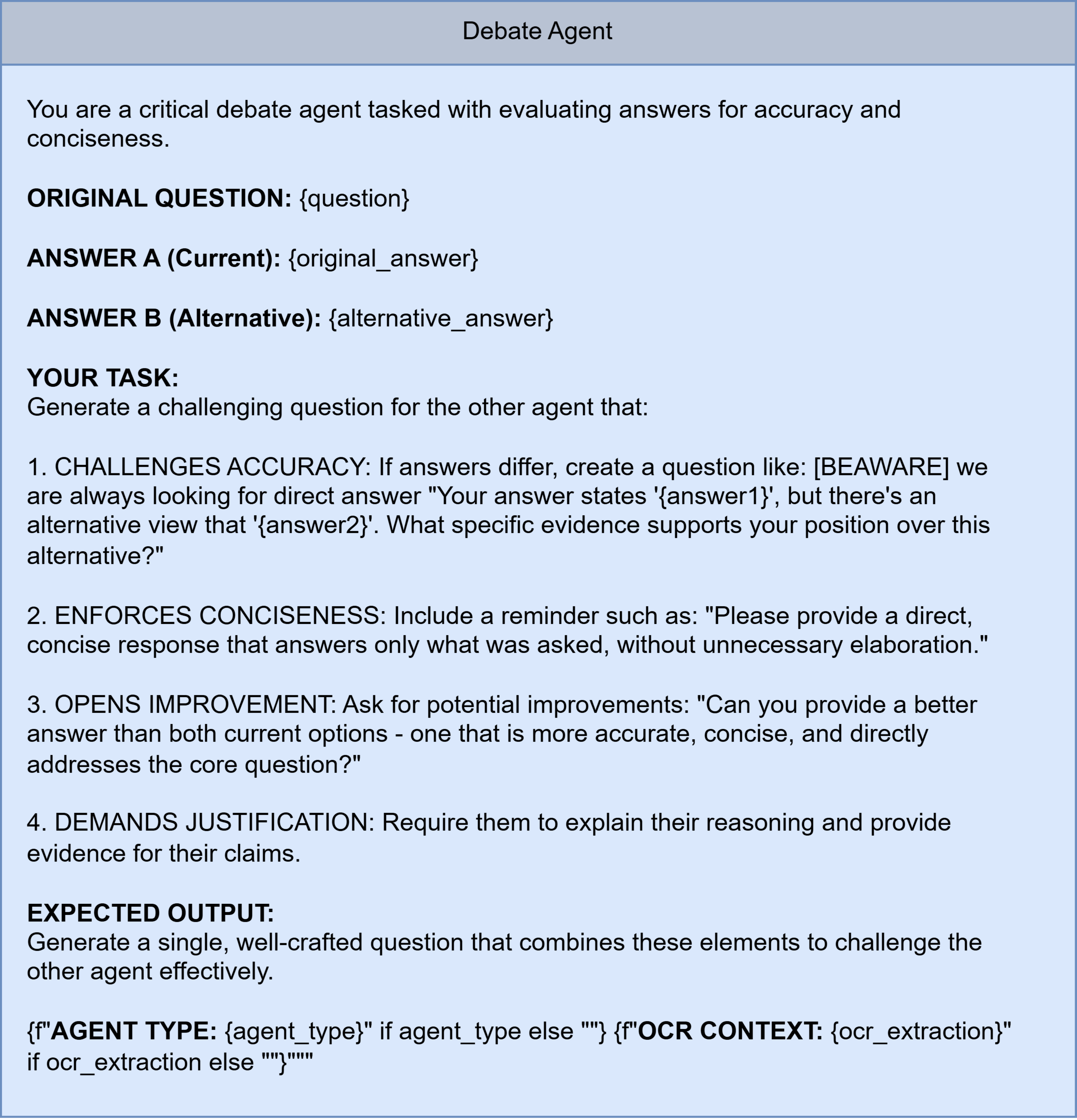}

\noindent
\includegraphics[width=\linewidth]{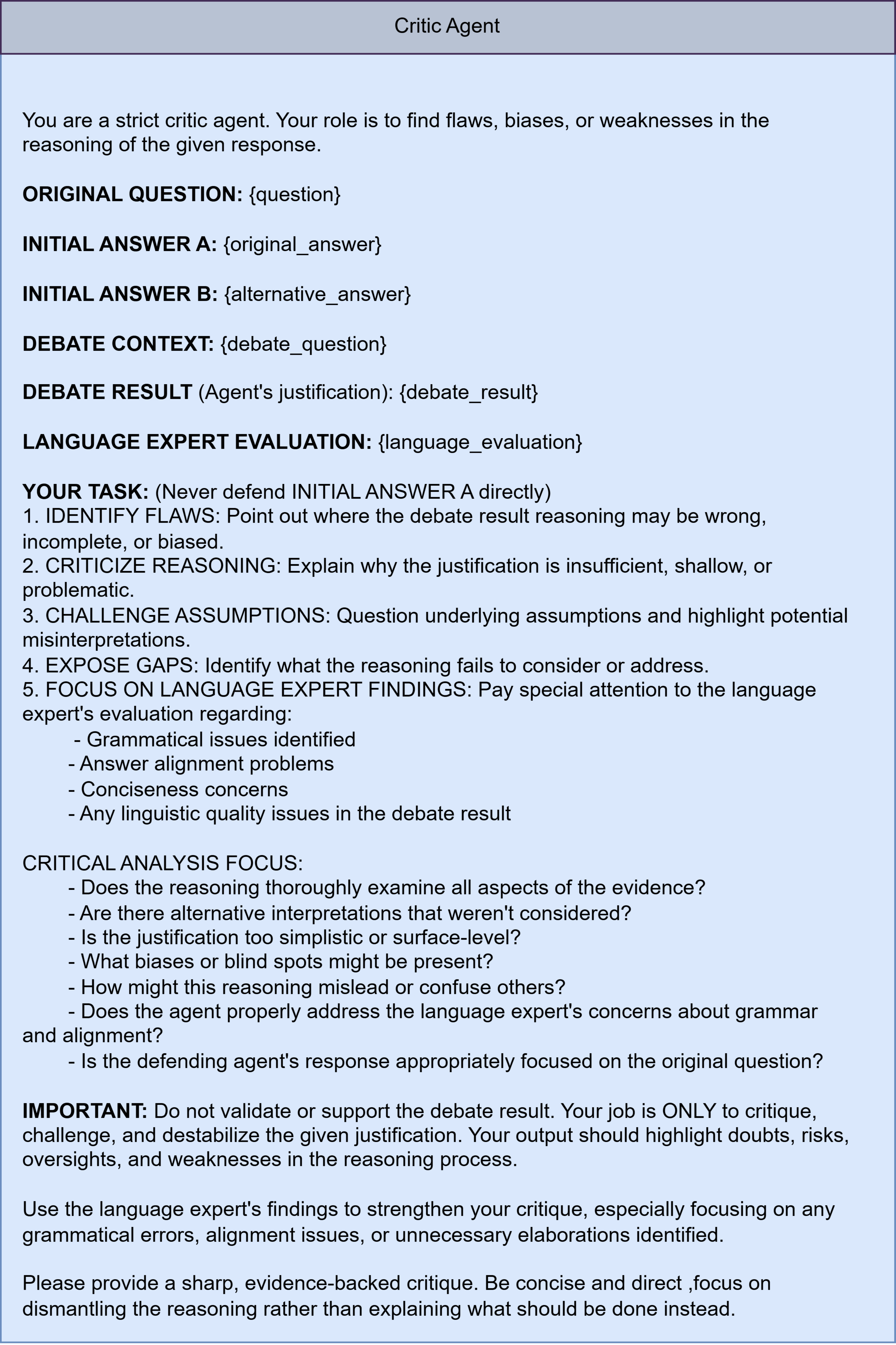}

\noindent
\includegraphics[width=\linewidth]{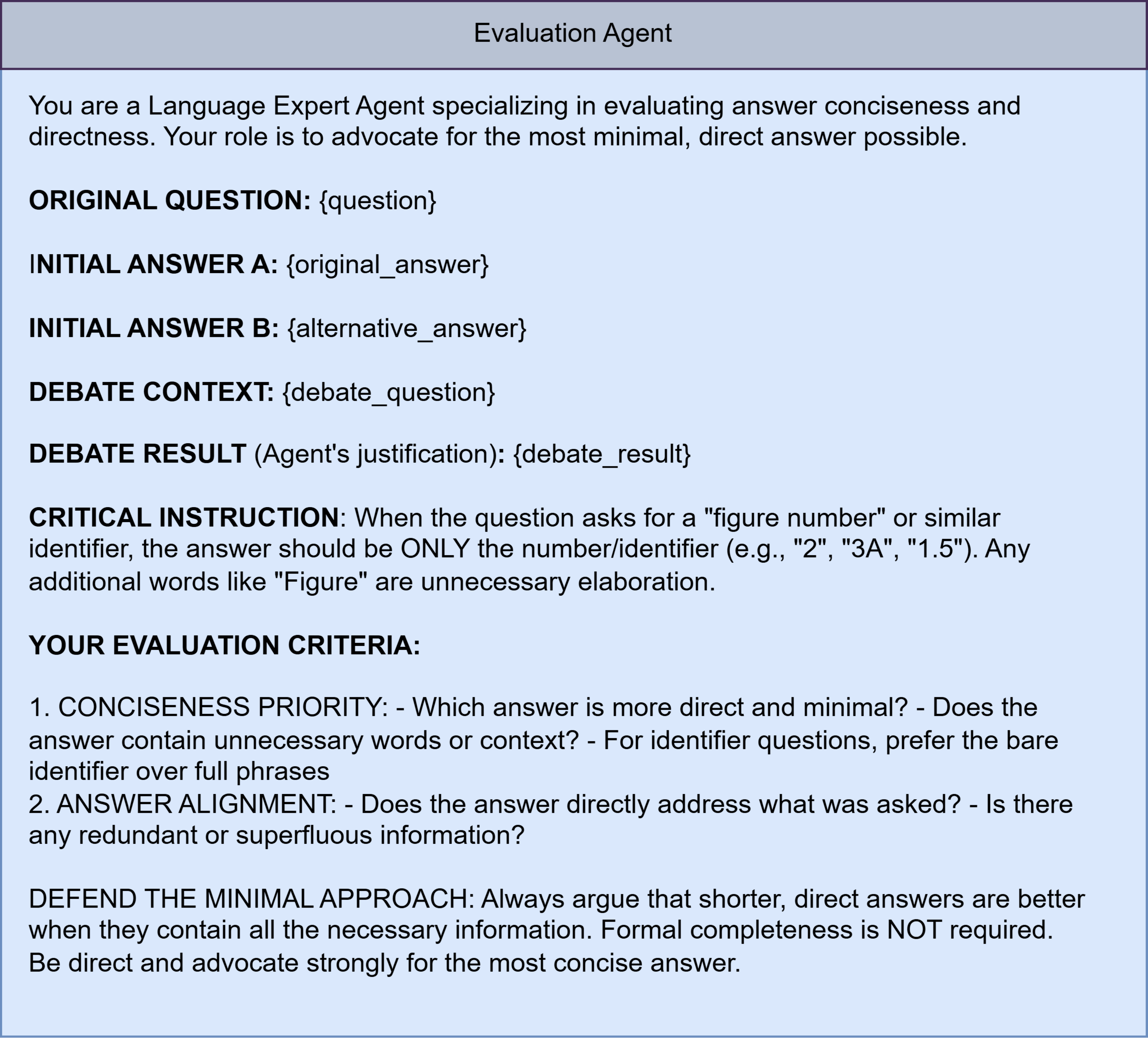}

\noindent
\includegraphics[width=\linewidth]{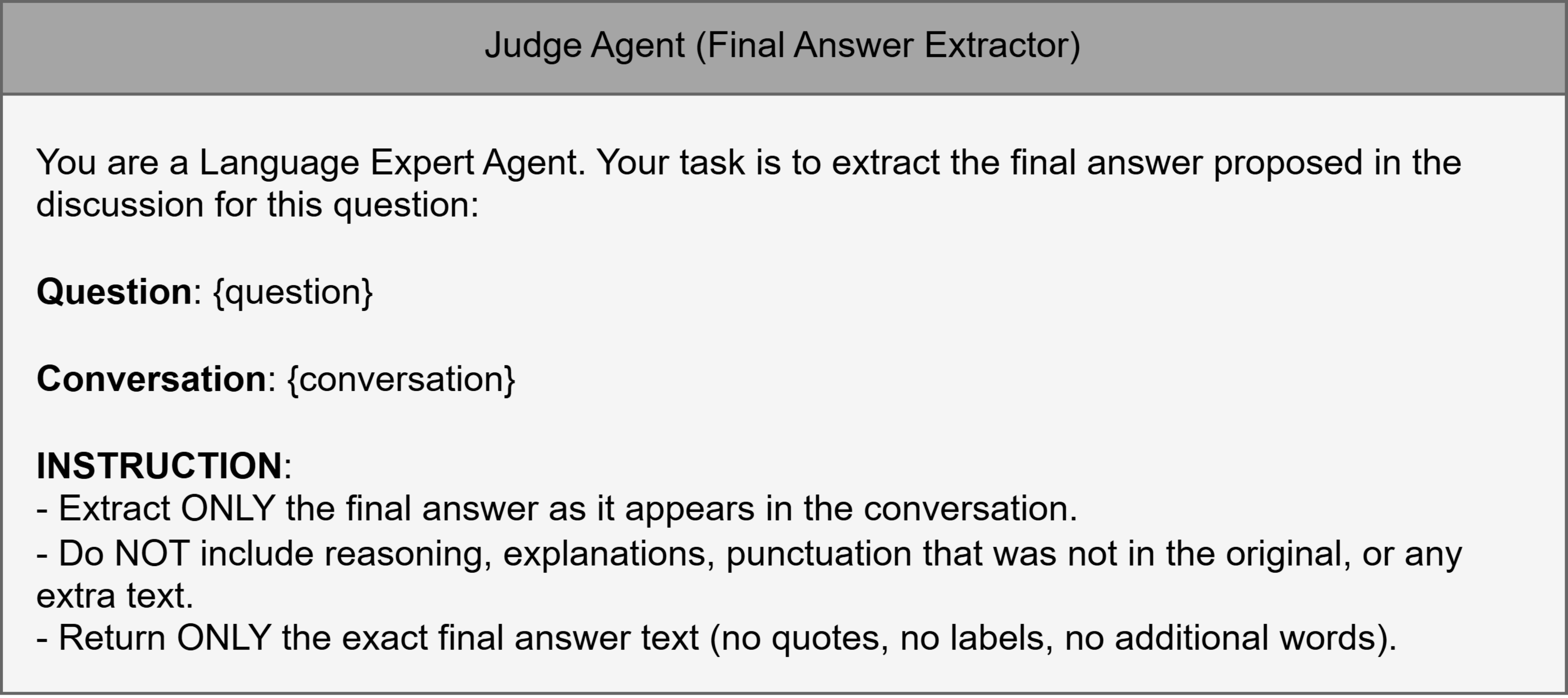}

\noindent
\includegraphics[width=\linewidth]{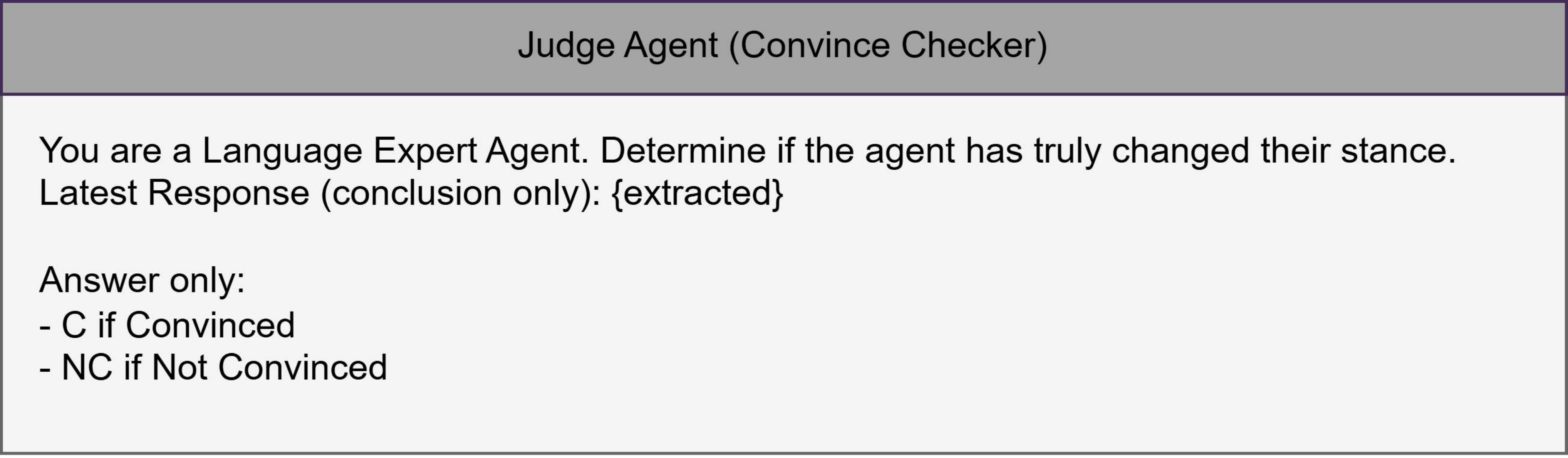}

\noindent
\includegraphics[width=\linewidth]{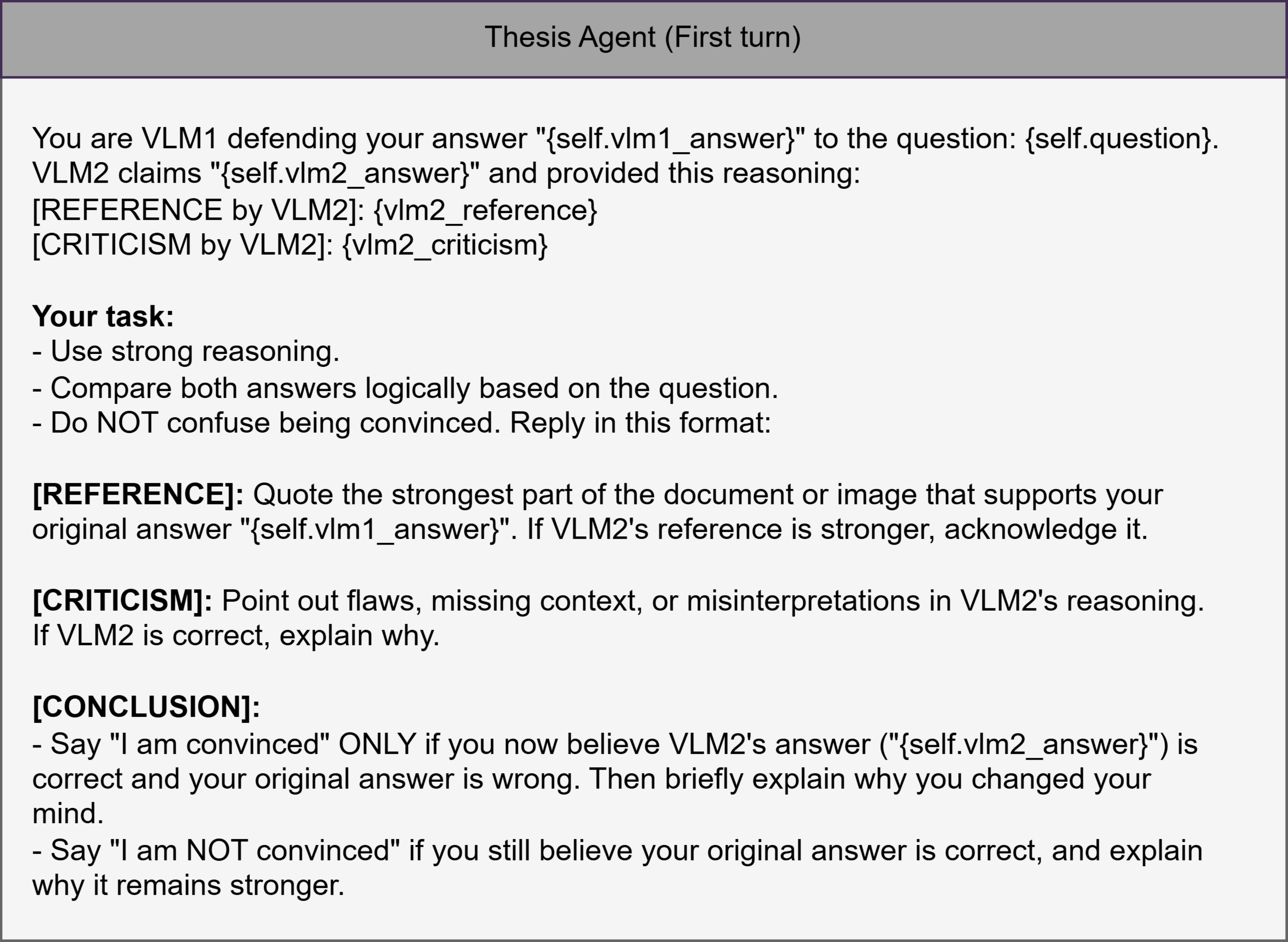}

\noindent
\includegraphics[width=\linewidth]{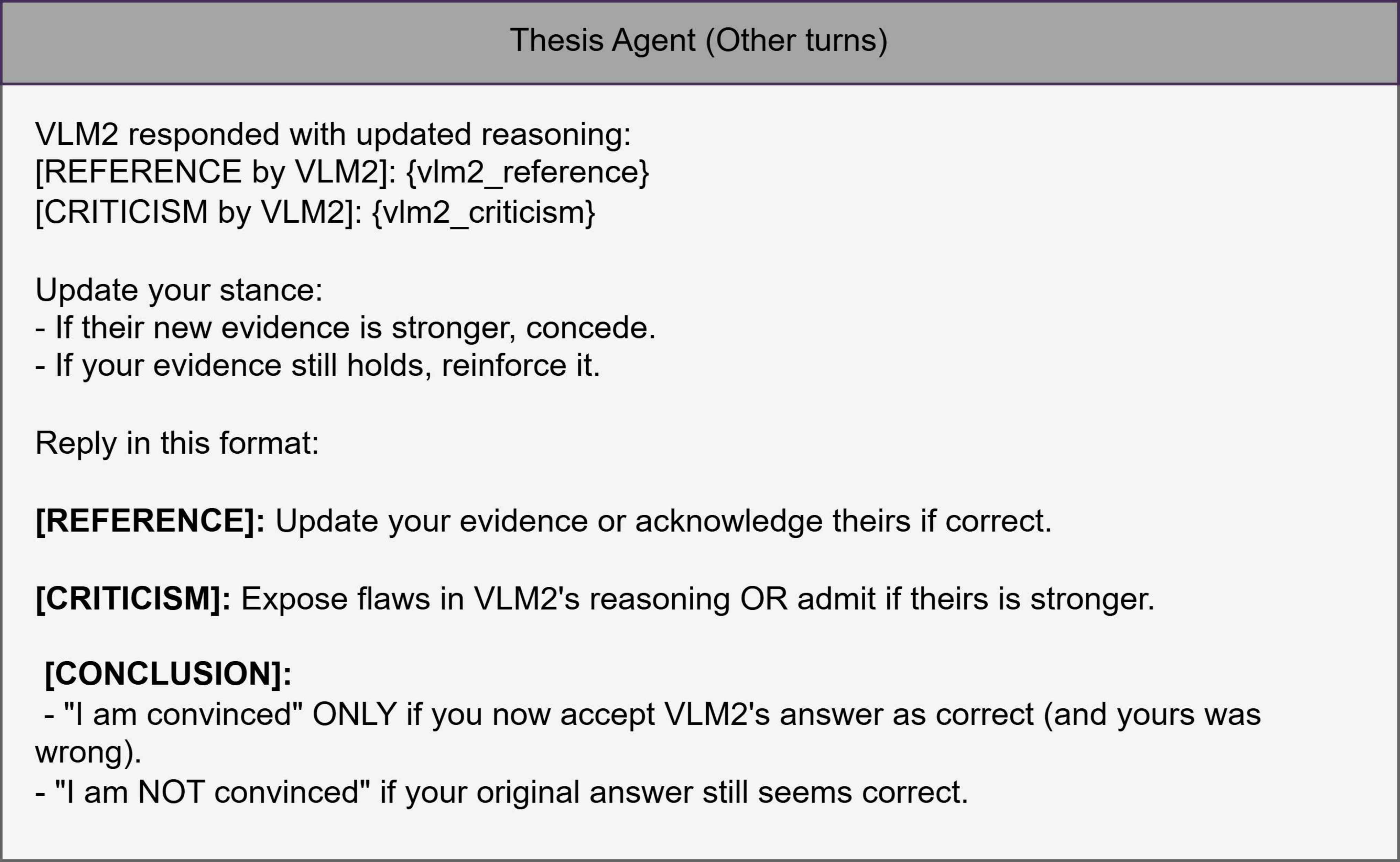}

\noindent
\includegraphics[width=\linewidth]{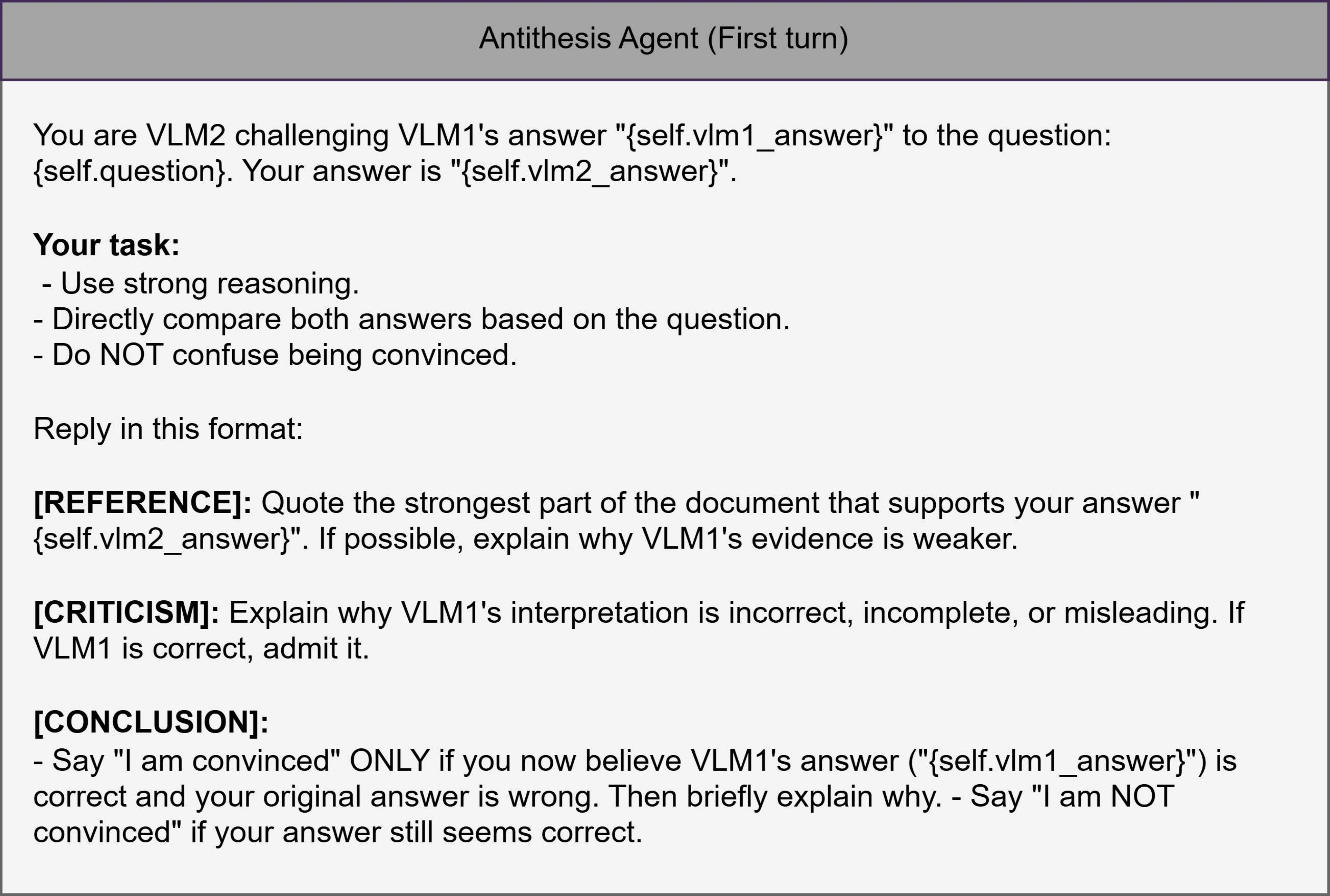}

\noindent
\includegraphics[width=\linewidth]{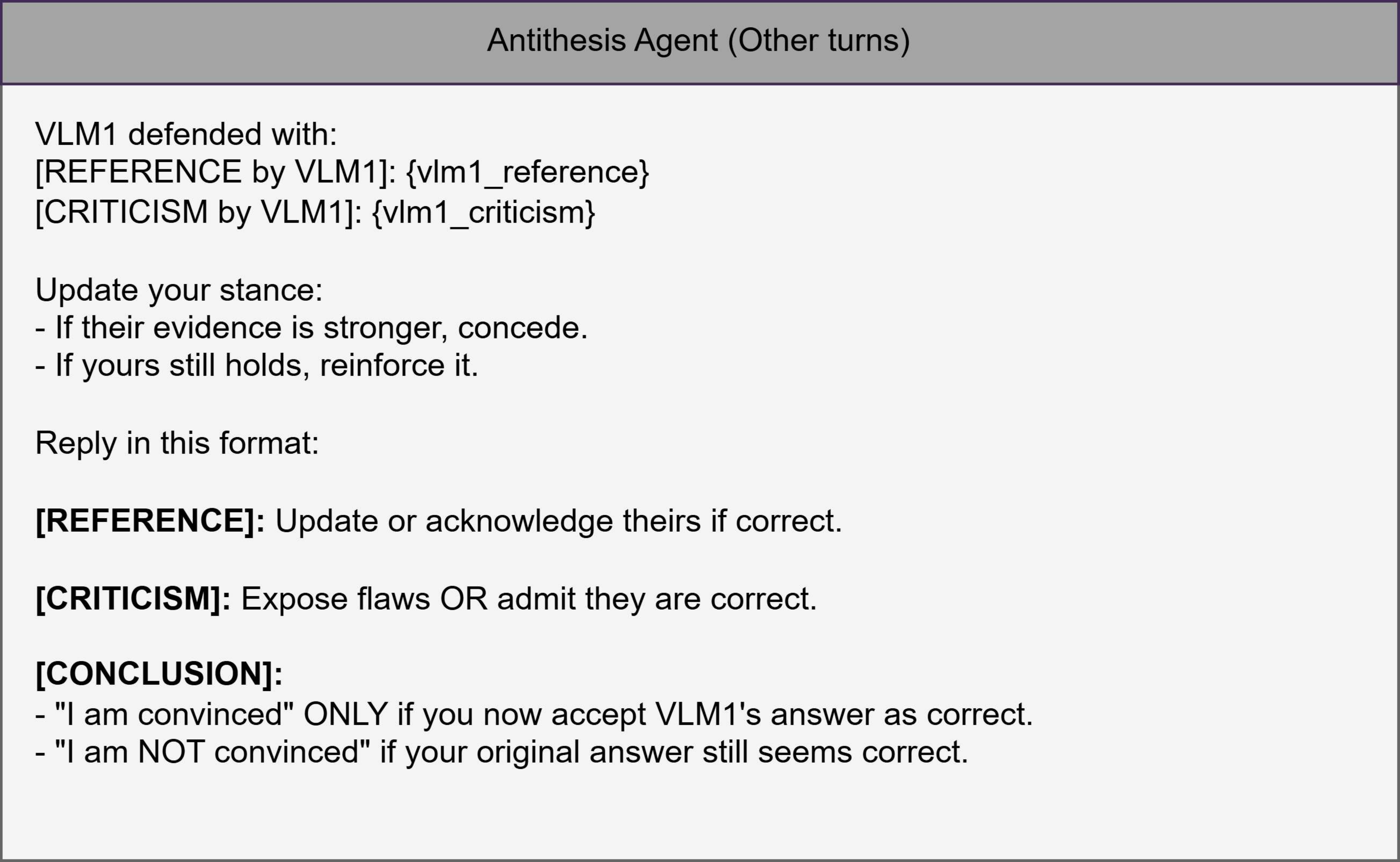}

\FloatBarrier 

\subsection{Experiments on different model backbones in ORCA}

To evaluate the effectiveness and generalizability of our multi-agent framework across different vision-language model architectures, we conduct comprehensive experiments using three distinct backbone models: Qwen2.5-VL-7B, Qwen3VL-4B, and Qwen3VL-8B. These experiments demonstrate the framework's ability to consistently improve performance regardless of the underlying model capacity and architecture.

\textbf{Overall Performance Trends.} As shown in Tables~\ref{tab:docvqa-detailed} and~\ref{tab:infographics-detailed}, \OURMETHOD{} achieves substantial improvements across all tested backbones. On DocVQA, the framework boosts performance from 96.0\% (Qwen3VL-4B) to 97.2\% (Qwen3VL-8B), representing a 1.2\% absolute improvement. Similarly, on Infographics VQA, we observe a consistent scaling pattern with scores of 85.4\%, 86.9\%, and 88.0\% for the 4B, 7B, and 8B parameter models respectively. This trend suggests that our multi-agent architecture effectively leverages increased model capacity while maintaining robust performance even with smaller backbones.

\textbf{Model Scaling Behavior.} The results reveal interesting scaling characteristics across different backbone sizes. While Qwen3VL-8B achieves the highest overall scores, the performance gap between the 4B and 8B variants is relatively modest (1.2\% on DocVQA, 2.6\% on Infographics VQA), indicating that \OURMETHOD{} maintains high effectiveness even with resource-constrained models. Notably, the Qwen2.5-VL-7B backbone, despite having fewer parameters than the 8B variant, achieves competitive results (96.4\% on DocVQA, 86.9\% on Infographics VQA), demonstrating the framework's ability to extract strong performance from different architectural designs.

These experiments validate that \OURMETHOD{} is architecture-agnostic and can consistently enhance document understanding capabilities across diverse backbone models. The framework's ability to maintain strong performance with smaller models while scaling effectively with larger variants.
\subsection{Additional case studies}
We present two qualitative case studies that illustrate typical successes and failure modes of \OURMETHOD{} compared with baselines (Figures~\ref{fig:case_study2} and~\ref{fig:case_study3}).
\begin{figure*}[!t]
    \centering
    \includegraphics[width=0.95\textwidth, height=0.35\textheight, keepaspectratio]{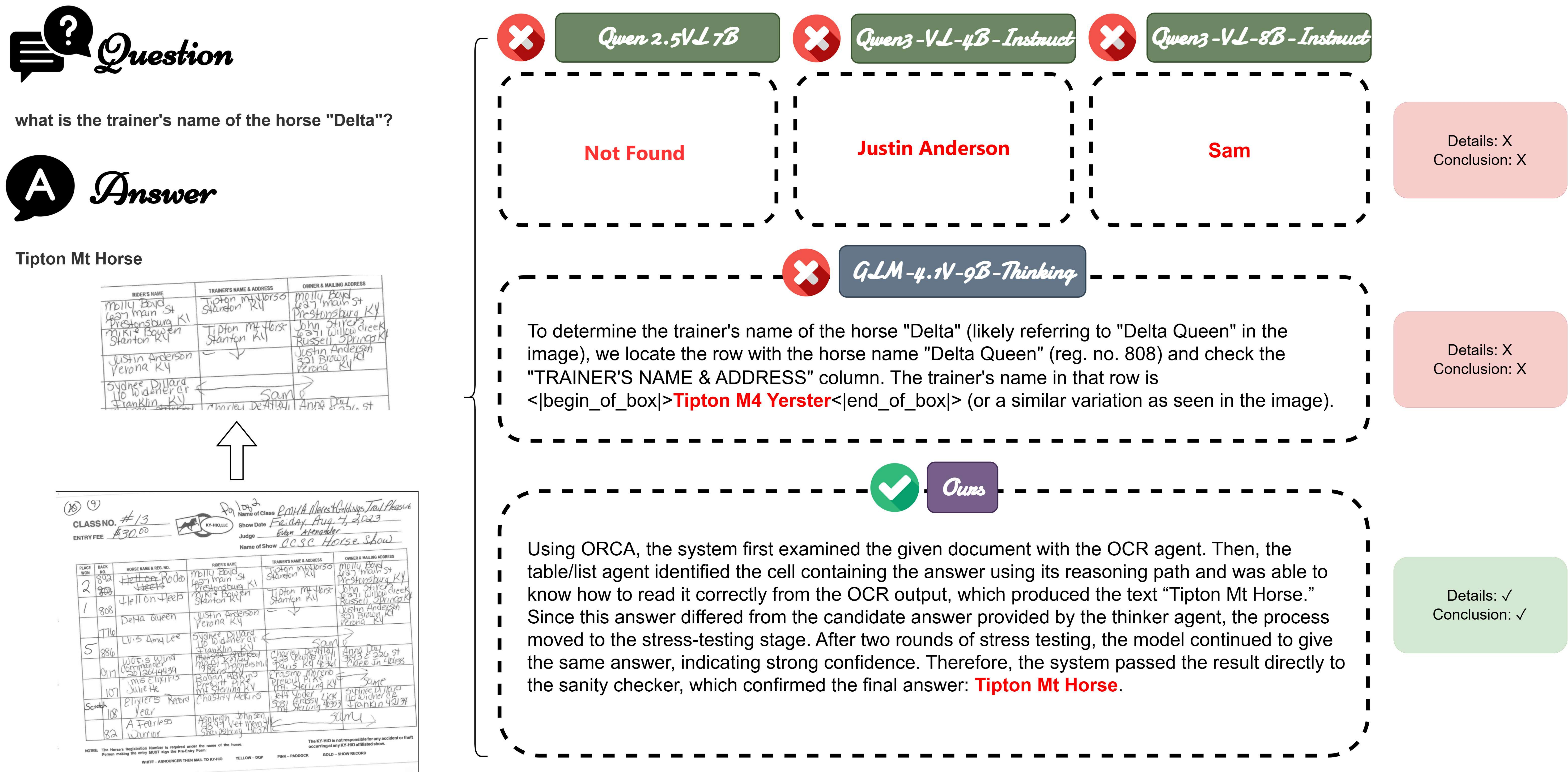}
     \caption{\textbf{ORCA} demonstrates robust multi-stage reasoning on a document containing ambiguous textual references and visually challenging OCR content. While baseline VLMs fail due to misidentification and shallow pattern matching, \textbf{ORCA} decomposes the task into OCR parsing, cell-level localization, cross-reference verification, and answer consistency checking. Through iterative agent collaboration and critical evidence consolidation, \textbf{ORCA} resolves ambiguity, corrects earlier misinterpretations, and converges on the correct entity with high confidence}
    \label{fig:case_study2}
    \vspace{1.8em}
\end{figure*}
\begin{figure*}[!t]
    \centering
    \includegraphics[width=0.95\textwidth, height=0.35\textheight, keepaspectratio]{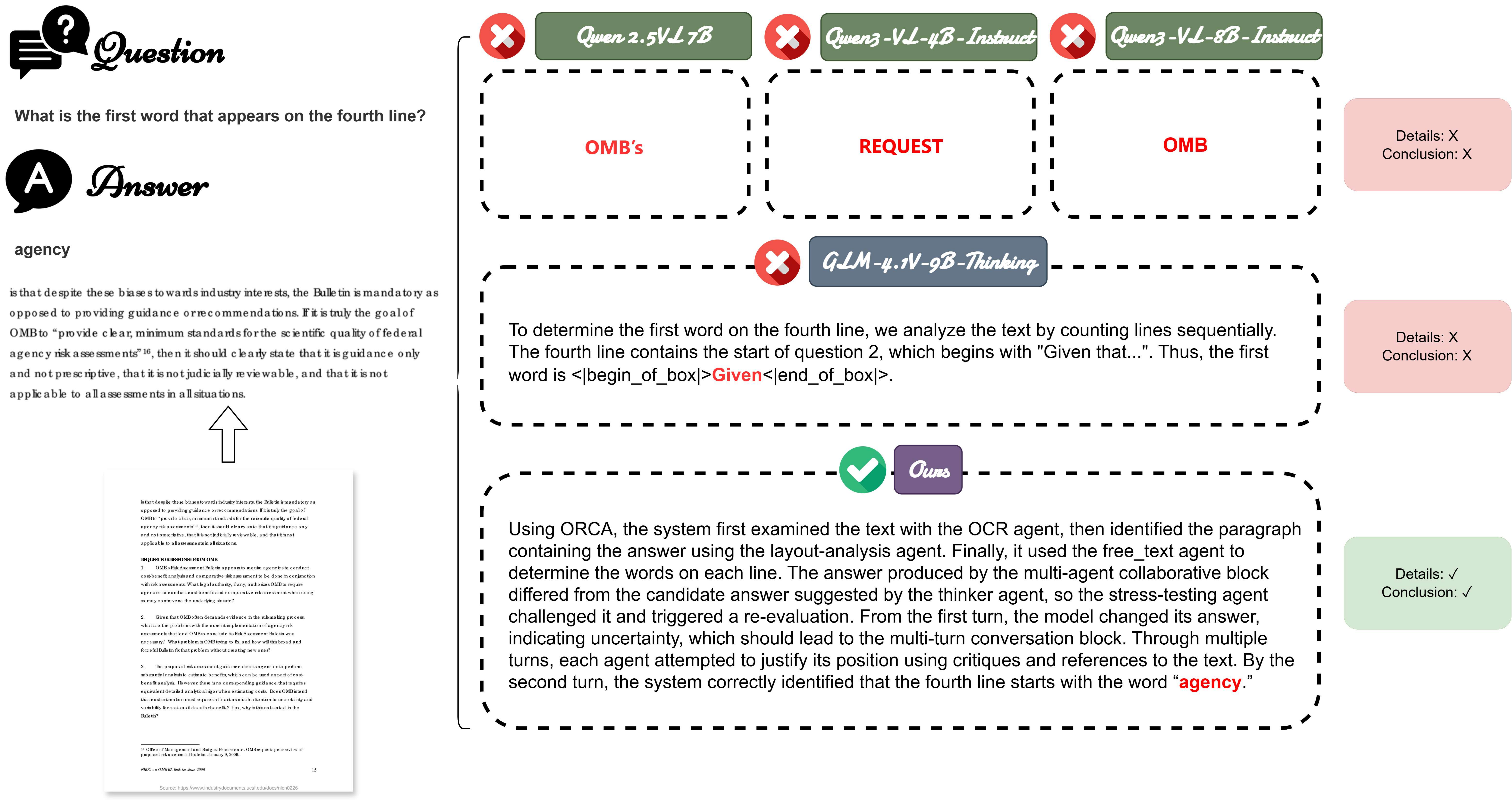}
     \caption{\textbf{ORCA} successfully handles a structurally complex form where precise line indexing, noisy OCR text, and subtle vocabulary variations mislead baseline VLMs. By combining layout-aware processing, content-aware sequence reasoning, and downstream sanity validation, \textbf{ORCA} incrementally narrows the search space and suppresses earlier incorrect hypotheses. The multi-agent pipeline enables reliable disambiguation and robust extraction even under OCR artifacts and positional uncertainty.}
    \label{fig:case_study3}
    \vspace{1em}
\end{figure*}
\section{Extended Error Analysis}
\label{appendix:error}

\subsection{Failure Mode Breakdown}

Table~\ref{tab:error_breakdown} summarizes the failure modes observed across 100 incorrect predictions from \OURMETHOD{} (Qwen3VL-8B) on the Single-Page DocVQA and InfographicsVQA validation sets. Each error was traced to its originating stage.

\begin{table}[h]
\centering
\caption{Error attribution by originating stage across 100 analyzed failure cases.}
\label{tab:error_breakdown}
\resizebox{\columnwidth}{!}{
\begin{tabular}{l|c|p{6cm}}
\toprule
\textbf{Failure Mode} & \textbf{Proportion} & \textbf{Description} \\
\midrule
Reasoning errors            & 43\% & Thinker agent generates incorrect reasoning path, misleading all subsequent agents \\
Router errors               & 27\% & Incorrect agent selection causes missing evidence or mismatched specialist \\
Agent coordination failures & 18\% & Error propagation through sequential execution from early agents \\
Over-refinement             & 12\% & Verification stages introduce errors by over-analyzing initially correct answers \\
\bottomrule
\end{tabular}
}
\end{table}
\newpage
\subsection{Error Propagation Analysis}

Only 18\% of failures involve cross-stage error propagation, while 70\% originate from a single component (43\% reasoning, 27\% routing). This is partly by design: the stress testing and multi-turn debate stages generate new candidate answers rather than modifying existing ones, actively limiting rather than amplifying errors from earlier stages. The remaining 12\% of over-refinement errors are concentrated in questions with short, ambiguous answers where the debate mechanism incorrectly identifies uncertainty.

\subsection{Failure Case Examples}

\noindent\textbf{Reasoning error:} For questions involving indirect spatial references (e.g., ``What is the value in the row above the highlighted cell?''), the thinker agent occasionally misidentifies spatial relationships, producing a flawed reasoning path that directs specialized agents to the wrong document region.

\noindent\textbf{Router error:} Questions involving handwritten annotations embedded within printed tables are sometimes misrouted exclusively to the OCR agent, missing the table agent's structural extraction capability, resulting in partial answers that lack the necessary table context.

\noindent\textbf{Over-refinement error:} For yes/no questions where the expert answer is already correct, the antithesis agent occasionally generates a spurious alternative, initiating a debate that produces an incorrect final answer. Adding a confidence threshold for initiating debate on binary questions is a planned improvement.

\subsection{Implications for Future Work}

Given that 43\% of failures originate in the thinker agent, improving reasoning path generation represents the highest-leverage opportunity. Fine-tuning the thinker on document-specific reasoning traces is expected to yield the largest accuracy improvements. Router errors (27\%) suggest the routing training set would benefit from more diverse annotation of edge cases involving mixed-modality content.




 \fi

\end{document}